%% file: main.tex
\documentclass{article}

 \usepackage[preprint]{neurips_2026}

\usepackage[utf8]{inputenc} 
\usepackage[T1]{fontenc}    
\usepackage{hyperref}       
\usepackage{url}            
\usepackage{booktabs}       
\usepackage{amsfonts}       
\usepackage{amsmath}        
\usepackage{nicefrac}       
\usepackage{microtype}      
\usepackage[table]{xcolor}  
\usepackage{multirow}       
\usepackage{subcaption}     

\usepackage{graphicx}
\usepackage{longtable}
\usepackage{tabularx}
\usepackage{graphbox}
\usepackage{pifont}
\usepackage{tikz}
\usetikzlibrary{shapes.geometric, arrows.meta, positioning, calc, fit}
\usepackage{svg}
\usepackage{subcaption}

\newcommand{\cmark}{\textcolor{green!70!black}{\ding{51}}}
\newcommand{\xmark}{\textcolor{red!80!black}{\ding{55}}}
  
\newcommand{\ours}{Jolia}
\newcommand{\oursmethod}{ConQuer}
\newcommand{\oursmethodfull}{Concept Queries}
\newcommand{\ctrate}{CT-RATE}
\newcommand{\ccnchest}{EXT-Chest-CT}
\newcommand{\ccnabdo}{EXT-Abd-CT}
\newcommand{\stanford}{Merlin-Abd-CT}
\newcommand{\inspect}{INSPECT}

\title{Jolia: Concept-Level Vision-Language Alignment \\ for 3D CT Contrastive Learning}

\author{%
  \normalfont
  Julien Khlaut$^{1,2,3}$, Charles Corbi\`ere$^{*,1}$, Baptiste Callard$^{*,1}$, Amaury Prat$^{*,1}$, \\
  Leo Butsanets$^{*,1}$, Antoine Saporta$^{1}$, Th\'eo Danielou$^{1}$, Leo Machado$^{1,5}$, \\
  Korentin Le Floch$^{1,2,3,4}$, Tom Boeken$^{2,3,4}$, Pierre Manceron$^{1}$, Corentin Dancette$^{1}$ \\[5pt]
  {\small $^{*}$Equal contribution.} \\[3pt]
  {\small $^{1}$Raidium, Paris, 75014, France} \\
  {\small $^{2}$Department of Vascular and Oncological Interventional Radiology,} \\
  {\small H\^opital Europ\'een Georges Pompidou, AP-HP, Paris, France} \\
  {\small $^{3}$Facult\'e de Sant\'e, Universit\'e Paris-Cit\'e, Paris, France} \\
  {\small $^{4}$HEKA, INRIA, Paris, France} \\
  {\small $^{5}$Imaging Department, Fondation Ophtalmologique Adolphe de Rothschild, Paris, France} \\[3pt]
  \texttt{julien.khlaut@raidium.eu}
}

\begin{document}

\maketitle

\input{chapters/0.abstract}
\input{chapters/1.introduction}
\input{chapters/2.related_works}
\input{chapters/3.method}
\input{chapters/4.experiments}
\input{chapters/5.conclusion}



\bibliographystyle{unsrt}
\bibliography{references}
\newpage
\appendix

\input{chapters/appendix}

\clearpage

\end{document}

%% file: chapters/0.abstract.tex
\begin{abstract}
Vision-language contrastive pretraining has become the dominant recipe for 3D medical foundation models, leveraging the large volumes of paired scans and reports produced in clinical practice. 
However, medical images usually span dozens of organs, and radiological reports are much longer than typical natural image captions and are composed of multiple structured sections.
CLIP-style pretraining compresses this structure by encoding each modality into a single global token, at the risk of losing important details.
We introduce \oursmethod{} (\oursmethodfull{}), an image-text pretraining method that augments CLIP's global alignment with a set of localized alignments, one per \emph{concept}.  \oursmethod{} splits the report into concept-specific sections and learns cross-attention queries that pool the matching image features without using any segmentation mask or spatial supervision. Contrastive learning is then applied independently for each concept. Concepts can be any unit of semantic localization; here, they are anatomical regions, one query per organ or gross body region. As a byproduct, each query learns attention maps focused on its concept, providing built-in spatial interpretability.
We use \oursmethod{} to train \ours{}, a 3D CT foundation model on chest and abdominal CT. \ours{} consistently outperforms a CLIP baseline on findings classification, report generation, and cross-center transfer, and sets a new state of the art across multiple public benchmarks. \ours{}'s weights are available \href{https://huggingface.co/raidium/Jolia}{here}.
\end{abstract}

%% file: chapters/1.introduction.tex
\section{Introduction}
\label{sec:intro}

Foundation models~\cite{bommasani2021foundation} have transformed machine learning by demonstrating that large-scale pretraining on broad data, followed by lightweight adaptation, can outperform task-specific models trained from scratch. In medical imaging, this paradigm is particularly appealing: labeled data is scarce, annotation requires clinical expertise, and the diversity of downstream tasks (classification, segmentation, report generation, retrieval) makes it impractical to train dedicated models for each.
Contrastive vision-language (VL) pretraining~\cite{radford2021clip} sidesteps these constraints: it turns the image--report pairs already produced in clinical practice into free supervision, requiring no manual labels, and yields representations that transfer to linear probing, zero-shot classification, and cross-modal retrieval.

A fundamental limitation of standard CLIP-style alignment, however, is that both the image and the report are compressed into \emph{single} global vectors, while real inputs and their text typically decompose along structural axes that a single token cannot preserve~\cite{kang2025clipideal}. In CT studies, a single scan may span dozens of organs, and a report typically describes findings organ by organ (e.g., ``\textit{Lungs: no nodule. Liver: 2cm cyst in segment VII.}''), producing text much longer than in typical CLIP pretraining. Recent fine-grained 3D CT methods recover this lost structure using segmentation masks~\cite{fvlm2025, totalfm2026, ctglip2024}, but at the cost of organ-level supervision, which limits coverage to the organs the segmentation tool supports.

We propose \oursmethod{} (\oursmethodfull{}), a vision-language pretraining method that addresses these limitations by complementing the global CLIP objective with a parallel set of \emph{concept-level} contrastive alignments. Any semantic axis along which one might want to localize the image--text alignment can serve. In this work, concepts are anatomical regions, and we use \oursmethod{} to train \ours{}, a 3D foundation model, on public abdominal and chest CT datasets. Our key contributions are:
\begin{enumerate}

    \item \oursmethod{}, a concept-level contrastive pretraining method without spatial supervision. We complement the global CLIP objective with a parallel set of concept-level contrastive losses, learned end-to-end from raw volume--report pairs. 
    Each report is split by an LLM into concept-specific sections; on the image side, a small set of learnable cross-attention queries pools concept-specific features directly from the encoder, without requiring masks or spatial annotations. Contrastive alignment is then performed independently per concept, comparing the same concept across samples. 


    \item We use \oursmethod{} to train the \ours{} foundation model on 74,434 public chest and abdominal CT--report pairs from \ctrate{}~\cite{ctrate2024}, \inspect{}~\cite{inspect2023}, and \stanford{}~\cite{merlin2024}. Model weights will be released to the research community.

    \item We benchmark \ours{} against a CLIP baseline and competing models, and set a new state of the art on findings classification (linear probing and zero-shot, in- and out-of-distribution), cross-center transfer, and radiology report generation, while remaining competitive on image--text retrieval. We show that \oursmethod{} generalizes across architectures and concept choices.


\end{enumerate}

\begin{figure}[t]
    \centering
    \includegraphics[width=\linewidth]{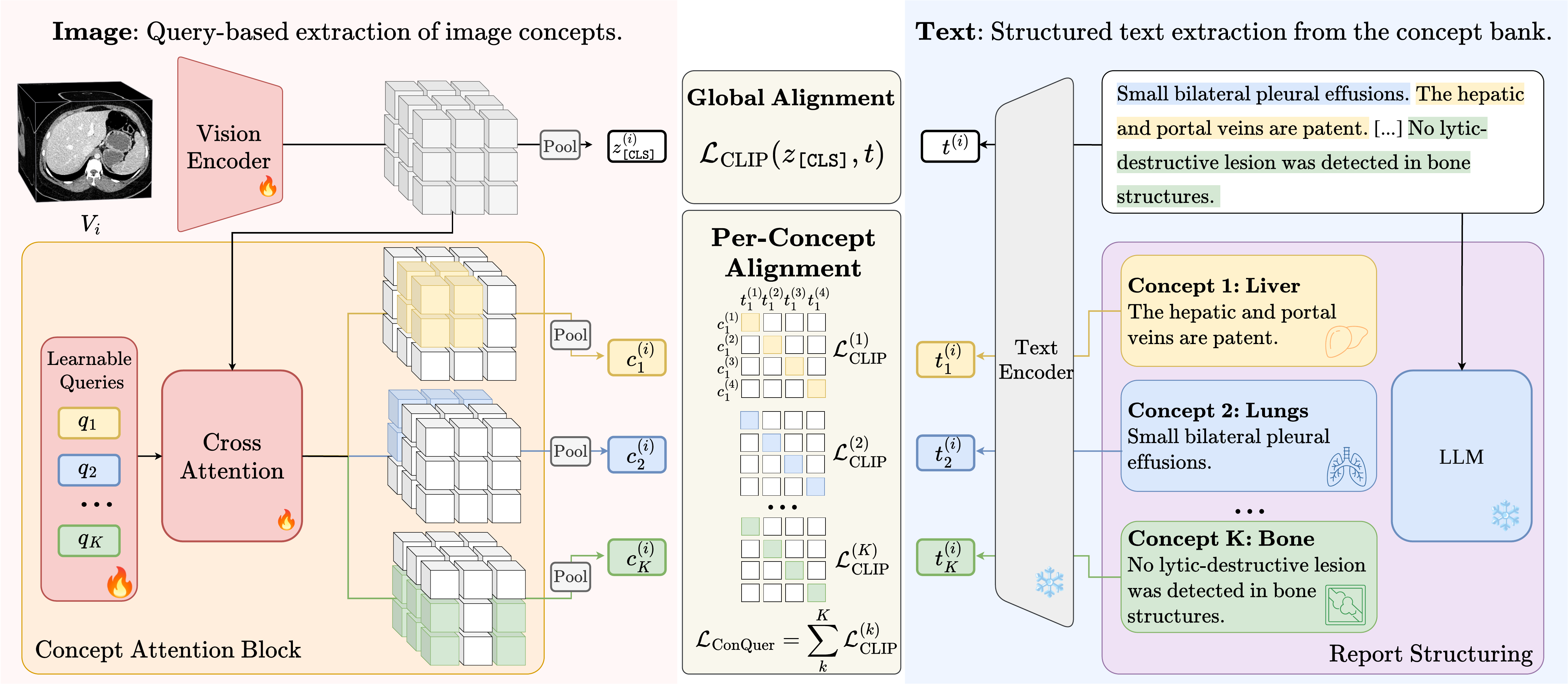}
    \caption{\textbf{Overview of \oursmethod{}.} We decompose both the image (via learnable cross-attention queries) and the report (via concept-specific sections produced by an LLM) into a set of concept-level representations, and apply contrastive alignment independently for each concept in addition to the standard global $\mathcal{L}_{\text{CLIP}}(z_\texttt{[CLS]},\,t)$.}
    \label{fig:method_overview_v3}
\end{figure}

%% file: chapters/2.related_works.tex
\section{Related Work}
\label{sec:related}

\textbf{Foundation models in medical imaging.}
Vision-language pretraining has become a popular recipe for medical foundation models~\cite{zhangBiomedCLIPMultimodalBiomedical2024,codellaMedImageInsightOpenSourceEmbedding2024,sellergrenMedGemmaTechnicalReport2025}, with paired images and clinical text routinely available from radiology workflows. In 3D CT specifically, three families have emerged: \emph{(1)} global CLIP-style alignment between volume and report (CT-CLIP~\cite{ctclip2024}, Merlin~\cite{merlin2024} on abdominal CT (whose dataset we use), Pillar-0~\cite{pillar0_2025}, COLIPRI~\cite{colipri2025}); \emph{(2)} self-supervised or multi-modal scaling without text-side locality (CT-FM~\cite{ctfm2025}, SPECTRE~\cite{spectre2025}, Curia~\cite{dancette2025curia}, Curia-2~\cite{saporta2026curia2}); and \emph{(3)} explicit anatomical decomposition via segmentation masks (fVLM~\cite{fvlm2025}, TotalFM~\cite{totalfm2026}, CT-GLIP~\cite{ctglip2024}).

\textbf{Local and region-based vision-language alignment.}
Several prior works recover the structure lost by global CLIP through finer-grained alignment: FILIP~\cite{filip2022} and DenseCLIP~\cite{denseclip2022} use token-wise or dense-pixel late interaction in natural images, while GLoRIA~\cite{huang2021gloria} and MGCA~\cite{mgca2022} extend the idea to chest X-rays at the word--patch and pathological-region level. \oursmethod{} differs from these along three axes. \emph{(i)} we produce one image and one text embedding \emph{per concept} and apply an independent contrastive objective per concept, comparing the same concept across samples, whereas global CLIP-style methods (CT-CLIP, Merlin, Pillar-0, BiomedCLIP) compress each volume and report into a single token. \emph{(ii)} unlike token-wise alignment (FILIP, GLoRIA, MGCA), where the alignment must emerge from the data, our queries impose an \emph{explicit} concept-level decomposition that mirrors how reports already split their content along anatomical lines. \emph{(iii)} unlike segmentation-based 3D approaches (fVLM, TotalFM, CT-GLIP), which depend on a pretrained 3D CT segmenter (e.g., RadSAM~\cite{radsam}) to crop organs before alignment, our queries learn where to attend purely through the per-concept contrastive loss, with no spatial supervision.

\textbf{Learnable-query visual representations.}
Compressing high-resolution visual inputs into a small set of learnable tokens via cross-attention is a well-established mechanism: DETR~\cite{detr2020} uses object queries for end-to-end detection, the Perceiver~\cite{jaegle2021perceiver} uses latent queries as a general bottleneck for arbitrary input modalities, and BLIP-2's Q-Former~\cite{blip2_2023} uses 32 learnable query tokens to bridge a frozen image encoder and a language model. Our cross-attention queries follow the same design, but anchor each query to a specific concept (in our CT setting, an anatomical region) and tie it to a per-concept contrastive objective rather than to a downstream detection or generation task.

%% file: chapters/3.method.tex
\section{Method}
\label{sec:method}

\oursmethod{} is a vision-language pretraining method that aligns volumes and reports at the level of \emph{concepts}, rather than at the global level alone.
A \emph{concept} is a single semantic unit along which we localize the alignment; in this paper, concepts are the $K$ entries of an anatomical taxonomy, typically a single organ (liver, spleen, pancreas), but in some cases an anatomical compartment grouping several organs (mediastinum, retroperitoneum).
The pipeline, illustrated in Figure~\ref{fig:method_overview_v3}, has three building blocks: (i) a data processing step that uses an LLM to label findings against a pre-defined taxonomy and split each report into \textbf{concept-specific sections}; (ii) an image encoder augmented with learnable cross-attention queries that pool \textbf{concept-specific image representations}; and (iii) a training objective combining a per-concept contrastive loss with a global CLIP loss. In this paper, we instantiate concepts as anatomical regions and use \oursmethod{} to train \ours{}, a 3D CT foundation model on chest and abdominal scans.


\subsection{Data Processing}
\label{subsec:data_processing}

Each paired radiology report is processed through a two-step LLM pipeline, using GPT-5.2. A first call infers the exam type and routes the report to a modality-specific finding taxonomy, developed with input from board-certified radiologists, for chest or abdominal CT examinations. A second call decomposes the \emph{findings} part of the report into atomic, single-sentence observations, each tagged with a finding category and an organ drawn from the modality taxonomy; an example of this decomposition and the full taxonomies are shown in Appendix~\ref{app:data_processing_details}.

The pipeline serves two purposes. For pretraining, findings are grouped by their mapped organ and remapped to the pre-defined concepts.
For downstream evaluation, we map each finding to a finding category,  defined by a unified taxonomy of $252$ labels ($172$ abdominal, $80$ chest) used by the linear-probing and zero-shot benchmarks of Section~\ref{subsec:findings_classification}. 
We validated the pipeline on 50 chest CT and 50 abdomen CT exams from the CT-RATE \cite{ctclip2024} and Merlin-Abd-CT \cite{merlin2024} datasets, respectively; a radiologist was tasked with flagging false-positive and false-negative findings. In those examples, the LLM-based finding extraction pipeline achieves 0.965/0.938 precision/recall on chest CT, and 0.975/0.951 on abdomen CT.

\subsection{Concept-Level Contrastive Pretraining}
\label{subsec:organ_contrastive}

We describe the four components of the pretraining pipeline in turn: the image encoder, the per-concept cross-attention pooling, the text encoder, and the contrastive alignment objective.

\textbf{Image encoder.}
The image encoder takes a CT volume resampled to $1.5$\,mm isotropic resolution and cropped to $192 \times 192 \times 192$ voxels, and produces a multi-scale sequence of patch token representations $\{z_1, \ldots, z_N\}$ at three feature resolutions together with a global \texttt{[CLS]} token $z_{\texttt{[CLS]}}$. We experiment with three backbones: a 3D Atlas transformer~\cite{Agrawal2025Atlas} (22\,M parameters, $6{\times}6{\times}6$ patches), a multi-scale 3D ResNet-101 (48\,M parameters), and a single-scale 3D ViT-B (120\,M parameters); the two multi-scale backbones produce patch tokens of dimension $d = 192$ per scale, while ViT-B yields a single set of patch tokens of the same dimension. The global \texttt{[CLS]} token is obtained by pooling and concatenating patch tokens across feature resolutions (three for Atlas/ResNet-101, one for ViT-B), giving a dimension of $d_\texttt{[CLS]}=576$ for the multi-scale variants. Encoders are trained from scratch.


\textbf{Concept-level cross-attention pooling.}
To extract a per-concept visual representation, we introduce $K$ learnable query tokens, one per concept. The image encoder produces patch tokens $\mathbf{z}\langle s\rangle = \{z_1\langle s\rangle, \ldots, z_{N_s}\langle s\rangle\}$ at $S$ scales (here, $S=3$ for Atlas and ResNet-101, $S=1$ for ViT-B). At each scale, we instantiate its own set of queries $Q^{(s)} = \{q_1^{(s)}, \ldots, q_K^{(s)}\}$ and apply an independent cross-attention module $\mathrm{CrossAttn}\langle s\rangle$, which first LayerNorms the queries and patch tokens and then performs a 1-head cross-attention of the queries onto the patches at that scale. The per-scale outputs are concatenated into the final per-concept representation,
\begin{equation}
    c_k = \big[\,\mathrm{CrossAttn}\langle s \rangle\bigl(q^{(s)}_k,\, \mathbf{z}\langle s \rangle\bigr)\,\big]_{s=1}^{S} \in \mathbb{R}^{576}.
\end{equation}
Each query learns where to attend through the per-concept contrastive loss alone, with no spatial supervision. The pooling stack ($K$ queries plus the cross-attention modules at every scale) adds only $\sim$0.6\,M parameters, roughly $3\%$ of Atlas and $1\%$ of ResNet-101.

\textbf{Text encoder.}
Following Pillar-0~\cite{pillar0_2025}, we use a frozen pretrained Qwen3-Embedding-8B~\cite{qwen3_embedding} as our text encoder. Each per-concept report section is encoded independently, yielding for each sample a set of $K$ text representations $\{t_1, \ldots, t_K\}$; we additionally encode the full (unsegmented) report to obtain a global text representation $t$ used by the global CLIP loss below. The 4096-dimensional embeddings are projected to the visual feature dimension via a single learnable linear layer, giving $t_k, t \in \mathbb{R}^{576}$.

\textbf{Concept-level contrastive alignment and training objective.}
Given a batch $b$ of image--text pairs, we form per-concept representations $(c_k^{(i)}, t_k^{(i)})$ for each sample $i$ and concept $k$. Reports do not always mention every concept, so a concept presence mask keeps only $(i, k)$ pairs with a non-empty text section. The per-concept contrastive loss is a standard symmetric InfoNCE applied independently per concept and combined with a global CLIP loss $\mathcal{L}_\text{CLIP}^{\texttt{[CLS]}}$ between $z^{(i)}_\texttt{[CLS]}$ and the embedding $t^{(i)}$ of the full report:
\begin{equation}
    \mathcal{L}_\text{\oursmethod{}} = \frac{1}{|\mathcal{K}_b|} \sum_{k \in \mathcal{K}_b} \mathcal{L}^{(k)}_{\text{CLIP}},
    \qquad
    \mathcal{L} = \lambda_\text{CLIP}\,\mathcal{L}_\text{CLIP}^{\texttt{[CLS]}} + \lambda_\text{\oursmethod{}}\,\mathcal{L}_\text{\oursmethod{}},
\end{equation}
where $\mathcal{K}_b$ is the set of concepts with at least two valid samples in the batch and $\mathcal{L}^{(k)}_{\text{CLIP}}$ is the symmetric InfoNCE on the similarity matrix $(S_k)_{ij} = \mathrm{sim}(c_k^{(i)}, t_k^{(j)})$ scaled by a learnable per-concept temperature. We use $\lambda_\text{CLIP} = 1.0$ and $\lambda_\text{\oursmethod{}} = 0.5$ in all experiments. Because the per-concept loss only ever compares the same concept across samples, each query is pushed to encode features that distinguish patients within its own concept rather than features of the volume as a whole.

\subsection{Evaluation Protocol}
\label{subsec:evaluation}

\begin{figure}[t]
    \centering
    \includegraphics[width=1.0\linewidth]{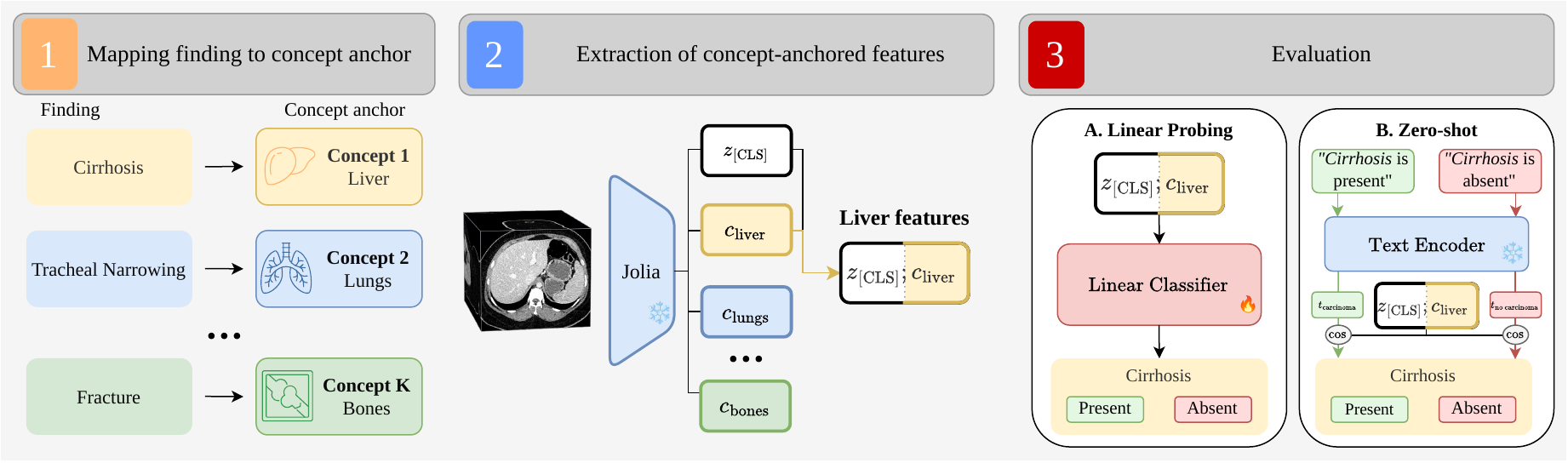}
    \caption{Evaluation protocol for findings classification (linear probing and zero-shot): we concatenate \ours{}'s global \texttt{[CLS]} and concept tokens into a finding-anchored representation $[z_\texttt{[CLS]};\, c_k]$.}
    \label{fig:organ_anchor}
\end{figure}

\ours{} exposes three frozen representations for downstream tasks: the global \texttt{[CLS]} token $z_\texttt{[CLS]}$ summarising the whole volume, the $K$ concept tokens $\{c_k\}$ pooled by the cross-attention queries, and their concatenation $[z_\texttt{[CLS]};\, c_k]$ that combines both sides, illustrated in Figure~\ref{fig:organ_anchor}. 

\textbf{Linear probing.} Each finding is mapped to a single concept (e.g.\ \textit{hepatic cyst} $\rightarrow$ liver), and \ours{}'s headline configuration feeds the combined representation $[z_\texttt{[CLS]};\, c_k]$ (with $k$ the concept associated with the finding) to a linear classifier (Figure~\ref{fig:organ_anchor}). We additionally report two ablation variants, \ours{} - [CLS] (only $z_\texttt{[CLS]}$) and \ours{} - Query (only $c_k$), to isolate what each side contributes.


\textbf{Zero-shot classification.}
Zero-shot evaluation lacks a canonical protocol, and we identify two key concerns when it is applied to radiological data. 
First, exact prompt wording shifts model rankings, making comparisons depend on prompt choice rather than representation quality. 
CT-CLIP~\cite{ctclip2024} uses a single prompt per finding, while Merlin~\cite{merlin2024} extends this with several radiologist-written prompts; however, neither yields a uniform, prompt-agnostic evaluation. Some prior work mitigates this by augmenting pretraining with prompt-like
statements and dedicated losses (e.g., COLIPRI's
$[\text{finding}]$\,/\,$[\emptyset\;\text{finding}]$ contrastive objective),
which may bias scores toward that specific prompt.
Second, there is a distribution shift between prompts and reports: for VLMs pretrained on full reports, short prompts lie far from the long-form clinical text seen during training. To address this we propose the following zero-shot evaluation schemes: \\
\emph{\textbf{Short.}}
To tackle prompt sensitivity, for each finding we define a broad set of paired positive and negative prompt variants asserting presence vs.\ absence, extending those used by Merlin~\cite{merlin2024} and CT-CLIP~\cite{ctclip2024}. Aggregating across sets removes dependence on any single wording, enabling a more fair comparison between models.
\\
\emph{\textbf{Long.}} Inspired from COLIPRI~\cite{colipri2025}, to reduce the domain shift between short prompts and capture the understanding of the full radiological reports seen at training time that is missing in the short protocol, we replace templates with empirical prototypes: for each finding, 50 reports from patients with the abnormality and
50 from patients without are embedded and averaged within each group, yielding a single positive and negative reference embedding.

\textbf{Report generation.}
We condition an autoregressive language model on the $z_\texttt{[CLS]}$ token, projected through a 2-layer MLP and prepended to the text prompt as a visual token; the per-concept tokens $\{c_k\}$ are not used in this setup. The LM is trained using a two-stage pipeline. We give additional details in Appendix~\ref{app:rrg_details}.

%% file: chapters/4.experiments.tex
\section{Experiments}
\label{sec:experiments}

\subsection{Experimental Setup}
\label{subsec:exp_setup}

\textbf{Pre-Training}
We instantiate \oursmethod{} with $K=102$ anatomical concepts spanning chest and abdomen (full list in Table~\ref{tab:anatomical_taxonomy} of Appendix~\ref{app:data_processing_details}), and train \ours{} using the Atlas~\cite{Agrawal2025Atlas} architecture on $74{,}434$ chest and abdominal CT--report pairs from \ctrate{}~\cite{ctrate2024} and \inspect{}~\cite{inspect2023} (chest) and \stanford{}~\cite{merlin2024} (abdomen). We train ResNet and ViT-Based variants as well (Section~\ref{subsec:ablations}). All models are trained using 8 NVIDIA H100 GPUs with a global batch size of 48 for 120,000 steps with AdamW. Training hyperparameters are detailed in Appendix~\ref{app:training_details}.

\textbf{Baselines.}
We compare \ours{} against recent CT foundation models (CT-CLIP~\cite{ctclip2024}, CT-FM~\cite{ctfm2025}, fVLM~\cite{fvlm2025}, COLIPRI~\cite{colipri2025}, SPECTRE~\cite{spectre2025}, Merlin~\cite{merlin2024}, and the chest and abdominal Pillar-0 variants~\cite{pillar0_2025}); a side-by-side comparison of their pretraining data and methodological choices is in Appendix~\ref{app:baseline_comparison}. To isolate the contribution of $\mathcal{L}_\text{\oursmethod{}}$, we additionally train an in-house \emph{Baseline CLIP} variant: same encoders and optimisation, $\mathcal{L}_\text{global}$ alone.

\textbf{Evaluation}
We benchmark three tasks: findings classification in multiple settings (linear-probing, zero-shot, and cross-center transfer), radiology report generation and image--text retrieval (reported in Appendix~\ref{app:retrieval}).
All numbers reported in the result tables are mean\,$\pm$\, std over five seeds for \ours{} and Baseline CLIP, 
Evaluation uses each dataset's held-out test split (in-distribution, ID) and two private clinical datasets from a separate center, \ccnchest{} and \ccnabdo{} (out-of-distribution, OOD); volume counts and finding taxonomies for each dataset are listed in Table~\ref{tab:datasets} (Appendix~\ref{app:dataset_statistics}).

\subsection{Findings Classification}
\label{subsec:findings_classification}


\textbf{Linear probing.} We train linear classifiers on frozen representations for binary findings classification (Section~\ref{subsec:evaluation}) and report AUROC in Table~\ref{tab:findings_classification_results}.
The per-concept loss is the main driver. Against our in-house Baseline CLIP (same encoders, $\mathcal{L}_\text{global}$ alone), \ours{} gains $+1.7$ AUROC on average ($+1.9$/$+2.4$ abdomen, $+1.0$/$+1.5$ chest), largest on the OOD splits; \ours{} - [CLS] alone already improves Baseline CLIP by $+0.4$ to $+1.0$, and adding the queries adds a further $+0.2$ to $+1.8$ (per-finding breakdown in Appendix~\ref{app:per_finding_breakdown}).
\ours{} beats the strongest public baseline by $+2.2$ on \ctrate{} (over Pillar-0-Chest) and $+2.0$ on \ccnabdo{} (over SPECTRE), and crucially the segmentation-based fVLM (evaluated without masks) by $+8.2$/$+9.4$ on chest. Seed-to-seed variance is also markedly smaller than for most baselines ($0.01$--$0.12$ vs up to $2.4$ for CT-FM and $0.6$ for SPECTRE/COLIPRI).

\input{tables/findings_classification_results}

\textbf{Comparison on different architectures.}
Table~\ref{tab:ablation_loss_eval} extends the linear-probing comparison of Table~\ref{tab:findings_classification_results} to two additional backbones (a plain ViT-B and a multi-scale 3D ResNet-101) and to all three ways of combining \ours{}'s representation at inference (\texttt{[CLS]} alone, Query alone, \texttt{[CLS]}+Query).
We show that using \oursmethod{} improves the \texttt{[CLS]} representation on every dataset for the Atlas backbone~\cite{Agrawal2025Atlas} and ResNet-101; ViT-B sits $3$--$5$ AUROC below either on every column, and the gain on top is smaller and mixed.
\texttt{[CLS]}+Query is the best evaluation; Query alone is competitive. The combined row outperforms either token alone in every (architecture, dataset) cell. On ResNet-101, even Query alone beats the global CLIP baseline on every column. We see similar results for Atlas except on \ctrate{}, whose $18$-finding lung/pleural-centric taxonomy leaves less room for per-concept pooling.
Atlas is best architecture on three of four columns and within $0.2$ AUROC of ResNet-101 on the fourth, while ViT-B trails by $3$--$6$ AUROC, 
We therefore use Atlas as the main architecture for \ours{}.

\input{tables/ablation_loss_eval}

\textbf{Cross-center transfer learning.}
We test whether the frozen representations transfer across clinical centers by training a linear probe on a in-training dataset (\stanford{} for abdomen, \ctrate{} for chest) and evaluating on an OOD test set from a separate institution (\ccnabdo{} and \ccnchest{}, respectively), under a unified taxonomy.
Results are in Table~\ref{tab:findings_taxonomy_comparison}. 
Overall, \ours{} beats all other baselines on the external benchmarks (77.05 / 75.88 without and with queries), as well as on source datasets. Against our Baseline CLIP, $\mathcal{L}_\text{\oursmethod{}}$ adds $+2.7$ AUROC on average ($+2.9$/$+4.0$ abdomen, $+1.6$/$+2.5$ chest), echoing the results of Table~\ref{tab:findings_classification_results}.
\ours{} is best on both source datasets ($+1.9$ over Pillar-0-Abd on \stanford{}, $+1.7$ over Pillar-0-:qChest on \ctrate{}). Our baseline and Jolia obtain lower results on EXT-Chest-CT compared to Pillar-0, which suggests a difference in quality between the pre-training dataset.

\input{tables/findings_taxonomy_comparison}



\subsection{Zero-shot classification}

\textbf{Short zero-shot classification.} We evaluate zero-shot abnormality detection using 8 prompt templates
(Appendix~\ref{app:zero_shot_templates}). As shown in
Figure~\ref{fig:zs_combined}(a), model performance could be highly sensitive to
template choice, particularly for Merlin and Pillar variants (best-to-worst
gap up to $24$ AUROC points), undermining fair and robust evaluation. To
eliminate validation-based prompt selection, we aggregate general prompts via
mean cosine similarity across positive and negative variations, removing the
need for prompt selection and yielding a more reliable cross-model comparison. This is further detailed in Appendix~\ref{app:zero_shot_short}).
Among the evaluated models, \ours{} and COLIPRI stand out for their
robustness to prompt choice. Results are reported in
Table~\ref{tab:findings_classification_results_zeroshot}.

\ours{} narrowly leads abdomen ($+0.3$ over Merlin on \stanford{}; \ours{} - \texttt{[CLS]} $+0.03$ over Pillar-0-Abd on \ccnabdo{}) and trails only COLIPRI on chest ($-2.8$ on \ctrate{}, $-4.0$ on \ccnchest{}). COLIPRI is the only baseline trained with zero-shot-specific text augmentation (\textsuperscript{$\star$}). Notably, we observe a bias toward the "\{$\emptyset$/no\} findings" phrasing encountered during pretraining. Best-prompt selection thus inflates performance by exploiting this artifact rather than measuring generalization, making it an unfair evaluation protocol for such methods.

\textbf{Long zero-shot classification.} As shown in Figure~\ref{fig:zs_combined}(b), AUROC improves steadily across all models as the number $r$ of reports used to create the prototype grows. In this setup, \ours{} shows a robust and strong understanding of fine-grained differences between long-report texts. With $r=50$, \ours{} leads among external baselines on \ctrate{} ($79.05$ vs Pillar-0-Chest's $78.60$) and remains competitive on \stanford{} with $83.03$, sitting behind Merlin ($84.26$). More details are reported in Appendix~\ref{app:zero_shot_long}.

\input{figures/zero_shot_robustness_heatmap}

\subsection{Radiology Report Generation}
\label{subsec:report_generation}

We pair the frozen \ours{} encoder with a Qwen3.5-9B~\cite{qwen35} language model and fine-tune it to generate the \emph{Findings} section of an abdominal CT report from the volume alone, using two-stage training (projector alignment then LoRA fine-tuning, vision encoder frozen throughout). At inference, each concept section is generated in its own autoregressive pass following~\cite{merlin2024}. We evaluate on the \stanford{} test split with LLM-as-judge clinical fidelity (GREEN~\cite{ostmeier2024green}, CRIMSON~\cite{baharoon2026crimson}), clinical entity overlap (RadGraph-F1~\cite{jain2021radgraph}), and standard NLG metrics (BLEU, ROUGE-L, BERTScore), all via the RadEval framework~\cite{xu2025radeval}, against Merlin~\cite{merlin2024} (trained in-domain on this dataset), the generalist 2D MedGemma-1.5~\cite{sellergrenMedGemmaTechnicalReport2025}, and Med3DVLM~\cite{xin2025med3dvlm}. Full details are in Appendix~\ref{app:rrg_details}.

\input{tables/rrg_stanford_results}

\ours{} is the strongest model on this benchmark (Table~\ref{tab:rrg_results_stanford}), leading on four of six metrics with a $+34\%$ relative gain on RadGraph-F1 over Merlin and the best GREEN score across all baselines. Qualitative inspection (Appendix~\ref{app:rrg_details}) nonetheless confirms a bias already noted by Merlin~\cite{merlin2024}: the model underreports positive findings, defaulting to normal descriptions; CRIMSON scores remain negative across all models.

\subsection{Further analysis}
\label{subsec:ablations}

\textbf{Concept-grouping ablation.} We probe how much the recipe depends on the specific choice of concepts by unsupervised K-means clusters, predefined anatomical groupings at two granularities ($K\in\{10,32\}$), and our default $K=102$ taxonomy (Table~\ref{tab:ablation_grouping}). Every variant improves over the global CLIP baseline, even with arbitrary concepts: K-means grouping adds $\approx +1$ AUROC, confirming that the load-bearing ingredient is localization itself, not the specific concepts. Anatomical groupings sharpen the signal further (best: $K=32$ and $K=102$, tied), so meaningful concepts add a second-order gain on top.

\input{tables/ablation_grouping}

\textbf{Interpretability of the learned queries.} Each concept query $q_k$ produces an attention map highlighting where the model looks for concept $k$ (Figure~\ref{fig:attention_maps}); failure cases on less precisely localized regions and a PCA comparison of the encoders' feature maps are in Appendix~\ref{app:interpretability_failures} and~\ref{app:pca_visualization}.

\input{figures/attention_maps}

%% file: tables/findings_classification_results.tex
\begin{table}[t]
    \centering
    \caption{\textbf{Linear Probing Finding classification results} (\% AUROC). \ccnchest{} and \ccnabdo{} are unseen during training for all models and use our in-house taxonomy of 172 abdomen and 80 chest findings; \stanford{} and \ctrate{} use their original taxonomies. \textsuperscript{\textdagger}Pillar-0-Best is the combination of the two Pillar models on their respective domains.
    Results in {\color{gray}grey} signify an evaluation outside the model's pre-training domain.
    }
    \label{tab:findings_classification_results}
    \resizebox{\textwidth}{!}{%
    \setlength{\tabcolsep}{4pt}
    \renewcommand{\arraystretch}{0.92}
    \begin{tabular}{l c c c c c}
        \toprule
        & \multicolumn{2}{c}{\textbf{Abdomen}}
        & \multicolumn{2}{c}{\textbf{Chest}}
        & \multirow{2}{*}{\textbf{Average}} \\
        \cmidrule(lr){2-3} \cmidrule(lr){4-5}
        \textbf{Model}
        & \stanford{}
        & \ccnabdo{}
        & \ctrate{}
        & \ccnchest{}
        & \\
        \midrule

        CT-CLIP~\cite{ctclip2024}
        & \textcolor{gray}{64.94} {\scriptsize \textcolor{gray}{$\pm$0.13}}
        & \textcolor{gray}{60.76} {\scriptsize \textcolor{gray}{$\pm$0.25}}
        & 76.28 {\scriptsize \textcolor{gray}{$\pm$0.03}}
        & 79.40 {\scriptsize \textcolor{gray}{$\pm$0.06}}
        & 70.35 \\

        CT-FM~\cite{ctfm2025}
        & 70.41 {\scriptsize \textcolor{gray}{$\pm$0.13}}
        & 64.16 {\scriptsize \textcolor{gray}{$\pm$2.08}}
        & 76.08 {\scriptsize \textcolor{gray}{$\pm$1.26}}
        & 79.65 {\scriptsize \textcolor{gray}{$\pm$2.41}}
        & 72.58 \\

        fVLM~\cite{fvlm2025}
        & \textcolor{gray}{64.52} {\scriptsize \textcolor{gray}{$\pm$0.35}}
        & \textcolor{gray}{60.45} {\scriptsize \textcolor{gray}{$\pm$0.76}}
        & 78.29 {\scriptsize \textcolor{gray}{$\pm$0.16}}
        & 79.65 {\scriptsize \textcolor{gray}{$\pm$ 0.60 }}
        & 70.73 \\

        COLIPRI~\cite{colipri2025}
        & \textcolor{gray}{75.28} {\scriptsize \textcolor{gray}{$\pm$0.12}}
        & \textcolor{gray}{66.46} {\scriptsize \textcolor{gray}{$\pm$0.24}}
        & 83.23 {\scriptsize \textcolor{gray}{$\pm$0.06}}
        & 82.81 {\scriptsize \textcolor{gray}{$\pm$0.36}}
        & 76.94 \\

        SPECTRE~\cite{spectre2025}
        & 78.82 {\scriptsize \textcolor{gray}{$\pm$0.19}}
        & 75.39 {\scriptsize \textcolor{gray}{$\pm$0.62}}
        & 82.44 {\scriptsize \textcolor{gray}{$\pm$0.06}}
        & 86.09 {\scriptsize \textcolor{gray}{$\pm$0.25}}
        & 80.69 \\

        Merlin~\cite{merlin2024}
        & 82.93 {\scriptsize \textcolor{gray}{$\pm$0.03}}
        & 72.75 {\scriptsize \textcolor{gray}{$\pm$0.07}}
        & \textcolor{gray}{76.59} {\scriptsize \textcolor{gray}{$\pm$0.15}}
        & \textcolor{gray}{81.55} {\scriptsize \textcolor{gray}{$\pm$0.20}}
        & 78.45 \\

        Pillar-0-Abd~\cite{pillar0_2025}
        & 83.18 {\scriptsize \textcolor{gray}{$\pm$0.04}}
        & 75.31 {\scriptsize \textcolor{gray}{$\pm$0.13}}
        & \textcolor{gray}{75.45} {\scriptsize \textcolor{gray}{$\pm$0.04}}
        & \textcolor{gray}{84.26} {\scriptsize \textcolor{gray}{$\pm$0.09}}
        & 79.55 \\

        Pillar-0-Chest~\cite{pillar0_2025}
        & \textcolor{gray}{77.69} {\scriptsize \textcolor{gray}{$\pm$0.06}}
        & \textcolor{gray}{69.13} {\scriptsize \textcolor{gray}{$\pm$0.11}}
        & 84.27 {\scriptsize \textcolor{gray}{$\pm$0.04}}
        & 88.94 {\scriptsize \textcolor{gray}{$\pm$0.08}}
        & 80.01 \\

        Pillar-0-Best\textsuperscript{\textdagger}~\cite{pillar0_2025}
        & 83.18 {\scriptsize \textcolor{gray}{$\pm$0.04}}
        & 75.31 {\scriptsize \textcolor{gray}{$\pm$0.13}}
        & 84.27 {\scriptsize \textcolor{gray}{$\pm$0.04}}
        & 88.94 {\scriptsize \textcolor{gray}{$\pm$0.08}}
        & 82.92 \\

        \midrule

        CLIP Baseline
        & 81.66 {\scriptsize \textcolor{gray}{$\pm$0.18}}
        & 75.01 {\scriptsize \textcolor{gray}{$\pm$0.79}}
        & 85.48 {\scriptsize \textcolor{gray}{$\pm$0.06}}
        & 87.53 {\scriptsize \textcolor{gray}{$\pm$0.31}}
        & 82.42 \\

        \ours{} - [CLS]
        & 82.69 {\scriptsize \textcolor{gray}{$\pm$0.11}}
        & 75.64 {\scriptsize \textcolor{gray}{$\pm$0.55}}
        & 86.23 {\scriptsize \textcolor{gray}{$\pm$0.05}}
        & 87.91 {\scriptsize \textcolor{gray}{$\pm$0.62}}
        & 83.12 \\

        \textbf{\ours{} \textit{(Ours)}}
        & \textbf{83.59} {\scriptsize \textcolor{gray}{$\pm$0.01}}
        & \textbf{77.39} {\scriptsize \textcolor{gray}{$\pm$0.07}}
        & \textbf{86.44} {\scriptsize \textcolor{gray}{$\pm$0.03}}
        & \textbf{89.06} {\scriptsize \textcolor{gray}{$\pm$0.12}}
        & \textbf{84.12} \\




        \bottomrule
    \end{tabular}
    }
\end{table}

%% file: tables/ablation_loss_eval.tex
\begin{table}[h]
\caption{\textbf{Ablation of alignment strategy, loss, and evaluation strategy.} Linear probing AUROC (\%). We progressively add our contributions on top of a global CLIP baseline. Architectures are listed by ascending overall performance (Atlas, the headline backbone, last).}
\label{tab:ablation_loss_eval}
\resizebox{\textwidth}{!}{%
\setlength{\tabcolsep}{4pt}
\renewcommand{\arraystretch}{0.92}
\begin{tabular}{lllrrrr}
\toprule
\multicolumn{1}{c}{} & \multicolumn{1}{c}{} & \multicolumn{1}{c}{}  & \multicolumn{2}{c}{\textbf{Abdomen}} & \multicolumn{2}{c}{\textbf{Chest}}    \\
\cmidrule(lr){4-5} \cmidrule(lr){6-7} \multicolumn{1}{c}{Archi.} & \multicolumn{1}{l}{Loss} & \multicolumn{1}{l}{Eval method} & \multicolumn{1}{c}{\stanford} & \multicolumn{1}{c}{\ccnabdo} & \multicolumn{1}{c}{\ctrate} & \multicolumn{1}{c}{\ccnchest} \\
\midrule
\multirow{4}{*}[-0.5em]{\shortstack[l]{ViT-B\\[-1pt]{\scriptsize\textcolor{gray}{120M}}}}
& CLIP & \texttt{[CLS]}
& 78.13 {\scriptsize \textcolor{gray}{$\pm$0.06}}
& 72.02 {\scriptsize \textcolor{gray}{$\pm$0.23}}
& 81.03 {\scriptsize \textcolor{gray}{$\pm$0.02}}
& 85.96 {\scriptsize \textcolor{gray}{$\pm$0.13}} \\
\cmidrule(lr){2-7}
& \multirow{3}{*}{CLIP + \oursmethod{}}
& \texttt{[CLS]}
& 78.52 {\scriptsize \textcolor{gray}{$\pm$0.04}}
& 71.73 {\scriptsize \textcolor{gray}{$\pm$0.49}}
& 80.40 {\scriptsize \textcolor{gray}{$\pm$0.02}}
& 86.28 {\scriptsize \textcolor{gray}{$\pm$0.15}} \\
&
& Query
& 77.89 {\scriptsize \textcolor{gray}{$\pm$0.01}}
& 70.67 {\scriptsize \textcolor{gray}{$\pm$0.13}}
& 78.99 {\scriptsize \textcolor{gray}{$\pm$0.04}}
& 84.97 {\scriptsize \textcolor{gray}{$\pm$0.01}} \\
&
& \texttt{[CLS]} + Query
& 78.66 {\scriptsize \textcolor{gray}{$\pm$0.01}}
& 72.63 {\scriptsize \textcolor{gray}{$\pm$0.13}}
& 80.04 {\scriptsize \textcolor{gray}{$\pm$0.02}}
& 86.23 {\scriptsize \textcolor{gray}{$\pm$0.01}} \\
\midrule
\multirow{4}{*}[-0.5em]{\shortstack[l]{ResNet-101\\[-1pt]{\scriptsize\textcolor{gray}{48.2M}}}}
& CLIP & \texttt{[CLS]}
& 82.08 {\scriptsize \textcolor{gray}{$\pm$0.11}}
& 74.70 {\scriptsize \textcolor{gray}{$\pm$0.62}}
& 85.45 {\scriptsize \textcolor{gray}{$\pm$0.04}}
& 86.84 {\scriptsize \textcolor{gray}{$\pm$0.36}} \\
\cmidrule(lr){2-7}
& \multirow{3}{*}{CLIP + \oursmethod{}}
& \texttt{[CLS]}
& 82.40 {\scriptsize \textcolor{gray}{$\pm$0.13}}
& 76.05 {\scriptsize \textcolor{gray}{$\pm$0.31}}
& 85.93 {\scriptsize \textcolor{gray}{$\pm$0.03}}
& 87.89 {\scriptsize \textcolor{gray}{$\pm$0.16}} \\
&
& Query
& 82.89 {\scriptsize \textcolor{gray}{$\pm$0.02}}
& 76.55 {\scriptsize \textcolor{gray}{$\pm$0.12}}
& 85.80 {\scriptsize \textcolor{gray}{$\pm$0.03}}
& 87.94 {\scriptsize \textcolor{gray}{$\pm$0.01}} \\
&
& \texttt{[CLS]} + Query
& 83.35 {\scriptsize \textcolor{gray}{$\pm$0.02}}
& \textbf{77.53} {\scriptsize \textcolor{gray}{$\pm$0.07}}
& 86.15 {\scriptsize \textcolor{gray}{$\pm$0.02}}
& 88.58 {\scriptsize \textcolor{gray}{$\pm$0.08}} \\
\midrule
\multirow{4}{*}[-0.5em]{\shortstack[l]{Atlas\\[-1pt]{\scriptsize\textcolor{gray}{21.8M}}}}
& CLIP & \texttt{[CLS]}
& 81.66 {\scriptsize \textcolor{gray}{$\pm$0.18}}
& 75.01 {\scriptsize \textcolor{gray}{$\pm$0.79}}
& 85.48 {\scriptsize \textcolor{gray}{$\pm$0.06}}
& 87.53 {\scriptsize \textcolor{gray}{$\pm$0.31}} \\
\cmidrule(lr){2-7}
& \multirow{3}{*}{CLIP + \oursmethod{}}
& \texttt{[CLS]}
& 82.69 {\scriptsize \textcolor{gray}{$\pm$0.11}}
& 75.64 {\scriptsize \textcolor{gray}{$\pm$0.55}}
& 86.23 {\scriptsize \textcolor{gray}{$\pm$0.05}}
& 87.91 {\scriptsize \textcolor{gray}{$\pm$0.62}} \\
&
& Query
& 83.12 {\scriptsize \textcolor{gray}{$\pm$0.02}}
& 76.35 {\scriptsize \textcolor{gray}{$\pm$0.07}}
& 85.38 {\scriptsize \textcolor{gray}{$\pm$0.02}}
& 87.87 {\scriptsize \textcolor{gray}{$\pm$0.15}} \\
&
& \texttt{[CLS]} + Query
& \textbf{83.59} {\scriptsize \textcolor{gray}{$\pm$0.01}}
& 77.39 {\scriptsize \textcolor{gray}{$\pm$0.07}}
& \textbf{86.44} {\scriptsize \textcolor{gray}{$\pm$0.03}}
& \textbf{89.06} {\scriptsize \textcolor{gray}{$\pm$0.12}} \\
\bottomrule
\end{tabular}
}
\end{table}

%% file: tables/findings_taxonomy_comparison.tex
\begin{table}[t]
    \centering
    \caption{\textbf{Cross-center transfer learning performance} (\% AUROC). In-distribution (ID) results on source test sets and out-of-distribution (OOD) results on target sets, aligned to a unified taxonomy. Transfer scenarios: (1) \stanford{} $\rightarrow$ \ccnabdo{} and (2) \ctrate{} $\rightarrow$ \ccnchest{}.
    Results in {\color{gray}grey} signify an evaluation outside the model's pre-training domain.
    Avg EXT is the mean of \ccnabdo{} and \ccnchest{}. Per-column best is in bold; values within one standard deviation of the best are also in bold. \textsuperscript{\textdagger}Pillar-0-Best is the combination of the two Pillar models on their respective domains.
    }
    \label{tab:findings_taxonomy_comparison}
    \small
    \setlength{\tabcolsep}{4pt}
    \renewcommand{\arraystretch}{0.92}
    \begin{tabular}{l c c c c c}
        \toprule
        & \multicolumn{2}{c}{\textbf{Abdomen}} & \multicolumn{2}{c}{\textbf{Chest}} & \\
        \cmidrule(lr){2-3} \cmidrule(lr){4-5}
        \textbf{Model}  & \stanford & \ccnabdo & \ctrate & \ccnchest & \textbf{Avg EXT} \\
        \midrule

        CT-CLIP~\cite{ctclip2024}
        & \textcolor{gray}{62.56} {\scriptsize \textcolor{gray}{$\pm$0.10}}
        & \textcolor{gray}{61.34} {\scriptsize \textcolor{gray}{$\pm$0.33}}
        & 75.54 {\scriptsize \textcolor{gray}{$\pm$0.23}}
        & 58.16 {\scriptsize \textcolor{gray}{$\pm$0.52}}
        & \textcolor{gray}{59.75} \\

        CT-FM~\cite{ctfm2025}
        & 63.92 {\scriptsize \textcolor{gray}{$\pm$1.23}}
        & 60.57 {\scriptsize \textcolor{gray}{$\pm$0.86}}
        & 71.87 {\scriptsize \textcolor{gray}{$\pm$2.01}}
        & 63.01 {\scriptsize \textcolor{gray}{$\pm$1.13}}
        & 61.79 \\

        fVLM~\cite{fvlm2025}
        & \textcolor{gray}{58.99} {\scriptsize \textcolor{gray}{$\pm$0.43}}
        & \textcolor{gray}{56.59} {\scriptsize \textcolor{gray}{$\pm$0.46}}
        & 77.03 {\scriptsize \textcolor{gray}{$\pm$0.33}}
        & 59.12 {\scriptsize \textcolor{gray}{$\pm$1.02}}
        & \textcolor{gray}{57.86} \\

        COLIPRI~\cite{colipri2025}
        & \textcolor{gray}{69.62} {\scriptsize \textcolor{gray}{$\pm$0.26}}
        & \textcolor{gray}{67.44} {\scriptsize \textcolor{gray}{$\pm$0.16}}
        & 82.91 {\scriptsize \textcolor{gray}{$\pm$0.23}}
        & 75.05 {\scriptsize \textcolor{gray}{$\pm$0.26}}
        & \textcolor{gray}{71.25} \\

        SPECTRE~\cite{spectre2025}
        & 78.77 {\scriptsize \textcolor{gray}{$\pm$0.19}}
        & 71.90 {\scriptsize \textcolor{gray}{$\pm$0.62}}
        & 82.25 {\scriptsize \textcolor{gray}{$\pm$0.35}}
        & 72.12 {\scriptsize \textcolor{gray}{$\pm$0.88}}
        & 72.01 \\

        Merlin~\cite{merlin2024}
        & 79.45 {\scriptsize \textcolor{gray}{$\pm$0.04}}
        & 71.98 {\scriptsize \textcolor{gray}{$\pm$0.11}}
        & \textcolor{gray}{75.61} {\scriptsize \textcolor{gray}{$\pm$0.29}}
        & \textcolor{gray}{63.48} {\scriptsize \textcolor{gray}{$\pm$0.38}}
        & \textcolor{gray}{67.73} \\

        Pillar-0-Abd~\cite{pillar0_2025}
        & 80.49 {\scriptsize \textcolor{gray}{$\pm$0.16}}
        & \textbf{76.07} {\scriptsize \textcolor{gray}{$\pm$0.12}}
        & \textcolor{gray}{75.26} {\scriptsize \textcolor{gray}{$\pm$0.20}}
        & \textcolor{gray}{68.19} {\scriptsize \textcolor{gray}{$\pm$0.15}}
        & \textcolor{gray}{72.13} \\

        Pillar-0-Chest~\cite{pillar0_2025}
        & \textcolor{gray}{73.10} {\scriptsize \textcolor{gray}{$\pm$0.03}}
        & \textcolor{gray}{68.74} {\scriptsize \textcolor{gray}{$\pm$0.17}}
        & 84.71 {\scriptsize \textcolor{gray}{$\pm$0.11}}
        & 73.92 {\scriptsize \textcolor{gray}{$\pm$0.42}}
        & \textcolor{gray}{71.33} \\

        Pillar-0-Best\textsuperscript{\textdagger}~\cite{pillar0_2025}
        & 80.49 {\scriptsize \textcolor{gray}{$\pm$0.16}}
        & \textbf{76.07} {\scriptsize \textcolor{gray}{$\pm$0.12}}
        & 84.71 {\scriptsize \textcolor{gray}{$\pm$0.11}}
        & 73.92 {\scriptsize \textcolor{gray}{$\pm$0.42}}
        & 75.00 \\

        \midrule
        CLIP Baseline & 79.57 {\scriptsize \textcolor{gray}{$\pm$0.25}} & 71.00 {\scriptsize \textcolor{gray}{$\pm$0.58}} & 84.83 {\scriptsize \textcolor{gray}{$\pm$0.57}} & 74.22 {\scriptsize \textcolor{gray}{$\pm$1.40}} & 72.61 \\
        \ours{} - [CLS] & \textbf{82.44} {\scriptsize \textcolor{gray}{$\pm$0.10}} & 74.13 {\scriptsize \textcolor{gray}{$\pm$0.27}} & \textbf{86.48} {\scriptsize \textcolor{gray}{$\pm$0.09}} & \textbf{79.97} {\scriptsize \textcolor{gray}{$\pm$0.30}} & \textbf{77.05} \\
        \textbf{\ours{}} & \textbf{82.43} {\scriptsize \textcolor{gray}{$\pm$0.14}} & 75.02 {\scriptsize \textcolor{gray}{$\pm$0.27}} & \textbf{86.40} {\scriptsize \textcolor{gray}{$\pm$0.12}} & 76.74 {\scriptsize \textcolor{gray}{$\pm$0.25}} & 75.88 \\
        \bottomrule
    \end{tabular}
\end{table}

%% file: figures/zero_shot_robustness_heatmap.tex
\begin{figure}[htbp]
    \centering


    \begin{tikzpicture}
        \node[anchor=south west, inner sep=0] (image) at (0,0) {\includegraphics[width=0.59\textwidth]{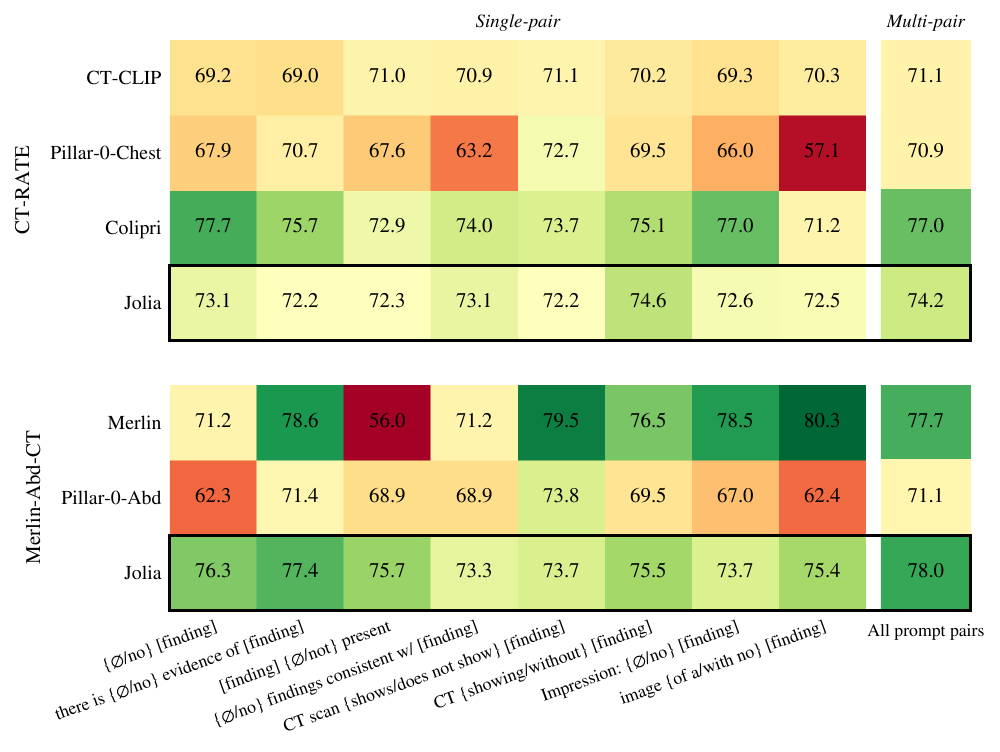}};
        \begin{scope}[x={(image.south east)},y={(image.north west)}]
            \node[anchor=north west, font=\bfseries] at (0.02,0.03) {(a)};
        \end{scope}
    \end{tikzpicture}
    \begin{tikzpicture}
        \node[anchor=south west, inner sep=0] (image1) at (0,0) 
            {\includegraphics[width=0.4\textwidth]{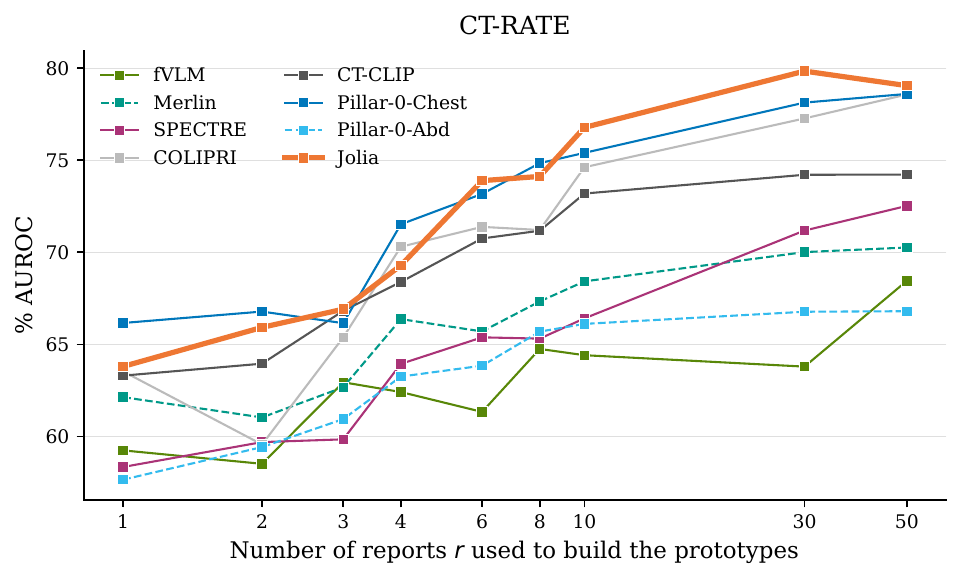}};
            
        \node[anchor=north west, inner sep=0] (image2) at (0,0) 
            {\includegraphics[width=0.4\textwidth]{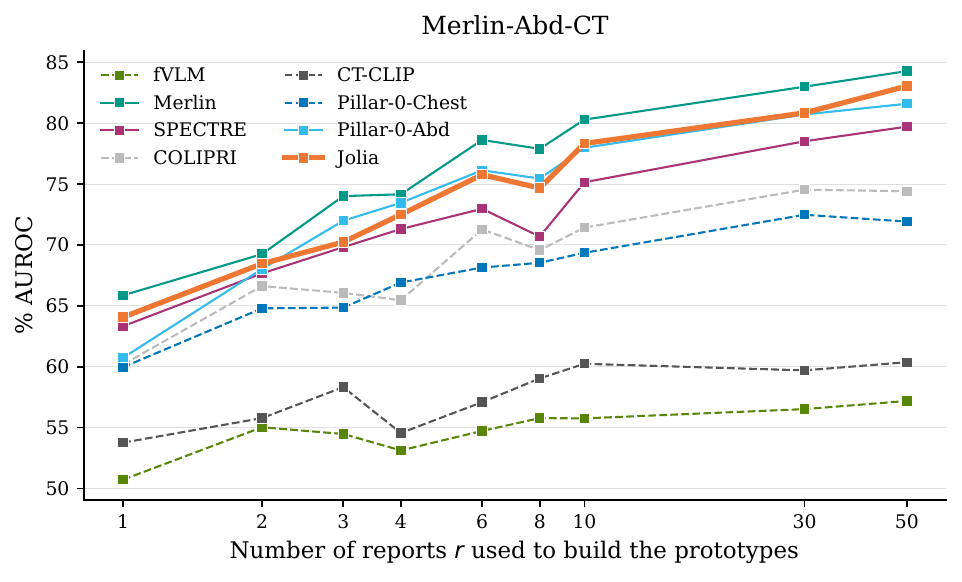}};
    
        \node[anchor=south west, font=\bfseries] at ([xshift=0pt, yshift=0pt]image2.south west) {(b)};
    \end{tikzpicture}
    
    \caption{\textbf{Zero-shot evaluation results.} \textbf{(a)} Per-template and aggregate zero-shot AUROC on the test set.
        Each cell reports the macro-AUROC obtained with a single prompt template
        pair (columns 1--8) or their aggregate (rightmost column),
        evaluated on CT-RATE (top) and Stanford Abd CT (bottom). \textbf{(b)} Ablation study on the number of prompt templates ($r$) in long-form zero-shot evaluation.}
    \label{fig:zs_combined}
\end{figure}

%% file: tables/rrg_stanford_results.tex
\begin{table}[h]
    \centering
    \caption{\textbf{Report generation on \stanford{}} test split ($N{=}5{,}125$) for the findings section. Higher is better; CRIMSON ranges over $[-1, 1]$ with $0$ corresponding to a normal report.}
    \label{tab:rrg_results_stanford}
    \resizebox{\textwidth}{!}{%
    \setlength{\tabcolsep}{6pt}
    \begin{tabular}{l cccccc}
        \toprule
        \textbf{Model} & BLEU & ROUGE-L & BERTScore & RadGraph-F1 & GREEN & CRIMSON \\
        \midrule
        Med3DVLM~\cite{xin2025med3dvlm}                          & 0.004 & 0.079 & 0.256 & 0.010 & 0.007 & $-0.545$ \\
        MedGemma-1.5~\cite{sellergrenMedGemmaTechnicalReport2025} & \textbf{0.124} & 0.106 & 0.361 & 0.032 & 0.038 & $-0.526$ \\
        Merlin~\cite{merlin2024}                                 & 0.063 & 0.285 & 0.551 & 0.237 & 0.316 & $\mathbf{-0.138}$ \\
        \midrule
        \textbf{\ours{}} \textit{(Ours)}                         & 0.119 & \textbf{0.323} & \textbf{0.567} & \textbf{0.317} & \textbf{0.324} & $-0.194$ \\
        \bottomrule
    \end{tabular}
    }
\end{table}

%% file: tables/ablation_grouping.tex
\begin{table}[t]
    \centering
    \caption{\textbf{Concept grouping granularity.} Linear probing AUROC (\%). Even arbitrary concept groupings  K-means improve over the CLIP baseline; natural anatomical groupings further sharpen the signal.}
    \label{tab:ablation_grouping}
    \resizebox{\textwidth}{!}{%
    \begin{tabular}{l c cc cc c}
        \toprule
        & & \multicolumn{2}{c}{\textbf{Abdomen}} & \multicolumn{2}{c}{\textbf{Chest}} & \\
        \cmidrule(lr){3-4} \cmidrule(lr){5-6}
        \textbf{Grouping} & \textbf{Queries} & Merlin-Abd-CT & EXT-Abd-CT & CT-RATE & EXT-Chest-CT & \textbf{Average} \\
        \midrule
        CLIP Baseline & --- & 82.08 {\scriptsize \textcolor{gray}{$\pm$0.11}}
& 74.70 {\scriptsize \textcolor{gray}{$\pm$0.62}}
& 85.45 {\scriptsize \textcolor{gray}{$\pm$0.04}}
& 86.84 {\scriptsize \textcolor{gray}{$\pm$0.36}}
& 82.27 \\
        \midrule
        Kmeans        & 32  & 82.56 {\scriptsize \textcolor{gray}{$\pm$0.02}} & 77.36 {\scriptsize \textcolor{gray}{$\pm$0.04}} & 85.37 {\scriptsize \textcolor{gray}{$\pm$0.04}} & 88.14 {\scriptsize \textcolor{gray}{$\pm$0.03}} & 83.36 \\
        \midrule
        \multirow{2}{*}{Natural}
                      & 10  & \textbf{83.38} {\scriptsize \textcolor{gray}{$\pm$0.02}} & 77.14 {\scriptsize \textcolor{gray}{$\pm$0.06}} & 85.89 {\scriptsize \textcolor{gray}{$\pm$0.03}} & 88.91 {\scriptsize \textcolor{gray}{$\pm$0.14}} & 83.83 \\
                      & 32  & 83.10 {\scriptsize \textcolor{gray}{$\pm$0.05}} & 78.03 {\scriptsize \textcolor{gray}{$\pm$0.07}} & 85.53 {\scriptsize \textcolor{gray}{$\pm$0.03}} & \textbf{88.92} {\scriptsize \textcolor{gray}{$\pm$0.06}} & \textbf{83.90} \\
        \midrule
        Fine (ours)   & 102 & \textbf{83.35} {\scriptsize \textcolor{gray}{$\pm$0.02}} & 77.53 {\scriptsize \textcolor{gray}{$\pm$0.07}} & \textbf{86.15} {\scriptsize \textcolor{gray}{$\pm$0.02}} & 88.58 {\scriptsize \textcolor{gray}{$\pm$0.08}} & \textbf{83.90} \\
        \bottomrule
    \end{tabular}
    }
\end{table}

%% file: figures/attention_maps.tex
\begin{figure}[h]
    \centering
    \setlength{\tabcolsep}{0.01pt} 
    \renewcommand{\arraystretch}{1.2} 

    \newlength{\gridwidth}
    \setlength{\gridwidth}{\dimexpr (\textwidth - 2em - 12\tabcolsep) / 5 \relax}

    \newcommand{\gridhdr}[1]{\footnotesize\textbf{#1}}
    \newcommand{\rowlabel}[1]{\rotatebox{90}{\makebox[0pt]{\gridhdr{#1}}}}

    \begin{tabular}{@{} c @{\hspace{4pt}}ccccc @{}}
        & \gridhdr{Liver} & \gridhdr{Lungs} & \gridhdr{Kidneys} & \gridhdr{Colon} & \gridhdr{Hip} \\[2pt]

        \rowlabel{Axial} &
        \includegraphics[width=\gridwidth, align=c]{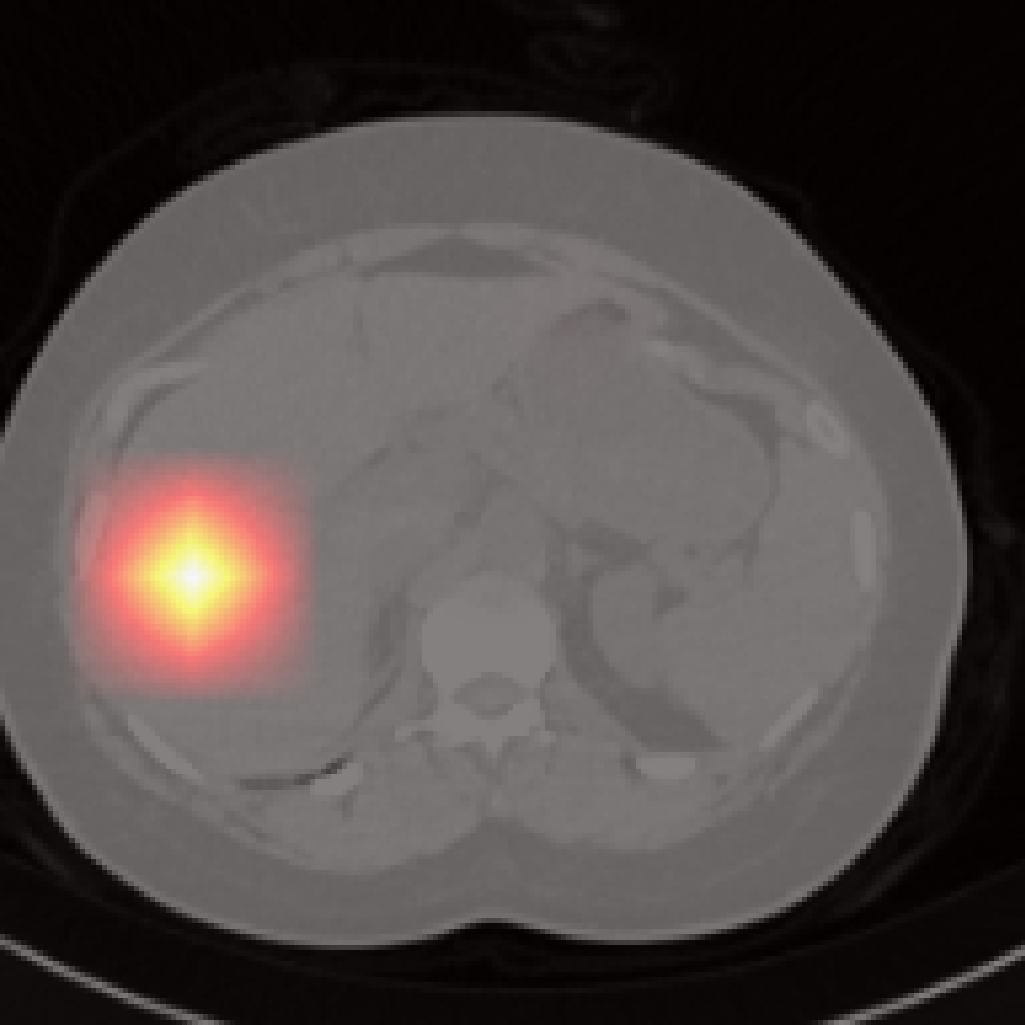} &
        \includegraphics[width=\gridwidth, align=c]{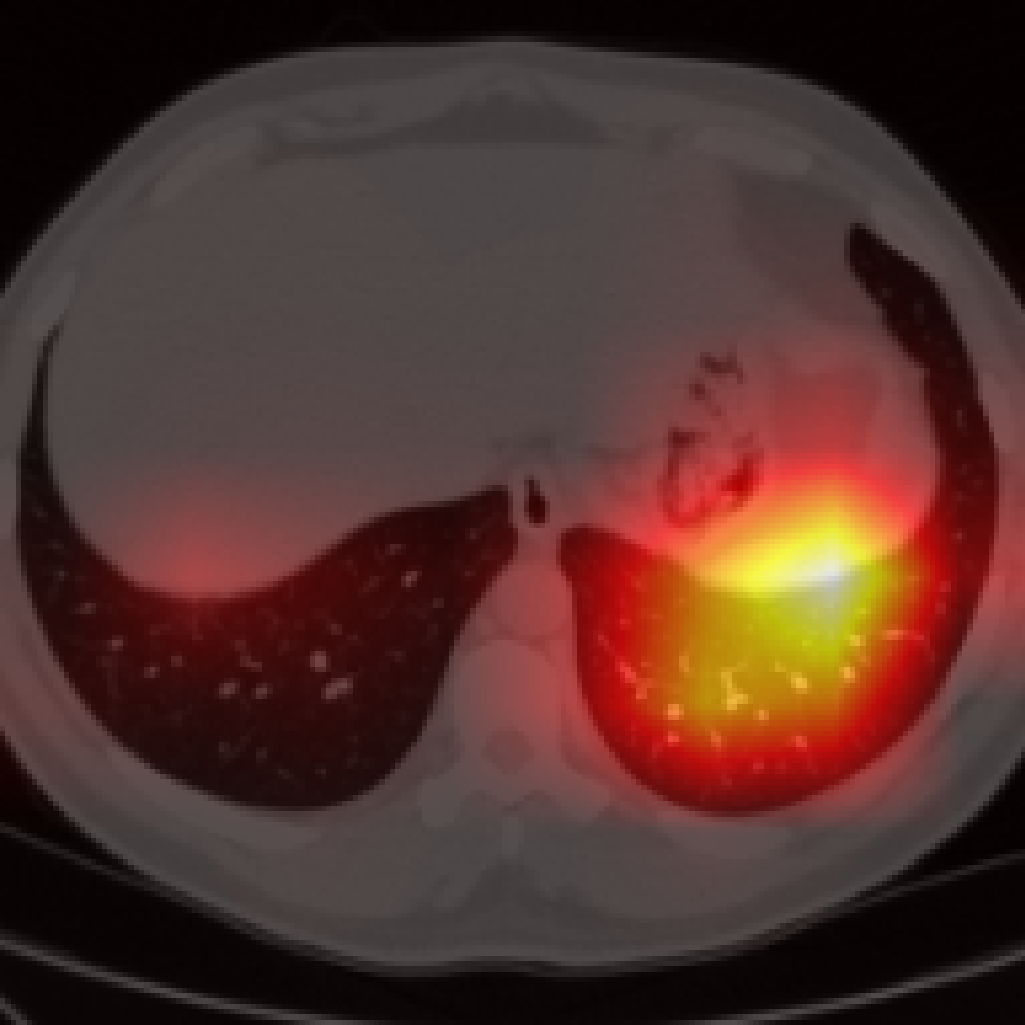} &
        \includegraphics[width=\gridwidth, align=c]{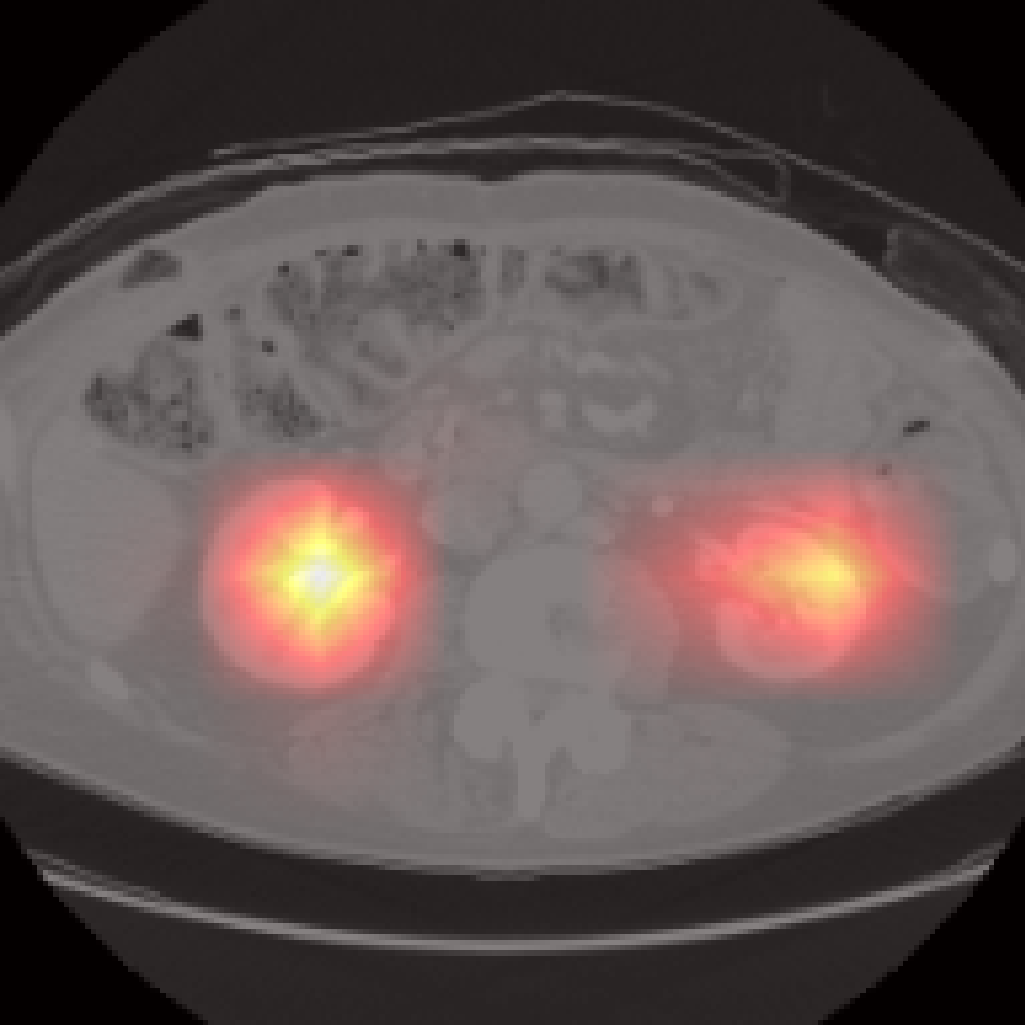} &
        \includegraphics[width=\gridwidth, align=c]{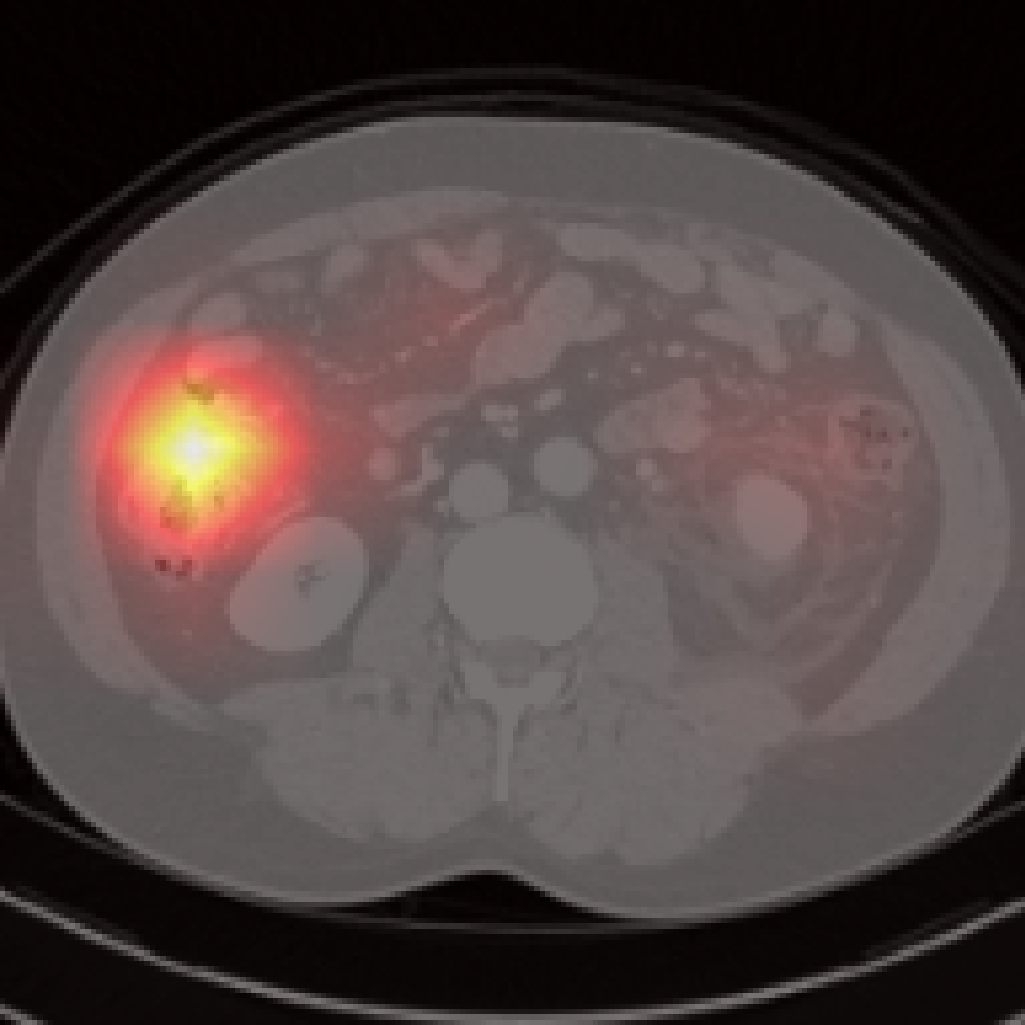} &
        \includegraphics[width=\gridwidth, align=c]{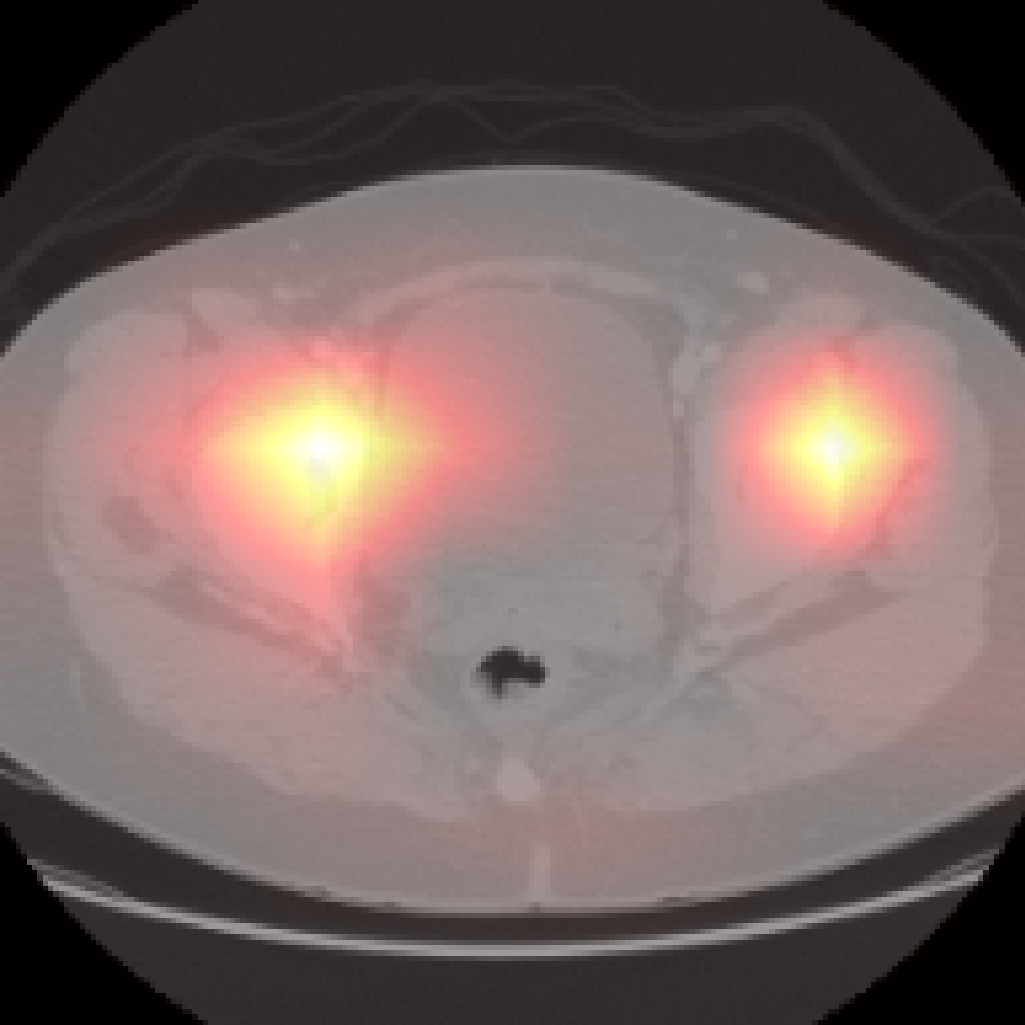} \\[6pt]

        \rowlabel{Coronal} &
        \includegraphics[width=\gridwidth, align=c]{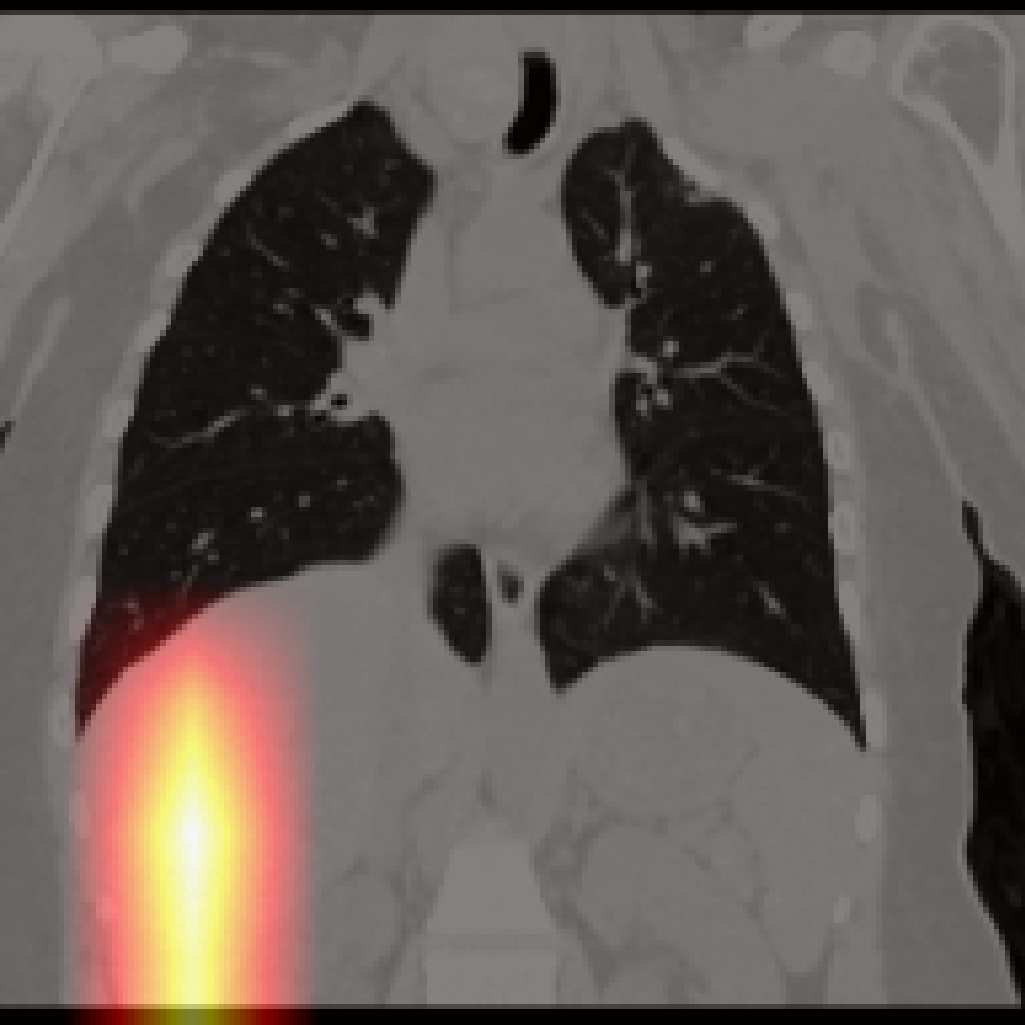} &
        \includegraphics[width=\gridwidth, align=c]{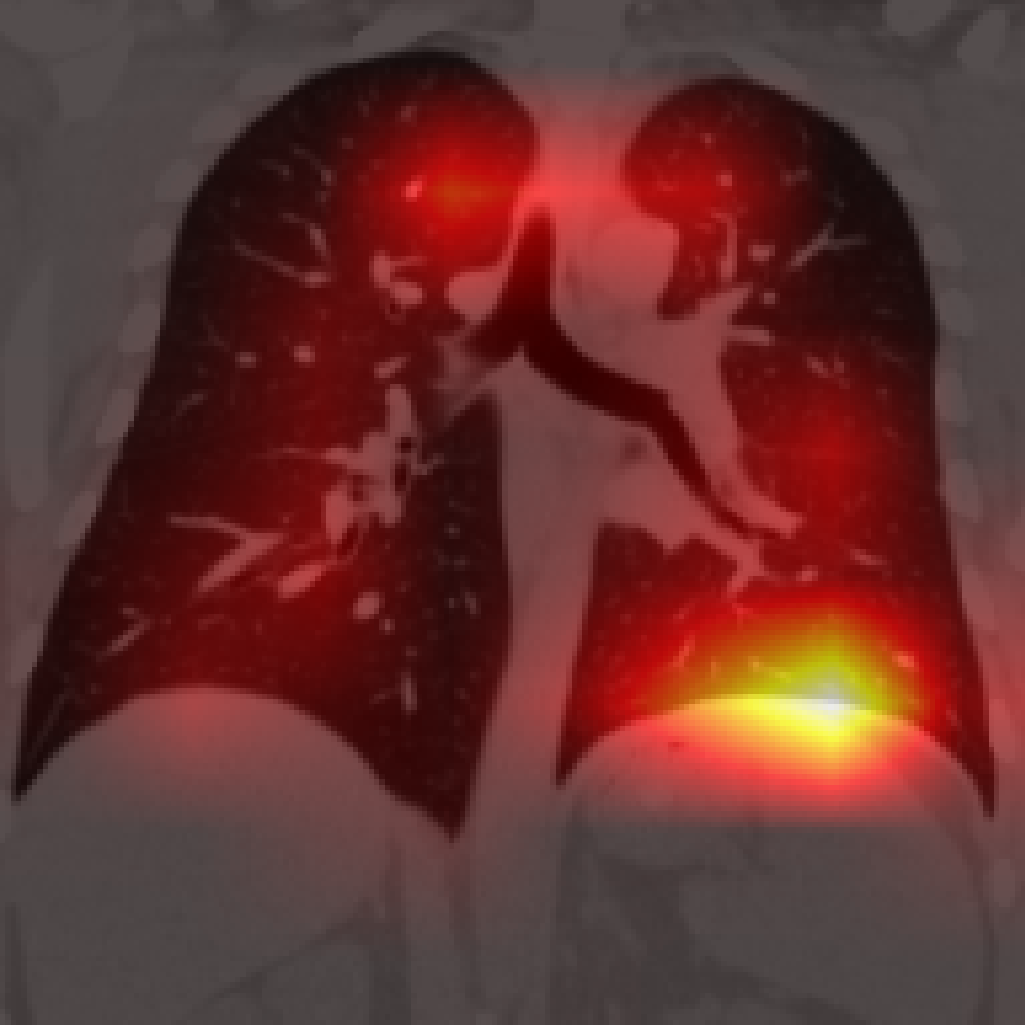} &
        \includegraphics[width=\gridwidth, align=c]{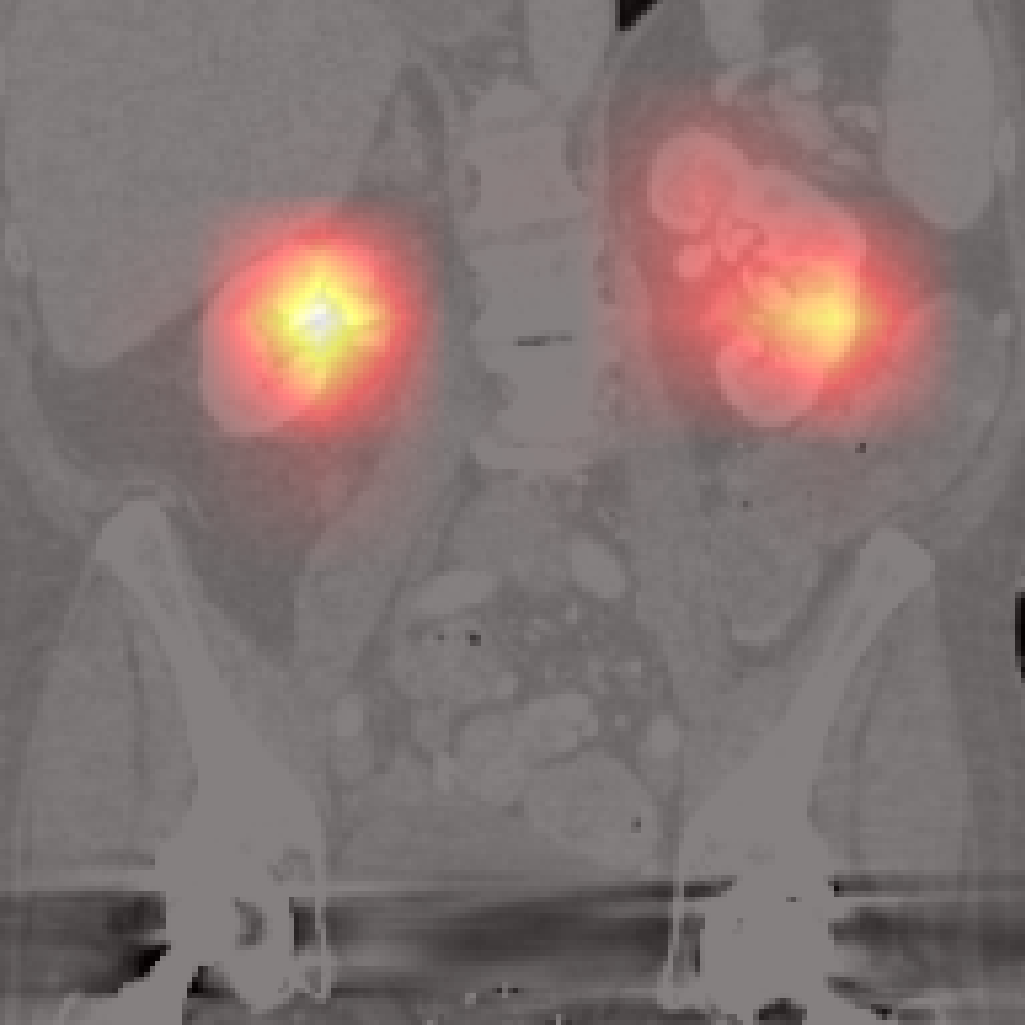} &
        \includegraphics[width=\gridwidth, align=c]{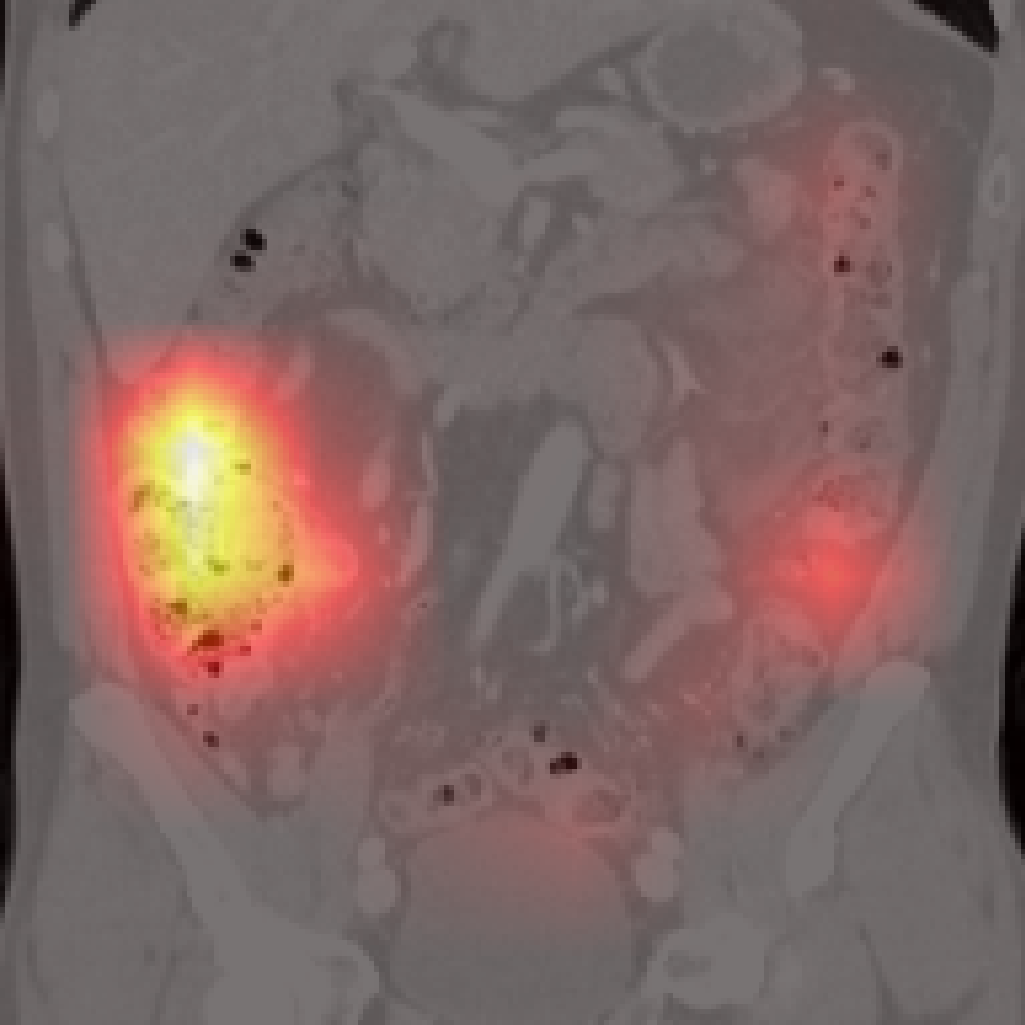} &
        \includegraphics[width=\gridwidth, align=c]{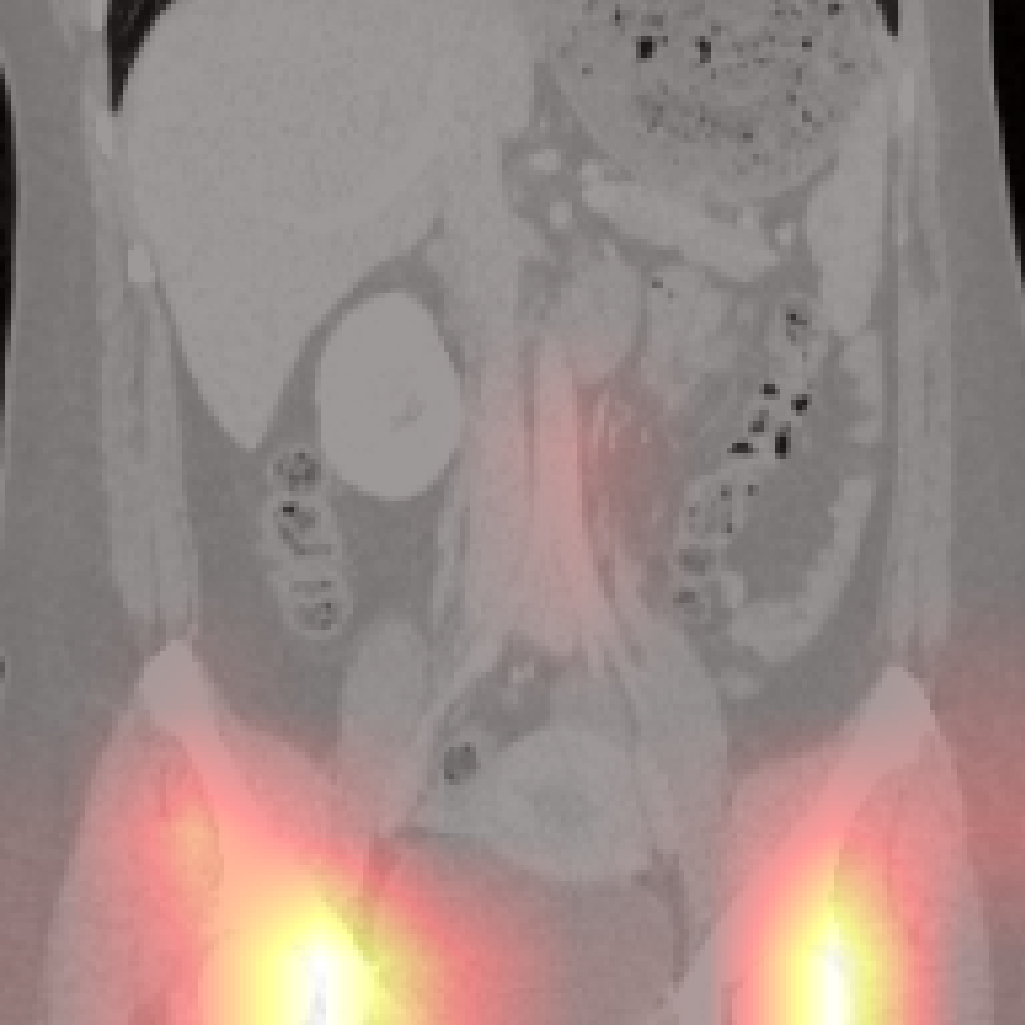} \\[6pt]

    \end{tabular}

    \caption{\textbf{Organ-level attention maps.} Cross-attention maps for different organ queries show that Jolia models learn to attend to anatomically meaningful regions without explicit supervision.}
    \label{fig:attention_maps}
\end{figure}

%% file: chapters/5.conclusion.tex
\section{Conclusion}
\label{sec:conclusion}

We introduced \oursmethod{} (\oursmethodfull{}), an image--text contrastive pretraining method that augments global CLIP alignment with localized, concept-level alignments learned from concept-specific report sections via cross-attention queries, without spatial supervision. By instanciating \oursmethod{} on anatomical regions, we train \ours{}, a 3D CT foundation model for abdominal and chest CT that sets a new state of the art on findings classification (linear probing, zero-shot, in- and out-of-distribution), cross-center transfer, and radiology report generation. Ablations show that the method is robust to architectural changes and various concept granularities. 
Our approach has a few limitations worth noting: first, gains are somewhat smaller on chest CT, where reports tend to be less detailed and span fewer organs than abdominal ones, leaving less per-concept structure for the queries to leverage. Second, the report-splitting pipeline relies on a large LLM (GPT-5.2); how sensitive the recipe is to the choice of splitter is an open question we leave to future work. Beyond these, a natural opening lies on the text side: we keep Qwen3-Embedding-8B frozen and do not train any language model, so radiology reports remain out-of-distribution for the text encoder; jointly training or fine-tuning a domain-adapted text encoder is a promising direction for further gains. Additionally, a promising direction is scaling \oursmethod{} to whole-body CT, instantiating concepts beyond anatomy (e.g., tumour subtypes), and extending the recipe to other 3D medical modalities like MRI.

\section*{Aknowledgments}
This work was granted access to the HPC resources of IDRIS under the allocations 2026-A0201017510, 2026-AS011017799,  2026-AD011017510R1, 2026-AD011017373 made by GENCI.

%% file: chapters/appendix.tex

\section*{Appendix}
\section{Dataset Statistics}
\label{app:dataset_statistics}

\input{tables/datasets}

\section{Data Processing}
\label{app:data_processing_details}

\paragraph{Pipeline.}
Both pipeline calls use GPT-5.2. All reports are anonymised (dates, names, and identifiers redacted) prior to LLM processing. Reports from \ccnabdo{} and \ccnchest{} originate from a French clinical institution and were translated to English with GPT-5 and manually verified by a board-certified radiologist before running the structuring pipeline.

\paragraph{Taxonomies.}
We maintain two modality-specific finding-category taxonomies relevant to this paper: abdomen ($172$ findings across $22$ anatomical groups) and chest ($80$ findings across $10$ groups), together yielding the $252$ binary findings reported in Section~\ref{subsec:data_processing}. Both share an organ taxonomy of $87$ entries covering chest and abdominal anatomy (the head region of the full taxonomy is not used in this paper).

\paragraph{Per-organ section construction.}
The fine-grained organ taxonomy is collapsed onto the $K = 102$ anatomical concepts used by our learnable queries (Section~\ref{subsec:organ_contrastive}; full list in Table~\ref{tab:anatomical_taxonomy}) via a hand-written mapping (e.g., \emph{liver}, \emph{biliary tree}, and \emph{hepatic vein} all map to the \emph{liver} concept). Each per-organ section fed to the text encoder is the concatenation of the \texttt{text} fields of all findings sharing the same concept; empty sections are masked out of the per-concept contrastive loss. 

\input{tables/anatomical_taxonomy}

\paragraph{Example of an organ-section decomposition.}
\label{app:report_split_example}

Figure~\ref{fig:report_split_example} illustrates the LLM-based report-splitting pipeline described in Section~\ref{subsec:data_processing} on a representative chest CT study. The left panel shows the raw \emph{Findings} part of a report (paraphrased and shortened from a public \ctrate{} study to fit the page); the right panel shows the corresponding output of the pipeline, with each atomic finding tagged with a \texttt{finding\_category} and mapped to a single entry of our $K = 102$ anatomical taxonomy. For pretraining, sentences sharing the same anatomical entry are concatenated into a single \emph{organ section} (e.g., the two lung sentences are merged into the \texttt{Lungs} section), and each section is independently embedded by the frozen text encoder before the per-organ contrastive loss.

\begin{figure}[h]
\centering
\fbox{\begin{minipage}{0.46\textwidth}
\textbf{Raw \emph{Findings} part of the report} \\[2pt]
\rule{\linewidth}{0.4pt}
\small\itshape
The thyroid gland is normal in size and contours. The trachea and main bronchi are patent.
The lungs are clear without focal consolidation, mass, or pleural effusion.
A 4\,mm calcified granuloma is present in the right upper lobe.
The heart is normal in size; no pericardial effusion.
The thoracic aorta is of normal calibre with mild atherosclerotic calcifications.
Mediastinal lymph nodes do not exceed short-axis 8\,mm.
The visualised upper abdominal organs are unremarkable.
\end{minipage}}%
\hfill
\fbox{\begin{minipage}{0.49\textwidth}
\textbf{Per-organ sections (after LLM pipeline)} \\[2pt]
\rule{\linewidth}{0.4pt}
\small
\textbf{Thyroid:} \textit{The thyroid gland is normal in size and contours.} \quad{\scriptsize\textcolor{gray}{[normal\_thyroid]}} \\[2pt]
\textbf{Trachea:} \textit{The trachea and main bronchi are patent.} \quad{\scriptsize\textcolor{gray}{[normal\_airway]}} \\[2pt]
\textbf{Lungs:} \textit{The lungs are clear without focal consolidation, mass, or pleural effusion.} \quad{\scriptsize\textcolor{gray}{[normal\_lungs]}} \;
\textit{A 4\,mm calcified granuloma is present in the right upper lobe.} \quad{\scriptsize\textcolor{gray}{[granuloma]}} \\[2pt]
\textbf{Heart:} \textit{The heart is normal in size; no pericardial effusion.} \quad{\scriptsize\textcolor{gray}{[normal\_heart]}} \\[2pt]
\textbf{Aorta:} \textit{The thoracic aorta is of normal calibre with mild atherosclerotic calcifications.} \quad{\scriptsize\textcolor{gray}{[atherosclerosis]}} \\[2pt]
\textbf{Mediastinum:} \textit{Mediastinal lymph nodes do not exceed short-axis 8\,mm.} \quad{\scriptsize\textcolor{gray}{[normal\_lymph\_nodes]}} \\[2pt]
\textbf{Upper abdomen:} \textit{The visualised upper abdominal organs are unremarkable.} \quad{\scriptsize\textcolor{gray}{[normal\_upper\_abdomen]}}
\end{minipage}}
\caption{\textbf{Example of the LLM-based report-splitting pipeline.} A raw chest CT \emph{Findings} report (left, paraphrased from a public \ctrate{} study) is decomposed into atomic, single-sentence observations, each tagged with a \texttt{finding\_category} (in grey, in brackets) and an organ drawn from our taxonomy. Sentences sharing an anatomical entry are concatenated into one \emph{organ section} (right) before being embedded by the frozen text encoder.}
\label{fig:report_split_example}
\end{figure}

\section{Baseline Comparison}
\label{app:baseline_comparison}

\paragraph{Pretraining-data comparison.}
Table~\ref{tab:pretraining_data} contrasts the size of \ours{}'s pretraining corpus (74{,}434 chest and abdominal CT--report pairs from \ctrate{}, \inspect{}, and \stanford{}) with the corpora used by the baselines we compare against. Several baselines train on overlapping data: CT-CLIP and COLIPRI on \ctrate{}; Merlin on \stanford{}; and SPECTRE on \emph{all three} of our pretraining datasets (\ctrate{}, \inspect{}, \stanford{}) plus five additional public collections, for a total over $3\times$ the size of ours. CT-FM is trained on $148$k public CT scans from the Imaging Data Commons but without paired text; the remaining baselines (Pillar-0, CT-GLIP, TotalFM, fVLM) train on private corpora, also typically larger than ours. \ours{} therefore sits at roughly the combined scale of CT-CLIP and Merlin, well below the larger private and public corpora used by the rest.

\paragraph{Method comparison.}
Table~\ref{tab:method_comparison} contrasts the methodological choices of \ours{} against the baselines along four axes: whether training requires segmentation masks (\emph{Mask-Free}), whether the model uses a global image--text alignment loss (\emph{Glob.\ Align.}), whether it adds a local, region-level alignment (\emph{Loc.\ Align.}), and whether the pretraining corpus is publicly available (\emph{Pub.\ Data}). \ours{} is the only model that combines all four: it learns localized concept-level alignment without any segmentation supervision, retains a global CLIP-style objective on top, and trains exclusively on public data.

\input{tables/pretraining_data}
\input{tables/method_comparison}

\section{Training Hyperparameters}
\label{app:training_details}

We train on $8$ NVIDIA H100 GPUs with a per-GPU batch size of $6$, giving an effective contrastive batch of $48$ after all-gathering features across DDP ranks. Training runs for $120$ epochs of $1{,}000$ steps each ($120{,}000$ optimiser steps total) with AdamW (weight decay $0.05$, no gradient clipping) and a base learning rate of $1.5 \times 10^{-4}$ following a warmup--stable--decay schedule: $8$ epochs warmup, then a stable plateau, then $8$ epochs cooldown to a final learning rate of $10^{-6}$. Mixed-precision training (\texttt{bfloat16}) is used throughout. 

\section{Per-Finding Gain Breakdown}
\label{app:per_finding_breakdown}

To complement the macro-AUROC numbers reported in Table~\ref{tab:findings_classification_results}, we break down the gain from adding the per-organ tokens to the global \texttt{[CLS]} representation finding-by-finding. For each finding we report the linear-probing AUROC of \ours{}-Atlas evaluated with \texttt{[CLS]} alone and with the full $[\texttt{[CLS]};\, c_k]$ configuration, both averaged over five seeds, and the difference $\Delta$. Findings with no positive or negative example in the test split (and therefore an undefined per-class AUROC) are excluded; for the in-house \ccnabdo{} and \ccnchest{} taxonomies, this leaves $118/172$ and $71/80$ findings respectively. Tables~\ref{tab:per_finding_merlin}--\ref{tab:per_finding_ccn_chest} are sorted by $\Delta$ in decreasing order. The full configuration improves over the \texttt{[CLS]}-only baseline on \textbf{$87\%$} of \stanford{} findings, $\mathbf{83\%}$ of \ctrate{}, $\mathbf{70\%}$ of \ccnabdo{}, and $\mathbf{72\%}$ of \ccnchest{}; the largest gains are concentrated on focal, anatomically-localised findings (e.g., aortic aneurysm, lymphadenopathy, perihilar cholangiocarcinoma, esophagus solid mass), while the rare drops sit on diffuse or whole-volume findings (e.g., splenomegaly, pleural effusion, gynecomastia) where the global token already suffices.

\input{tables/per_finding_tables}

\input{tables/zero_shot_per_finding_tables}
\section{Zero-Shot Classification}
\label{app:zero_shot_templates}

\subsection{Short Zero-Shot Classification}
\label{app:zero_shot_short}

We evaluate zero-shot abnormality detection by aggregating 8 prompt pairs
(Table~\ref{tab:prompt_templates}) via mean cosine similarity across positive
and negative variations, as illustrated in Figure~\ref{fig:organ_anchor}.
Table~\ref{tab:findings_classification_results_zeroshot} reports zero-shot AUROC
across four splits. \ours{} leads on both abdomen splits ($78.08$ on \stanford{},
$63.57$ on \ccnabdo{}) and trails only COLIPRI on chest ($-3.1$ on \ctrate{},
$-4.0$ on \ccnchest{}), which is the only baseline pretrained with a
label/no-label contrastive objective specifically designed for zero-shot inference
(\textsuperscript{$\star$}). Concept queries consistently add signal
beyond \texttt{[CLS]} on abdomen (up to $+3.5$ on \stanford{}), mirroring the
linear-probing pattern observed in Section~\ref{subsec:findings_classification}.

\input{tables/findings_classification_results_zeroshot}

We further analyse prompt sensitivity in Figure~\ref{fig:zs_aggregation_template_ablation}, which
reports test AUROC as the number of aggregated prompt pairs $p$ grows from $1$ to $8$, with shaded bands
spanning the best- and worst-performing single-pair scores.
Aggregation converges quickly: all curves plateau by $p{\approx}4$-$5$, and for
the most sensitive encoders the gain is substantial: Pillar-0-Abd recovers
$+8.4$ and Merlin $+6.6$ on \stanford{} relative to their worst single pair,
swings larger than the gap between most methods in
Table~\ref{tab:findings_classification_results_zeroshot}.
\ours{} is the only encoder combining a narrow single-pair band ($\leq 4$ AUROC)
with a competitive aggregate, showing that robustness and accuracy are not at
odds. Merlin and Pillar-0 exhibit the widest bands (up to $22$-$24$ AUROC),
making aggregation essential for their fair evaluation. COLIPRI is the
limiting case: its aggregate degrades ($-0.6$) because its label/no-label
pretraining objective already aligns with the simplest pair in
Table~\ref{tab:prompt_templates}, a further indication that its strong chest
performance reflects overfitting to a specific inference setup rather than
prompt-agnostic generalisation.

\input{tables/zero_shot_templates}

\input{figures/zero_shot_template_ablation}

\subsection{Long Zero-Shot Classification}
\label{app:zero_shot_long}

 We evaluate a \emph{long} zero-shot variant: for each finding, we sample $r=50$ reports from patients with the abnormality and $r=50$ without. We then compute the embedding of each report and average each group to yield two representative prototypes ($p_k^{pos}$, $p_k^{neg}$). We use the cosine similarity as in short-form zero-shot between $[z_\texttt{[CLS]};\, c_k]$ and the two prototypes. The results are reported in Table~\ref{tab:findings_classification_results_zeroshot_long}.

Under long prompts, \ours{} is competitive across all four splits. With the Atlas backbone, \ours{} leads among external baselines on \ctrate{} ($79.05$ vs Pillar-0-Chest's $78.60$); on \stanford{} ($83.03$) it is a close second to Merlin ($84.26$), on \ccnabdo{} ($72.52$) it sits behind Pillar-0-Abd ($76.35$), and on \ccnchest{} ($79.40$) it trails SPECTRE ($80.18$). Switching to the ResNet-101 backbone, \ours{} achieves the top score on \ccnchest{} ($82.69$). Notably, some methods that demonstrate strong understanding under short prompts fail to maintain this performance on full reports, revealing a gap between short-form recognition and long-form comprehension. \oursmethod{} avoids this trade-off, with the Atlas variant competitive across modalities and the ResNet-101 variant leading on \ccnchest{}, confirming that the gains from $\mathcal{L}_\text{\oursmethod{}}$ carry across the two zero-shot text regimes.

\input{tables/findings_classification_results_zeroshot_long}

\section{Image-Text Retrieval}
\label{app:retrieval}

\input{tables/retrieval}

\paragraph{Encoding.}
For each model and evaluation set, we encode every CT volume with the vision encoder and the corresponding \emph{Findings} section with the text encoder. 

\paragraph{Scoring.}
We score each (image $i$, report $j$) pair by combining the cosine similarity of the global $[\texttt{CLS}]$ embeddings with the mean cosine similarity over the per-concept tokens active in report $j$:
\begin{equation}
    \text{sim}_{i,j} \;=\; \lambda_\text{cls}\,\text{sim}^{[\texttt{CLS}]}_{i,j} \;+\;
    (1-\lambda_\text{cls})\,\frac{1}{|\mathcal{K}_j|} \sum_{k \in \mathcal{K}_j}
    \text{sim}^{(k)}_{i,j},
    \label{eq:retrieval-score}
\end{equation}
where $\text{sim}^{[\texttt{CLS}]}_{i,j} = \langle \hat{z}^{\text{img}}_i,\, \hat{z}^{\text{txt}}_j \rangle$ and $\text{sim}^{(k)}_{i,j} = \langle \hat{c}^{\text{img}}_{i,k},\, \hat{c}^{\text{txt}}_{j,k} \rangle$ are cosine similarities of the L2-normalised global ($z$) and per-concept ($c_k$) embeddings (denoted by $\hat{\cdot}$), and $\mathcal{K}_j \subseteq \{1, \dots, K\}$ is the set of concept slots for which report $j$ contains at least one finding field). When $\mathcal{K}_j = \emptyset$, the second term is omitted, and the score reduces to the $[\texttt{CLS}]$ similarity. We report two settings, $\lambda_\text{cls} = 1$ (CLS only) and $\lambda_\text{cls} = 0.5$ (CLS and per-concept tokens equally weighted), and use the same $\lambda_\text{cls}$ in both retrieval directions (I$\rightarrow$T and T$\rightarrow$I). Baselines without concept-specific tokens are evaluated only at $\lambda_\text{cls} = 1$.

\paragraph{Protocol.}
We partition the test set into $\lfloor |\mathcal{D}|/n \rfloor$ disjoint subsets of $n = 100$ (image, report) pairs drawn from a single seeded random permutation; the $|\mathcal{D}| \bmod n$ remaining samples are discarded. Within each subset we rank candidates by Equation~\ref{eq:retrieval-score} (with $\lambda_\text{cls}=1$ for baselines) and evaluate both directions, image-to-text (I$\rightarrow$T) and text-to-image (T$\rightarrow$I). We report Recall@$\{1, 5, 10\}$ averaged first over the $n$ queries within each subset and then across subsets. 

\paragraph{Retrieval performance.}
Retrieval performances can be found in Table~\ref{tab:retrieval}. On chest (\ctrate), \ours{} (Atlas) reaches R@1 = 29.8 / 20.4 (I$\rightarrow$T / T$\rightarrow$I), trailing COLIPRI (33.3 / 38.2), which is explicitly tuned for global volume--report alignment via a report-generation head, and SPECTRE in the T$\rightarrow$I direction (25.3), while remaining ahead of every other CT VLM baseline in both directions. On abdomen (\stanford), \ours{} reaches R@1 = 43.7 / 27.8: ahead of CT-CLIP, fVLM, COLIPRI, and Pillar-0-Chest in both directions, and ahead of SPECTRE on I$\rightarrow$T (38.2); only the abdomen-specialized Merlin (64.0 / 65.3, trained with EHR supervision) and Pillar-0-Abd (48.9 / 47.7, abdomen-specific pretraining) lead in both directions, with SPECTRE pulling ahead on T$\rightarrow$I (38.8). 
The per-concept improves I$\rightarrow$T R@1 over the $[\texttt{CLS}]$-only variant across both regions and both backbones (Atlas: +4.4 chest / +5.6 abdomen; ResNet-101: +5.0 / +5.5) and brings a comparable T$\rightarrow$I gain on chest (+5.1 Atlas, +4.9 ResNet-101); the abdomen T$\rightarrow$I direction is roughly neutral. 

A likely contributing factor is that the text encoder is kept frozen during training, so the report-side representation cannot adapt to \ours{}'s per-concept structure---which plausibly caps T$\rightarrow$I performance overall. \ours{} is trained without any retrieval-specific objective.

\section{Radiology Report Generation}
\label{app:rrg_details}

\paragraph{Architecture.}
A single $576$-dimensional global feature pooled by \ours{} from the input volume (Section~\ref{subsec:organ_contrastive}) is projected through a 2-layer MLP into the input embedding space of a Qwen3.5-9B decoder~\cite{qwen35}. The vision encoder is initialised from the pretrained \ours{} checkpoint.

\paragraph{Stage 1 --- projector alignment.}
Both the vision encoder and the LLM are frozen, and only the MLP projector is updated. Reports are split into anatomically-grouped sections following the \stanford{} report structure. At each step we sample one organ section as the target with probability $0.8$, and the full report as the target with probability $0.2$. We train for $3$ epochs of $500$ steps with cosine LR decay from $10^{-3}$ to $10^{-5}$, effective batch size $24$ (per-GPU batch $1$, gradient accumulation $24$), bf16-mixed precision.

\paragraph{Stage 2 --- LoRA fine-tuning.}
Initialised from the Stage~1 checkpoint, the vision encoder remains frozen; the projector and LoRA adapters ($r{=}8$, applied to the $Q$ and $V$ projections of every transformer block of the LM) are jointly trained on full Findings sections. We train for $10$ epochs of $500$ steps, with $1$ epoch warmup and $2$ epochs cooldown, projector LR $5{\times}10^{-5}$, LoRA LR $2{\times}10^{-4}$. Table~\ref{tab:rrg_stage_ablation} shows the comparative results of Stage~2 LoRA tuning over the Stage~1 projector-only model.

\paragraph{Decoding and prompt.}
At inference, the LLM is prompted per organ section (e.g., ``\textit{Report the imaging findings in the liver and biliary tree for this abdominal CT scan.}'') and decodes each section autoregressively in its own pass with greedy sampling. The per-section outputs are concatenated into the final report following~\cite{merlin2024}. For Merlin, MedGemma-1.5, and Med3DVLM we use a matched template adapted to each model's interface according to their original implementation.

\paragraph{LLM-as-judge configuration.}
For GREEN~\cite{ostmeier2024green} we use the released open-weights judge; for CRIMSON~\cite{baharoon2026crimson} we use the GPT-5.2 API. CRIMSON ranges over $[-1, 1]$ with $0$ corresponding to a normal report; positive values reflect reports rated above the normal baseline and negative values clinically-significant errors.

\paragraph{Qualitative examples.}
Figure~\ref{fig:rrg-qualitative-stanford} contrasts \ours{} with the three baselines on two held-out \stanford{} studies, with matched findings highlighted in green and hallucinations in red.

\textit{Case~(a) --- mostly-normal abdominal CT.} The reference report contains three incidental findings (steatosis, post-cholecystectomy state, ventral-hernia mesh repair). \ours{} is the only model that correctly reports the dominant abnormality (steatosis) while keeping the rest of the report normal. Merlin defaults to all-normal and additionally fabricates a hysterectomy with bilateral salpingo-oophorectomy. MedGemma describes hip arthroplasty instead of the abdomen. Med3DVLM mistakes diffuse steatosis for a giant focal hepatic mass.

\textit{Case~(b) --- complex post-pelvic-exenteration study.} The reference contains many positive findings. \ours{} captures the dominant signal (right pleural effusion, end colostomy, abdominopelvic fluid, midline incision, post-surgical anatomy) while still missing the dominant left effusion, the ileal conduit, and the discrete rim-enhancing collections, and mis-identifies the operation type. Even so, \ours{} is the only model in clinical proximity to the GT: Merlin defaults to all-normal, MedGemma confabulates bilateral hemothoraces with metallic foreign-body injuries, and Med3DVLM produces a one-sentence non-answer.

Together, the two cases illustrate (i)~\ours{}'s clear advantage over openly-available baselines on both regimes, and (ii)~the persistent under-reporting of positive findings, common to all evaluated models in 3D abdominal CT report generation.

\input{tables/rrg_stage_ablation}
\input{figures/rrg_examples}

\section{PCA Visualization of Learned Feature Maps}
\label{app:pca_visualization}

To qualitatively assess the spatial structure of learned representations, we visualize the feature maps of each model using Principal Component Analysis (PCA). The PCA is computed directly from the high-resolution feature map used to derive the \texttt{[CLS]} token; the first principal component is discarded, as it is heavily linked to the background, and the next three components are mapped to the RGB channels of the visualization. Each color thus reflects a distinct mode of variation in the feature space, and regions sharing similar colors exhibit similar learned representations. As shown in Figure~\ref{fig:pca}, models differ substantially in the spatial richness of their embeddings: Merlin produces maps with little spatial differentiation, while Pillar-0 and CLIP baselines show coarser but more structured activations following anatomical boundaries. \ours{} models produce the most anatomically coherent maps, with distinct color assignments for individual organs: for instance, the heart, bones, and liver are clearly separable.

\input{figures/pca}

\section{Interpretability: Attention-Map Failure Cases}
\label{app:interpretability_failures}

While the per-concept attention maps of Figure~\ref{fig:attention_maps} concentrate cleanly on the targeted anatomy for most concepts, a few queries are less precisely localised. Figure~\ref{fig:attention_maps_fail} shows representative cases for the pancreas and spleen, where attention spreads to neighbouring organs. We attribute this to the small size and adjacency of these structures and to the higher rate of normal mentions in their associated report sections, which weakens the contrastive signal each query receives.

\input{figures/attention_map_fail}

\section{Broader Impacts}
\label{app:broader_impacts}

Foundation models for 3D CT can broaden access to diagnostic support, accelerate radiology workflows, and standardize finding extraction across centers. They also carry risks: automation bias (over-reliance on model output), disparate performance across demographic and acquisition-protocol subgroups not represented in training data, and miscalibration when deployed outside the centers and populations seen at pretraining. We release \ours{} for research use only and do not advocate clinical deployment without prospective, multi-center validation. The LLM-based report-structuring pipeline introduces an additional dependency whose biases should be audited before any clinical reuse.

%

%% file: tables/datasets.tex
\begin{table}[h]
    \centering
    \small
    \setlength{\tabcolsep}{6pt}
    \caption{\textbf{Dataset statistics.} CT volumes per dataset, train / test split, and finding taxonomy size. \textsuperscript{$\star$}Private out-of-distribution evaluation set from a different clinical center. \# Findings: native taxonomies for \ctrate{} ($18$) and \stanford{} ($30$); our in-house taxonomy of $80$ chest / $172$ abdominal findings for \inspect{} and the OOD sets.}
    \label{tab:datasets}
    \begin{tabular}{l l r r r}
        \toprule
        \textbf{Domain} & \textbf{Dataset} & \textbf{\# Train} & \textbf{\# Test} & \textbf{\# Findings} \\
        \midrule
        \multirow{3}{*}{Chest CT}
            & \ctrate~\cite{ctrate2024}         & 24{,}128 & 1{,}564  & 18 \\
            & \inspect~\cite{inspect2023}       & 23{,}248 & --       & 80 \\
            & \ccnchest\textsuperscript{$\star$} & 30{,}873 & 3{,}851  & 80 \\
        \midrule
        \multirow{2}{*}{Abdominal CT}
            & \stanford~\cite{merlin2024}       & 20{,}357 & 5{,}137  & 30 \\
            & \ccnabdo\textsuperscript{$\star$}  & 6{,}503  & 811      & 172 \\
        \bottomrule
    \end{tabular}
\end{table}

%% file: tables/anatomical_taxonomy.tex
\begin{table}[h]
\centering\scriptsize\setlength{\tabcolsep}{6pt}
\renewcommand{\arraystretch}{0.95}
\caption{The $K = 102$ anatomical concepts used as \oursmethod{} cross-attention queries (Section~\ref{subsec:organ_contrastive}). Listed in alphabetical order.}
\label{tab:anatomical_taxonomy}
\begin{tabular}{l l l l l}
\toprule
Abdominal cavity & Common carotid artery & Inguinal hernia & Portal vein & Spleen \\
Abdominal wall & Device & Joint & Postsurgical changes & Splenic artery \\
Adrenal glands & Diaphragm & Kidneys & Prostate & Splenic vein \\
Anus & Duodenum & Large bowel & Pulmonary artery & Sternum \\
Aorta & Epididymis & Liver & Pulmonary vein & Stomach \\
Appendix & Esophagus & Lungs & Rectum & Subclavian artery \\
Atrial appendage & Fallopian tube & Lymph nodes & Renal artery & Subclavian vein \\
Biliary & Femurs & Mediastinum & Renal vein & Superior mesenteric artery \\
Bladder & Gallbladder & Mesenteric & Retroperitoneum & Superior mesenteric vein \\
Bone & Gastrointestinal tract & Muscle & Ribs & Superior vena cava \\
Brachiocephalic trunk & Gonadal vein & Orbit & Sacrum & Testis \\
Brachiocephalic vein & Heart & Ovarian artery & Scapulae & Thyroid \\
Brain & Hepatic artery & Ovarian vein & Scrotum & Total organs \\
Breast & Hepatic vein & Ovaries & Seminal vesicle & Trachea \\
Chest & Hernia & Pancreas & Sinus & Ureter \\
Chest soft tissue & Hip & Pelvis & Skin & Urinary \\
Chest soft tissue solid mass & Iliac artery & Penis & Skull & Uterus \\
Chest soft tissues & Iliac vein & Pericardium & Small bowel & Vagina \\
Chest wall & Inferior mesenteric artery & Peritoneum & Soft tissues &  \\
Clavicles & Inferior mesenteric vein & Pleura & Spinal cord &  \\
Colon & Inferior vena cava & Portal artery & Spine &  \\
\bottomrule
\end{tabular}
\end{table}

%% file: tables/pretraining_data.tex
\begin{table}[t]
    \centering
    \caption{\textbf{Pretraining data comparison.} CT volumes and paired radiology reports used by each baseline foundation model and by \ours{}. The last column lists which of our three pretraining datasets (\ctrate{}, \inspect{}, \stanford{}) the baseline also trains on.}
    \label{tab:pretraining_data}
    \resizebox{\textwidth}{!}{%
    \begin{tabular}{l l r r l}
        \toprule
        \textbf{Model} & \textbf{Domains} & \textbf{\# CT scans} & \textbf{\# Reports} & \textbf{Shared with our pretraining} \\
        \midrule
        CT-CLIP~\cite{ctclip2024}                  & chest               & 25{,}692            & 25{,}692            & \ctrate{} \\
        Merlin~\cite{merlin2024}                   & abdomen             & 25{,}494            & 25{,}494\,$^*$      & \stanford{} \\
        Pillar-0-Chest~\cite{pillar0_2025}         & chest               & 86{,}411            & 86{,}411            & --- (private corpus) \\
        Pillar-0-Abd~\cite{pillar0_2025}           & abdomen-pelvis      & 42{,}990            & 42{,}990            & --- (private corpus) \\
        CT-FM~\cite{ctfm2025}                      & multi-domain        & 148{,}000           & 0                   & --- \\
        SPECTRE~\cite{spectre2025}                 & multi-domain        & 229{,}619           & 85{,}689            & \ctrate{}, \inspect{}, \stanford{} \\
        COLIPRI~\cite{colipri2025}                 & chest               & 97{,}700            & 25{,}692            & \ctrate{} \\
        fVLM~\cite{fvlm2025}                       & chest               & 272{,}124           & 272{,}124           & --- (private) \\
        CT-GLIP~\cite{ctglip2024}                  & full-body           & 44{,}011            & 44{,}011            & --- (private) \\
        TotalFM~\cite{totalfm2026}                 & full-body           & 140{,}000           & 140{,}000           & --- (private) \\
        \midrule
        \textbf{\ours{} \textit{(Ours)}}           & chest + abdomen     & \textbf{74{,}434}   & \textbf{74{,}434}   & (\ctrate{}, \inspect{}, \stanford{}) \\
        \bottomrule
    \end{tabular}
    }
    \vspace{2pt}
    \\ {\scriptsize $^*$Merlin additionally pairs each scan with structured EHR data.}
\end{table}

%% file: tables/method_comparison.tex
\begin{table}[t]
    \centering
    \caption{\textbf{Method comparison.} Methodological choices of \ours{} against baselines along four axes: \emph{Mask-Free} (training does not require segmentation masks), \emph{Glob.\ Align.} (uses a global image--text contrastive loss), \emph{Loc.\ Align.} (adds a local, region-level alignment), and \emph{Pub.\ Data} (pretraining corpus is publicly available).}
    \label{tab:method_comparison}
    \begin{tabular}{l c c c c}
        \toprule
        \textbf{Model} & \textbf{Mask-Free} & \textbf{Glob.\ Align.} & \textbf{Loc.\ Align.} & \textbf{Pub.\ Data} \\
        \midrule
        CT-CLIP~\cite{ctclip2024}              & \cmark & \cmark & \xmark & \cmark \\
        Merlin~\cite{merlin2024}               & \cmark & \cmark & \xmark & \cmark \\
        Pillar-0-Chest~\cite{pillar0_2025}     & \cmark & \cmark & \xmark & \xmark \\
        Pillar-0-Abd~\cite{pillar0_2025}       & \cmark & \cmark & \xmark & \xmark \\
        CT-FM~\cite{ctfm2025}                  & \cmark & \xmark & \xmark & \cmark \\
        SPECTRE~\cite{spectre2025}             & \cmark & \cmark & \xmark & \cmark \\
        COLIPRI~\cite{colipri2025}             & \cmark & \cmark & \xmark & \cmark \\
        fVLM~\cite{fvlm2025}                   & \xmark & \xmark & \cmark & \xmark \\
        CT-GLIP~\cite{ctglip2024}              & \xmark & \xmark & \cmark & \xmark \\
        TotalFM~\cite{totalfm2026}             & \xmark & \xmark & \cmark & \xmark \\
        \midrule
        \textbf{\ours{} \textit{(Ours)}}       & \cmark & \cmark & \cmark & \cmark \\
        \bottomrule
    \end{tabular}
\end{table}

%% file: tables/per_finding_tables.tex
\begin{table}[h]
\centering\footnotesize\setlength{\tabcolsep}{4pt}
\caption{\textbf{Per-finding linear-probing AUROC on \stanford{} (Merlin-binary, 30 findings).} Comparison of \ours{}-Atlas evaluated with the global \texttt{[CLS]} token alone vs.\ the full configuration ($\texttt{[CLS]}+c_k$). Sorted by $\Delta = \text{Full} - \text{[CLS]}$.}
\label{tab:per_finding_merlin}
\begin{tabular}{l r r r|l r r r}
\toprule
\textbf{Finding} & \textbf{[CLS]} & \textbf{Full} & \textbf{$\Delta$} & \textbf{Finding} & \textbf{[CLS]} & \textbf{Full} & \textbf{$\Delta$} \\
\midrule
Lymphadenopathy & 70.53 & 75.79 & \textcolor{teal}{+5.26} & Prostatomegaly & 89.06 & 89.66 & \textcolor{teal}{+0.59} \\
Renal hypodensities & 63.10 & 66.15 & \textcolor{teal}{+3.05} & Hepatomegaly & 82.57 & 83.12 & \textcolor{teal}{+0.55} \\
Abdominal aortic aneurysm & 85.10 & 87.90 & \textcolor{teal}{+2.80} & Coronary calcification & 78.82 & 79.31 & \textcolor{teal}{+0.49} \\
Metastatic disease & 85.08 & 87.38 & \textcolor{teal}{+2.30} & Biliary ductal dilation & 81.66 & 82.00 & \textcolor{teal}{+0.34} \\
Pancreatic atrophy & 76.08 & 78.32 & \textcolor{teal}{+2.25} & Bowel obstruction & 92.73 & 93.06 & \textcolor{teal}{+0.33} \\
Submucosal edema & 74.91 & 76.45 & \textcolor{teal}{+1.54} & Fracture & 81.57 & 81.85 & \textcolor{teal}{+0.28} \\
Aortic valve calcification & 87.73 & 89.22 & \textcolor{teal}{+1.49} & Hydronephrosis & 89.49 & 89.64 & \textcolor{teal}{+0.15} \\
Thrombosis & 79.50 & 80.90 & \textcolor{teal}{+1.40} & Anasarca & 95.22 & 95.37 & \textcolor{teal}{+0.15} \\
Hiatal hernia & 75.35 & 76.48 & \textcolor{teal}{+1.14} & Free air & 84.79 & 84.93 & \textcolor{teal}{+0.15} \\
Osteopenia & 89.23 & 90.12 & \textcolor{teal}{+0.88} & Appendicitis & 88.73 & 88.82 & \textcolor{teal}{+0.09} \\
Renal cyst & 70.87 & 71.67 & \textcolor{teal}{+0.80} & Pleural effusion & 92.89 & 92.89 & \textcolor{teal}{+0.00} \\
Atelectasis & 63.97 & 64.74 & \textcolor{teal}{+0.77} & Surgically absent gallbladder & 96.21 & 96.01 & \textcolor{red}{-0.20} \\
Ascites & 89.94 & 90.68 & \textcolor{teal}{+0.74} & Hepatic steatosis & 89.13 & 88.83 & \textcolor{red}{-0.30} \\
Cardiomegaly & 83.04 & 83.77 & \textcolor{teal}{+0.73} & Gallstones & 72.22 & 71.63 & \textcolor{red}{-0.59} \\
Atherosclerosis & 81.01 & 81.64 & \textcolor{teal}{+0.62} & Splenomegaly & 90.21 & 89.39 & \textcolor{red}{-0.82} \\
\bottomrule
\end{tabular}
\end{table}

\begin{table}[h]
\centering\small\setlength{\tabcolsep}{4pt}
\caption{\textbf{Per-finding linear-probing AUROC on \ctrate{} (binary taxonomy, 18 findings).} Same setup as Table~\ref{tab:per_finding_merlin}.}
\label{tab:per_finding_ctrate}
\begin{tabular}{l r r r|l r r r}
\toprule
\textbf{Finding} & \textbf{[CLS]} & \textbf{Full} & \textbf{$\Delta$} & \textbf{Finding} & \textbf{[CLS]} & \textbf{Full} & \textbf{$\Delta$} \\
\midrule
Lymphadenopathy & 77.50 & 78.22 & \textcolor{teal}{+0.72} & Emphysema & 80.38 & 80.58 & \textcolor{teal}{+0.20} \\
Pulmonary fibrotic sequela & 73.15 & 73.83 & \textcolor{teal}{+0.68} & Mosaic attenuation pattern & 90.33 & 90.50 & \textcolor{teal}{+0.17} \\
Pericardial effusion & 90.67 & 91.22 & \textcolor{teal}{+0.55} & Arterial wall calcification & 94.17 & 94.31 & \textcolor{teal}{+0.14} \\
Interlobular septal thickening & 88.66 & 89.09 & \textcolor{teal}{+0.43} & Coronary artery wall calcification & 93.76 & 93.89 & \textcolor{teal}{+0.13} \\
Medical material & 92.97 & 93.35 & \textcolor{teal}{+0.38} & Bronchiectasis & 80.16 & 80.27 & \textcolor{teal}{+0.11} \\
Lung opacity & 87.73 & 88.08 & \textcolor{teal}{+0.35} & Atelectasis & 80.86 & 80.96 & \textcolor{teal}{+0.10} \\
Lung nodule & 72.74 & 73.01 & \textcolor{teal}{+0.28} & Peribronchial thickening & 81.99 & 81.85 & \textcolor{red}{-0.14} \\
Consolidation & 92.70 & 92.94 & \textcolor{teal}{+0.24} & Pleural effusion & 97.91 & 97.64 & \textcolor{red}{-0.27} \\
Cardiomegaly & 93.71 & 93.94 & \textcolor{teal}{+0.24} & Hiatal hernia & 82.83 & 82.25 & \textcolor{red}{-0.58} \\
\bottomrule
\end{tabular}
\end{table}

\begin{table}[h]
\centering\small\setlength{\tabcolsep}{4pt}
\caption{\textbf{Per-finding linear-probing AUROC on \ccnabdo{} (in-house taxonomy, 172 findings).} 15 largest gains and 5 largest drops out of 118 findings with computable AUC ($\Delta$ mean over all 118: +1.84 AUROC; $\Delta>0$ on 70.3\% of findings).}
\label{tab:per_finding_ccn_abd}
\begin{tabular}{l r r r}
\toprule
\textbf{Finding} & \textbf{[CLS]} & \textbf{Full} & \textbf{$\Delta$} \\
\midrule
Esophagus solid mass & 56.26 & 84.50 & \textcolor{teal}{+28.24} \\
Kidneys laceration & 17.47 & 43.19 & \textcolor{teal}{+25.71} \\
Perihilar cholangiocarcinoma & 59.52 & 83.11 & \textcolor{teal}{+23.59} \\
Small bowel solid mass & 61.14 & 75.15 & \textcolor{teal}{+14.01} \\
Colitis & 69.66 & 82.54 & \textcolor{teal}{+12.88} \\
Gastric wall thickening & 46.31 & 59.04 & \textcolor{teal}{+12.72} \\
Prostatic solid mass & 16.36 & 29.05 & \textcolor{teal}{+12.69} \\
Kidneys angiomyolipoma & 51.84 & 63.08 & \textcolor{teal}{+11.24} \\
Peritoneum carcinomatosis & 74.15 & 84.35 & \textcolor{teal}{+10.20} \\
Tubo ovarian abscess & 81.94 & 91.68 & \textcolor{teal}{+9.75} \\
Kidneys arterial stenosis & 69.50 & 78.71 & \textcolor{teal}{+9.22} \\
Iliopsoas abscess & 89.51 & 98.63 & \textcolor{teal}{+9.13} \\
Liver laceration & 21.52 & 29.56 & \textcolor{teal}{+8.04} \\
Pyosalpinx & 79.57 & 86.57 & \textcolor{teal}{+7.00} \\
Liver focal nodular hyperplasia & 85.16 & 91.23 & \textcolor{teal}{+6.07} \\
\midrule
\multicolumn{4}{c}{\textit{... \(98\) intermediate findings omitted ...}} \\
\midrule
Status post appendectomy & 38.20 & 29.96 & \textcolor{red}{-8.24} \\
Extrahepatic bile duct mass & 92.90 & 83.08 & \textcolor{red}{-9.81} \\
Nephrocalcinosis & 33.86 & 23.46 & \textcolor{red}{-10.39} \\
Mesenteric lymphadenopathy & 63.50 & 49.51 & \textcolor{red}{-13.99} \\
Aortic stent & 79.29 & 59.06 & \textcolor{red}{-20.23} \\
\bottomrule
\end{tabular}
\end{table}

\begin{table}[h]
\centering\small\setlength{\tabcolsep}{4pt}
\caption{\textbf{Per-finding linear-probing AUROC on \ccnchest{} (in-house taxonomy, 80 findings).} 15 largest gains and 5 largest drops out of 71 findings with computable AUC ($\Delta$ mean over all 71: +1.08 AUROC; $\Delta>0$ on 71.8\% of findings).}
\label{tab:per_finding_ccn_chest}
\begin{tabular}{l r r r}
\toprule
\textbf{Finding} & \textbf{[CLS]} & \textbf{Full} & \textbf{$\Delta$} \\
\midrule
Hematogenic micronodules & 19.57 & 51.67 & \textcolor{teal}{+32.10} \\
Pneumatocele & 64.69 & 82.15 & \textcolor{teal}{+17.46} \\
Diaphragmatic elevation & 78.76 & 84.49 & \textcolor{teal}{+5.73} \\
Mediastinal solid mass & 88.04 & 92.61 & \textcolor{teal}{+4.57} \\
Chest soft tissue solid mass & 87.09 & 91.00 & \textcolor{teal}{+3.91} \\
Goiter & 83.28 & 87.05 & \textcolor{teal}{+3.77} \\
Anomalous venous return & 68.75 & 72.22 & \textcolor{teal}{+3.47} \\
Lungs fissural distortion & 81.84 & 85.31 & \textcolor{teal}{+3.47} \\
Rib fracture & 74.83 & 77.44 & \textcolor{teal}{+2.62} \\
Aortic tube replacement & 90.15 & 92.72 & \textcolor{teal}{+2.57} \\
Pneumomediastinum & 95.46 & 97.41 & \textcolor{teal}{+1.95} \\
Pleural plaques & 88.07 & 89.79 & \textcolor{teal}{+1.72} \\
Thyroid nodule & 83.84 & 85.37 & \textcolor{teal}{+1.54} \\
Tracheal solid mass & 72.38 & 73.90 & \textcolor{teal}{+1.52} \\
Lobar pneumonia & 91.08 & 92.56 & \textcolor{teal}{+1.48} \\
\midrule
\multicolumn{4}{c}{\textit{... \(51\) intermediate findings omitted ...}} \\
\midrule
Bronchiolitis & 84.59 & 83.22 & \textcolor{red}{-1.37} \\
Lungs cyst & 91.28 & 88.22 & \textcolor{red}{-3.05} \\
Gynecomastia & 86.73 & 81.77 & \textcolor{red}{-4.96} \\
Diaphragmatic hernia & 76.01 & 69.79 & \textcolor{red}{-6.22} \\
Perilymphatic micronodules & 78.97 & 71.94 & \textcolor{red}{-7.03} \\
\bottomrule
\end{tabular}
\end{table}

%% file: tables/findings_classification_results_zeroshot.tex
\begin{table}[t]
    \centering
    \caption{\textbf{Short zero-shot finding classification results} (\% AUROC). \ccnchest{} and \ccnabdo{} are unseen during training for all models and use our in-house taxonomy of 172 abdomen and 80 chest findings; \stanford{} and \ctrate{} use their original taxonomies. Long zero-shot results, where class prototypes are built from real reports rather than templated prompts, are reported in Appendix~\ref{app:zero_shot_long}. \textsuperscript{$\star$}Models trained with zero-shot-specific text augmentation during pretraining (e.g., LLM-shortened statements or templated prompts mimicking the inference distribution).}
    \label{tab:findings_classification_results_zeroshot}
    \small
    \setlength{\tabcolsep}{4pt}
    \renewcommand{\arraystretch}{0.92}
    \begin{tabular}{l c c c c}
        \toprule
        & \multicolumn{2}{c}{\textbf{Abdomen}}
        & \multicolumn{2}{c}{\textbf{Chest}} \\
        \cmidrule(lr){2-3} \cmidrule(lr){4-5}
        \textbf{Model}
        & \stanford{} & \ccnabdo{}
        & \ctrate{} & \ccnchest{} \\
        \midrule
        CT-CLIP~\cite{ctclip2024}                       & \textcolor{gray}{60.70} & \textcolor{gray}{52.00} & 71.07 & 57.90 \\
        COLIPRI\textsuperscript{$\star$}~\cite{colipri2025} & \textcolor{gray}{71.65} & \textcolor{gray}{62.31} & \textbf{76.98} & \textbf{73.34} \\
        SPECTRE~\cite{spectre2025}                      & 64.20 & 60.90 & 57.00 & 59.70 \\
        Merlin~\cite{merlin2024}                        & 77.72 & 63.46 & \textcolor{gray}{65.02} & \textcolor{gray}{56.85} \\
        Pillar-0-Abd~\cite{pillar0_2025}                & 70.95 & 63.54 & \textcolor{gray}{64.04} & \textcolor{gray}{63.12} \\
        Pillar-0-Chest~\cite{pillar0_2025}              & \textcolor{gray}{67.36} & \textcolor{gray}{56.34} & 70.86 & 47.22 \\
        \midrule
        CLIP Baseline                                   & 71.07 & 60.40 & 71.38 & 72.85 \\
        \ours{} - [CLS]                                 & 74.58 & \textbf{63.57} & 73.46 & 71.81 \\
        \textbf{\ours{}}                                & \textbf{78.04} & 62.06 & 74.18 & 69.34 \\
        \bottomrule
    \end{tabular}
\end{table}

%% file: tables/zero_shot_templates.tex
\begin{table}[h!]
    \centering
    \small 
    \caption{Short-form prompt templates used for zero-shot abnormality detection. During evaluation, \texttt{[label]} is replaced with the specific clinical finding.}
    \label{tab:prompt_templates}
    \begin{tabular}{@{} l p{0.45\textwidth} p{0.45\textwidth} @{}}
        \toprule
        \textbf{\#} & \textbf{Positive Template} & \textbf{Negative Template} \\
        \midrule
        1 & \texttt{[label]} & \texttt{no [label]} \\
        2 & \texttt{there is evidence of [label]} & \texttt{there is no evidence of [label]} \\
        3 & \texttt{[label] present} & \texttt{[label] not present} \\
        4 & \texttt{findings consistent with [label]} & \texttt{no findings consistent with [label]} \\
        5 & \texttt{The CT scan shows [label]} & \texttt{The CT scan does not show [label]} \\
        6 & \texttt{a CT showing [label]} & \texttt{a CT without [label]} \\
        7 & \texttt{Impression: [label]} & \texttt{Impression: no [label]} \\
        8 & \texttt{this is an image of a [label]} & \texttt{this is an image with no [label]} \\
        \bottomrule
    \end{tabular}
\end{table}

%% file: figures/zero_shot_template_ablation.tex
\begin{figure}[!h]
    \centering
    \begin{minipage}{0.49\textwidth}
        \centering
        \includegraphics[width=\linewidth]{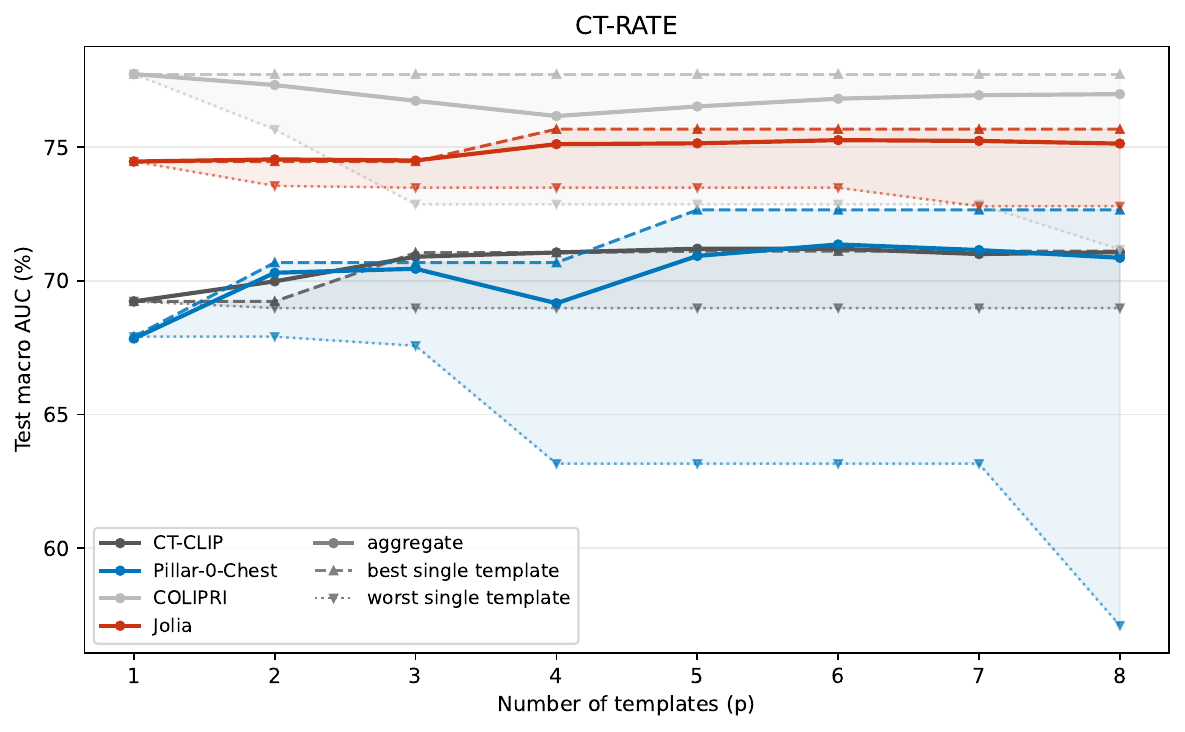}
    \end{minipage}\hfill
    \begin{minipage}{0.49\textwidth}
        \centering
        \includegraphics[width=\linewidth]{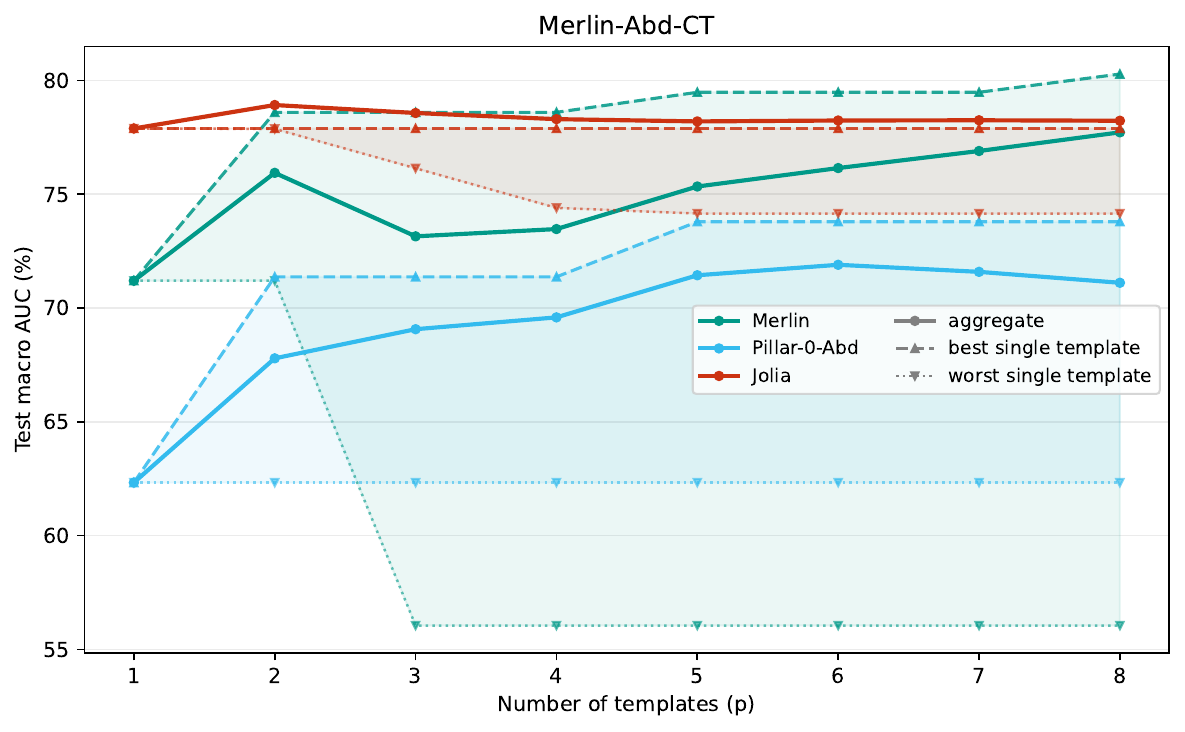}
    \end{minipage}
    
    \caption{\textbf{Aggregation converges to a stable zero-shot estimate.}
Test macro-AUC against the number of aggregated templates $p$ on CT-RATE
(left) and Merlin-Abd-CT (right). Solid lines show the aggregation protocol;
dashed and dotted lines show the val-best and val-worst single-template AUCs,
with the shaded tube denoting their envelope. Curves stabilise by
$p \approx 4\text{–}5$ on both datasets and no model collapses at $p=1$,
supporting aggregation over the full template pool as a model-agnostic
zero-shot protocol.}
    \label{fig:zs_aggregation_template_ablation}
\end{figure}

%% file: tables/findings_classification_results_zeroshot_long.tex
\begin{table}[htbp]
    \centering
    \caption{\textbf{Long zero-shot classification results} (\% AUROC). Class prototypes are built from the embeddings of $50$ positive and $50$ negative real reports per finding (Appendix~\ref{app:zero_shot_long}). \ccnchest{} and \ccnabdo{} are unseen during training for all models. \textsuperscript{$\star$}Models trained with zero-shot-specific text augmentation during pretraining.}
    \label{tab:findings_classification_results_zeroshot_long}
    \resizebox{\textwidth}{!}{%
    \setlength{\tabcolsep}{5pt}
    \begin{tabular}{l c c c c}
        \toprule
        & \multicolumn{2}{c}{\textbf{Abdomen}}
        & \multicolumn{2}{c}{\textbf{Chest}} \\
        \cmidrule(lr){2-3} \cmidrule(lr){4-5}
        \textbf{Model}
        & \stanford{} & \ccnabdo{}
        & \ctrate{} & \ccnchest{} \\
        \midrule
        CT-CLIP~\cite{ctclip2024}                       & 60.37 & 53.74 & 74.22 & 70.31 \\
        fVLM~\cite{fvlm2025}                            & 57.19    & 55.62    & 68.45    & 54.20    \\
        COLIPRI\textsuperscript{$\star$}~\cite{colipri2025} & 74.40 & 62.22 & 78.57 & 71.72 \\
        SPECTRE~\cite{spectre2025}                      & 79.70    & 71.27    & 72.52    & 80.18    \\
        Merlin~\cite{merlin2024}                        & \textbf{84.26} & 68.90 & 70.26 & 60.48 \\
        Pillar-0-Abd~\cite{pillar0_2025}                & 81.58 & \textbf{76.35} & 66.80 & 78.78 \\
        Pillar-0-Chest~\cite{pillar0_2025}              & 71.91 & 61.85 & 78.60 & 79.12 \\
        \midrule
        \textit{Atlas architecture} \\
        CLIP Baseline                                   & 82.78 & 72.80 & \textbf{80.33} & 79.41 \\
        \ours{} - [CLS]                                 & 81.99 & 73.58 & 79.40 & 80.70 \\
        \textbf{\ours{}}                                & 83.03 & 72.52 & 79.05 & 79.40 \\
        \midrule
        \textit{ResNet-101 architecture} \\
        CLIP Baseline                                   & 82.58 & 71.60 & 80.26 & 72.26 \\
        \ours{} - [CLS]                                 & 81.96 & 73.33 & 79.50 & 80.21 \\
        \textbf{\ours{}}                                & 83.49 & 74.28 & 80.22 & \textbf{82.69} \\
        \bottomrule
    \end{tabular}
    }
\end{table}

%% file: tables/retrieval.tex
\begin{table}[t]

    \centering
    \caption{\textbf{Image-text retrieval results.} Recall@1, @5, @10 for image-to-text (I$\rightarrow$T) and text-to-image (T$\rightarrow$I). Averages from the full dataset, based on subsets of 100 samples.}
    \label{tab:retrieval}
    \resizebox{\textwidth}{!}{%
    \setlength{\tabcolsep}{5pt}
    \begin{tabular}{l ccc ccc ccc ccc}
        \toprule
        & \multicolumn{6}{c}{\textbf{Chest (\ctrate)}} & \multicolumn{6}{c}{\textbf{Abdomen (\stanford)}} \\
        \cmidrule(lr){2-7} \cmidrule(lr){8-13}
        & \multicolumn{3}{c}{I$\rightarrow$T} & \multicolumn{3}{c}{T$\rightarrow$I} & \multicolumn{3}{c}{I$\rightarrow$T} & \multicolumn{3}{c}{T$\rightarrow$I} \\
        \cmidrule(lr){2-4} \cmidrule(lr){5-7} \cmidrule(lr){8-10} \cmidrule(lr){11-13}
        \textbf{Model} & @1 & @5 & @10 & @1 & @5 & @10 & @1 & @5 & @10 & @1 & @5 & @10 \\
        \midrule
            CT-CLIP~\cite{ctclip2024} & 4.71 & 18.50 & 30.21 & 9.07 & 28.43 & 44.14 & 0.80 & 5.12 & 10.61 & 1.43 & 6.53 & 12.39 \\
            fVLM~\cite{fvlm2025} & 1.86 & 11.50 & 20.21 & 1.64 & 7.00 & 13.71 & 1.31 & 5.73 & 11.16 & 1.39 & 6.59 & 12.12 \\
            COLIPRI~\cite{colipri2025} & 33.29 & 63.71 & 76.36 & \textbf{38.21} & \textbf{71.43} & \textbf{82.21} & 4.92 & 18.78 & 31.43 & 9.10 & 28.29 & 42.76 \\
            SPECTRE~\cite{spectre2025} & 25.07 & 54.93 & 67.93 & 25.29 & 56.43 & 68.79 & 38.18 & 70.80 & 82.53 & 38.84 & 72.47 & 84.20 \\
            Merlin~\cite{merlin2024} & 4.50 & 14.43 & 26.00 & 5.71 & 16.86 & 26.86 & \textbf{64.00} & \textbf{90.06} & \textbf{95.47} & \textbf{65.25} & \textbf{91.08} & \textbf{95.88} \\
            Pillar-0-Abd~\cite{pillar0_2025} & 2.64 & 10.07 & 17.29 & 2.50 & 8.86 & 16.71 & 48.90 & 79.33 & 89.33 & 47.69 & 76.82 & 87.18 \\
            Pillar-0-Chest~\cite{pillar0_2025} & 12.07 & 32.71 & 47.21 & 10.43 & 29.43 & 42.50 & 3.71 & 13.94 & 24.82 & 4.96 & 17.80 & 28.69 \\
        \midrule
        \textit{Atlas architecture} \\
            CLIP Baseline & \textbf{35.29} & \textbf{67.21} & \textbf{81.21} & 26.14 & 53.43 & 66.07 & 41.39 & 77.55 & 88.96 & 34.06 & 69.04 & 82.33\\
            \ours{} - [CLS] & 25.36 & 55.79 & 71.07 & 15.36 & 34.86 & 48.64 & 38.06 & 73.49 & 85.88 & 28.29 & 60.51 & 75.80 \\
            \textbf{\ours{}} & 29.79 & 61.29 & 74.43 & 20.43 & 45.57 & 59.79 & 43.69 & 78.61 & 89.98 & 27.75 & 58.37 & 73.65 \\
        \midrule
        \textit{ResNet-101 architecture} \\
            CLIP Baseline & 33.07 & 64.14 & 78.21 & 20.43 & 48.93 & 60.79 & 39.14 & 77.86 & 89.33 &35.92 & 72.63 & 84.86 \\
            \ours{} - [CLS] & 21.93 & 50.29 & 66.00 & 14.36 & 33.86 & 47.00 & 34.80 & 71.88 & 85.31 & 31.75 & 65.80 & 79.24 \\
            \textbf{\ours{}} & 26.93 & 56.00 & 70.57 & 19.29 & 42.71 & 56.50 & 40.25 & 76.49 & 88.51 & 31.90 & 64.76 & 78.14 \\
        \bottomrule
    \end{tabular}
    }
\end{table}

%% file: tables/rrg_stage_ablation.tex
\begin{table}[h]
    \centering
    \caption{\textbf{Stage 1 vs. Stage~2 LoRA fine-tuning} for \ours{} on \stanford{}.}
    \label{tab:rrg_stage_ablation}
    \begin{tabular}{l ccccc}
        \toprule
        \textbf{Stage} & BLEU & ROUGE-L & BERTScore & RadGraph-F1 & CRIMSON \\
        \midrule
        Stage 1 (projector only) & 0.029 & 0.247 & 0.502 & 0.244 & $-0.010$ \\
        Stage 2 (+ LoRA)         & 0.119 & 0.323 & 0.567 & 0.317 & $-0.194$ \\
        \bottomrule
    \end{tabular}
\end{table}

%% file: figures/rrg_examples.tex
\definecolor{RrgMatch}{rgb}{0.06, 0.50, 0.13}    
\definecolor{RrgHall}{rgb}{0.75, 0.10, 0.10}     
\definecolor{RrgGTRow}{rgb}{0.94, 0.94, 0.94}    
\definecolor{RrgOursRow}{rgb}{0.86, 0.92, 0.99}  
\newcommand{\rrgmatch}[1]{\textcolor{RrgMatch}{\textbf{#1}}}
\newcommand{\rrghall}[1]{\textcolor{RrgHall}{\textbf{#1}}}

\begin{figure*}[ht]
\centering
\scriptsize
\setlength{\tabcolsep}{4pt}
\renewcommand{\arraystretch}{1.1}

\noindent
\rrgmatch{matched finding}\quad\rrghall{hallucination}\quad{\itshape plain = filler / normal}\par
\vspace{0.5em}

\begin{tabularx}{\textwidth}{@{}l X@{}}
\toprule
\multicolumn{2}{@{}l}{\normalsize\textbf{(a) Mostly-normal abdominal CT}} \\
\midrule
\rowcolor{RrgGTRow}
\textbf{GT}
& Lower thorax: Normal. Liver and biliary tree: The liver is normal in size and contour. Redemonstration of hypoattenuating liver parenchyma compatible with \textbf{hepatic steatosis}. No focal liver lesions or biliary dilatation. Gallbladder: \textbf{Surgically absent}. Spleen: Normal. Pancreas: Normal. Adrenal glands: \textbf{Stable thickening of the left adrenal gland without discrete nodule}. The right adrenal gland appears normal. Kidneys and ureters: Normal. Bowel: Normal. The appendix appears normal. Peritoneal cavity: No free fluid. Bladder: Normal. Uterus and ovaries: Normal. Vasculature: Patent. Lymph nodes: Normal. Abdominal wall: \textbf{Postsurgical changes of ventral hernia repair with mesh}. Musculoskeletal: Normal. \\
\midrule
\rowcolor{RrgOursRow}
\textbf{\ours{}}
& Lower thorax: Normal. Liver and biliary tree: \rrgmatch{Hepatic steatosis. No focal hepatic lesions.} Gallbladder: \rrgmatch{Surgically absent}. Spleen: Normal. Pancreas: Normal. Adrenal glands: Normal. Kidneys and ureters: Normal. Gastrointestinal tract: Diverticulosis without evidence of acute diverticulitis. Peritoneal cavity: No free fluid. Bladder: Normal. Uterus and ovaries: Normal. Vasculature: Patent. Lymph nodes: Normal. Abdominal wall: Small fat-containing umbilical hernia. Musculoskeletal: Degenerative change of the spine. \\
\midrule
\textbf{Merlin}
& Lower thorax: normal. Liver and biliary tree: normal. Gallbladder: \rrgmatch{surgically absent}. Spleen: normal. Pancreas: normal. Adrenal glands: normal. Kidneys and ureters: normal. Gastrointestinal tract: no evidence of bowel obstruction. Normal appendix (3/274). Peritoneal cavity: no free fluid. Abdominal wall: normal. Bladder: normal. Uterus and ovaries: \rrghall{the uterus is surgically absent with bilateral salpingo-oophorectomy}. Vasculature: patent. Lymph nodes: normal. Musculoskeletal: no aggressive osseous lesion is identified in the abdomen or pelvis. \\
\midrule
\textbf{MedGemma}
& \rrghall{FINDINGS: The patient is status post right hip arthroplasty with associated hardware streak artifact obscuring portions of the pelvis and proximal femur. There appears to be mild left hip osteoarthritis characterized by joint space narrowing at the superior aspect of the acetabulum as well as subchondral sclerosis involving both femoral head and acetabular rim. No acute fracture or dislocation is identified within the visualized osseous structures. IMPRESSION: Mild left hip osteoarthritis. Status post right total hip replacement.} \\
\midrule
\textbf{Med3DVLM}
& \rrghall{There is a large, well-defined hypodense lesion occupying the entire right lobe of the liver. The lesion is predominantly fat attenuating with a few non-fat containing areas showing mild enhancement on post-contrast imaging.} \\
\bottomrule
\end{tabularx}

\vspace{0.7em}

\begin{tabularx}{\textwidth}{@{}l X@{}}
\toprule
\multicolumn{2}{@{}l}{\normalsize\textbf{(b) Complex post-pelvic-exenteration study}} \\
\midrule
\rowcolor{RrgGTRow}
\textbf{GT}
& Lower thorax: \textbf{Large left pleural effusion} with compressive atelectasis, progressed from prior. \textbf{Small right pleural effusion} with compressive atelectasis, improved from prior. There are \textbf{small bilateral pneumothoraces}; new pleural pigtail catheter on the right. Redemonstration of a small hiatal hernia. Liver and biliary tree: Normal. Gallbladder: \textbf{Surgically absent}. Spleen: Normal. Pancreas: Normal. Adrenal glands: Normal. Kidneys and ureters: Redemonstration of \textbf{ileal conduit}. Bowel: Extensive postsurgical changes of \textbf{pelvic exenteration} and small bowel anastomoses. \textbf{End colostomy at the left lower quadrant} and \textbf{ileal conduit at the right lower quadrant}. A few mildly prominent loops of small bowel within the pelvis (suggesting serositis/peritonitis), up to 3\,cm. Peritoneal cavity: \textbf{A rim-enhancing fluid collection in the pelvis with a surgical drain, $3.2 \times 9.8 \times 7.3$\,cm}, with rim enhancement and an air-fluid level. Interval placement of an anterior pigtail catheter; additional small rim-enhancing collections in the lower abdomen, anterior pelvis, and \textbf{perianal region extending to the gluteal fold}. Bladder: Normal. Uterus and ovaries: Surgically absent. Vasculature: \textbf{Atherosclerosis}. Lymph nodes: Normal. Abdominal wall: \textbf{Subcutaneous edema}. Some fluid subjacent to surgical incision. Musculoskeletal: Stable appearance with degenerative changes. \\
\midrule
\rowcolor{RrgOursRow}
\textbf{\ours{}}
& Lower thorax: \rrgmatch{small right pleural effusion with associated atelectasis} (left side missed). Liver and biliary tree: normal. No focal hepatic lesions. Gallbladder: \rrghall{Normal}. Spleen: \rrghall{Surgically absent}. Pancreas: Normal. Adrenal glands: Normal. Kidneys and ureters: enhance symmetrically without hydronephrosis (ileal conduit missed). Gastrointestinal tract: \rrghall{Postsurgical changes of total colectomy with ileoanal anastomosis.} \rrgmatch{There is a left lower quadrant end colostomy.} \rrghall{Diffuse bowel wall thickening involving the rectum and sigmoid colon} (rectum is absent post-exenteration). Peritoneal cavity: \rrgmatch{There is a large amount of ascites within the abdomen and pelvis} (discrete collections collapsed into ascites). Bladder: Foley catheter within the bladder lumen. Uterus and ovaries: \rrgmatch{Surgically absent.} Vasculature: Patent. Lymph nodes: \rrghall{Multiple prominent mesenteric lymph nodes measuring up to 10\,mm.} Abdominal wall: \rrgmatch{Midline abdominal incision with overlying skin staples.} Musculoskeletal: \rrgmatch{Degenerative changes of the spine.} \\
\midrule
\textbf{Merlin}
& Lower thorax: \rrgmatch{small right pleural effusion with associated compressive atelectasis} (left missed). The left lung base appears clear. Liver and biliary tree: the liver is normal in size. \rrghall{Scattered subcentimeter hypodensities throughout both lobes of the liver, which likely represent cysts or hamartomas (3/109).} No intrahepatic ductal dilatation. Gallbladder: \rrgmatch{surgically absent}. Spleen: normal. Pancreas: normal. Adrenal glands: normal. Kidneys and ureters: normal. Gastrointestinal tract: \rrghall{postsurgical changes of a right hemicolectomy with ileocolic anastomosis.} No evidence for bowel obstruction or inflammation. Peritoneal cavity: no free fluid. Abdominal wall: normal. Bladder: normal. Uterus and ovaries: surgically absent. Vasculature: patent. Lymph nodes: normal. Musculoskeletal: no aggressive osseous lesion is identified in the abdomen or pelvis. \\
\midrule
\textbf{MedGemma}
& FINDINGS: The patient is rotated to their left side with significant cardiac and mediastinal shift towards the right. There appears to be an artifact overlying the heart which may represent surgical material or calcification within the pericardium/pleura. \rrghall{A large amount of fluid is present throughout both pleural spaces (hemothorax). Multiple metallic foreign bodies are noted scattered throughout the chest wall including near the sternum, ribs, and axilla bilaterally.} No definite pneumothorax identified. IMPRESSION: \rrghall{Large bilateral hemothoraces with associated mass effect causing cardiomediastinal shift. Metallic foreign body related injuries involving multiple structures.} \\
\midrule
\textbf{Med3DVLM}
& \rrghall{CT confirms a large hematoma with active hemorrhage but no contrast extravasation, as demonstrated on the delayed imaging.} \\
\bottomrule
\end{tabularx}

\caption{\textbf{Qualitative comparison of report generation on two held-out \stanford{} abdominal CT studies}: a mostly-normal study \textbf{(a)} and a complex post-pelvic-exenteration study \textbf{(b)}. \rrgmatch{Green} marks findings supported by the radiologist ground truth; \rrghall{red} marks hallucinations; plain text is filler / normal phrasing.}
\label{fig:rrg-qualitative-stanford}
\end{figure*}

%% file: figures/pca.tex
\begin{figure}
    \centering
    \setlength{\tabcolsep}{1pt}
    \renewcommand{\arraystretch}{1.1}

    \newlength{\imgwidth}
    \setlength{\imgwidth}{\dimexpr (\textwidth - 20\tabcolsep) / 6 \relax}

    \newcommand{\modelhdr}[1]{\footnotesize\shortstack{#1}}

    \begin{tabular}{@{} *{6}{c} @{}}
        \modelhdr{Merlin} &
        \modelhdr{Pillar-0} &
        \modelhdr{Baseline\\ CLIP (Atlas)} &
        \modelhdr{Baseline\\ CLIP (ResNet)} &
        \modelhdr{\ours{} (Atlas)} &
        \modelhdr{\ours{} (ResNet)} \\

        \includegraphics[width=\imgwidth]{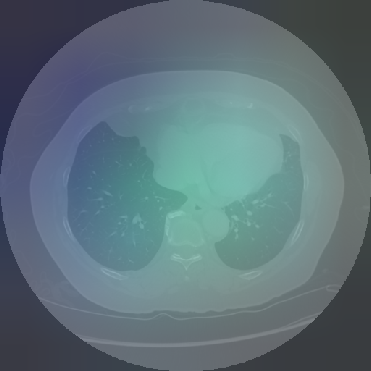} &
        \includegraphics[width=\imgwidth]{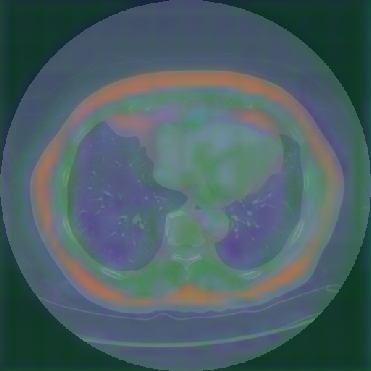} &
        \includegraphics[width=\imgwidth]{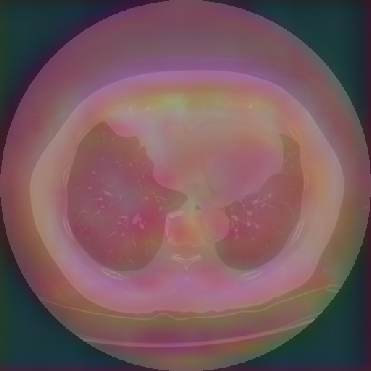} &
        \includegraphics[width=\imgwidth]{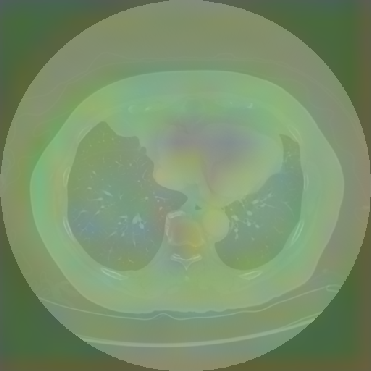} &
        \includegraphics[width=\imgwidth]{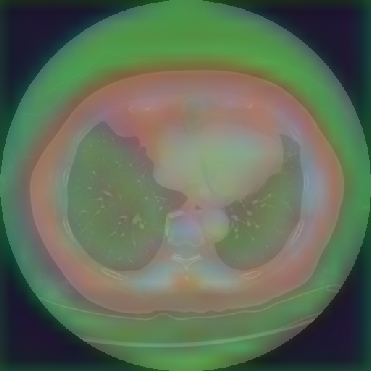} &
        \includegraphics[width=\imgwidth]{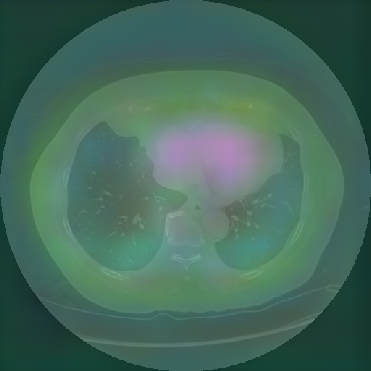} \\

        \includegraphics[width=\imgwidth]{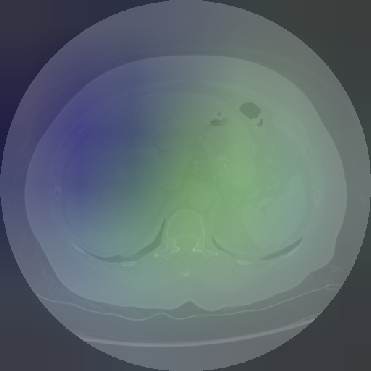} &
        \includegraphics[width=\imgwidth]{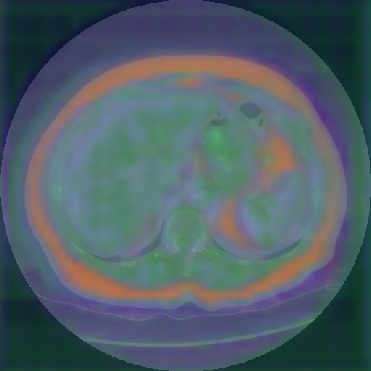} &
        \includegraphics[width=\imgwidth]{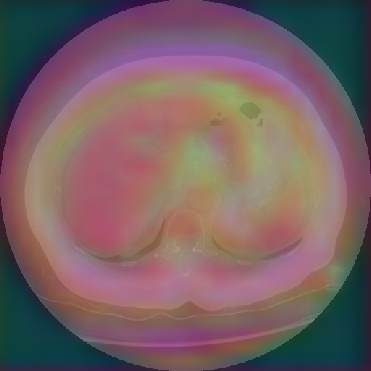} &
        \includegraphics[width=\imgwidth]{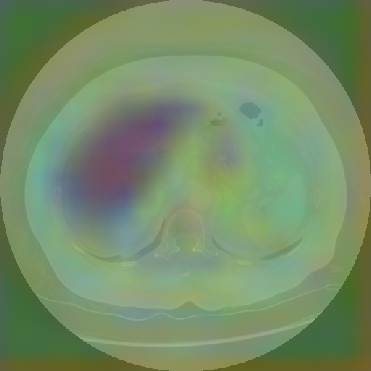} &
        \includegraphics[width=\imgwidth]{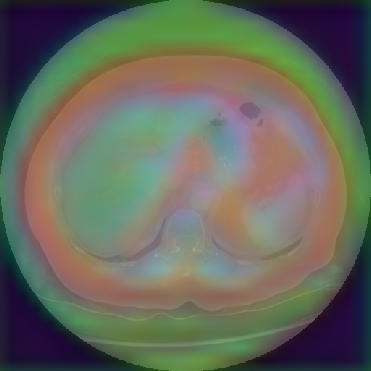} &
        \includegraphics[width=\imgwidth]{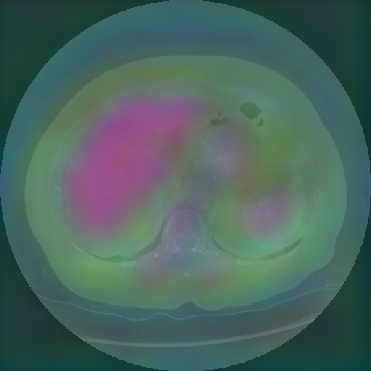} \\

        \includegraphics[width=\imgwidth]{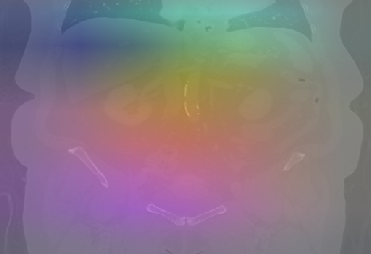} &
        \includegraphics[width=\imgwidth]{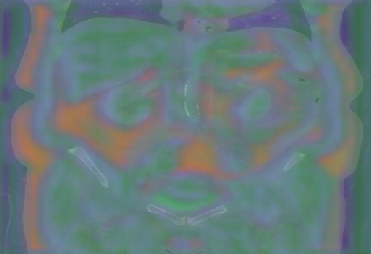} &
        \includegraphics[width=\imgwidth]{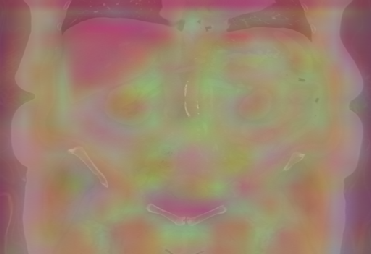} &
        \includegraphics[width=\imgwidth]{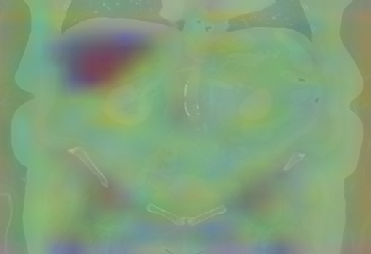} &
        \includegraphics[width=\imgwidth]{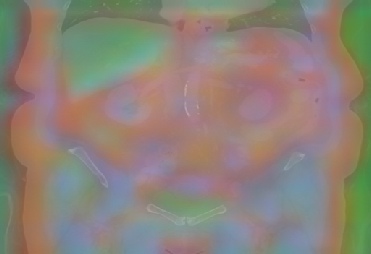} &
        \includegraphics[width=\imgwidth]{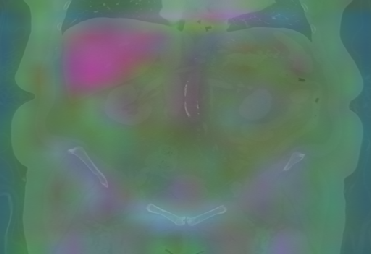} 
    \end{tabular}

    \caption{\textbf{Principal Component Analysis Visualization.} First three non-trivial PCA components of the high-resolution feature maps of the same volume mapped to RGB channels. \ours{} models produce the most spatially structured and anatomically coherent representations across all views.}
    \label{fig:pca}
\end{figure}

%% file: figures/attention_map_fail.tex
\begin{figure}[h]
    \centering
    \setlength{\tabcolsep}{0.001pt} 
    \renewcommand{\arraystretch}{0} 

    \setlength{\gridwidth}{\dimexpr (\textwidth - 3em - 10\tabcolsep) / 4 \relax}

    \newcommand{\gridhdr}[1]{\footnotesize\textbf{#1}}
    \newcommand{\rowlabel}[1]{\rotatebox{90}{\makebox[0pt]{\gridhdr{#1}}}}

    \begin{tabular}{@{} c @{\hspace{4pt}} cccc @{}}
        & \multicolumn{2}{c}{\gridhdr{Pancreas}} & \multicolumn{2}{c}{\gridhdr{Spleen}} \\[6pt]

        \rowlabel{Axial} &
        \includegraphics[width=\gridwidth, align=c]{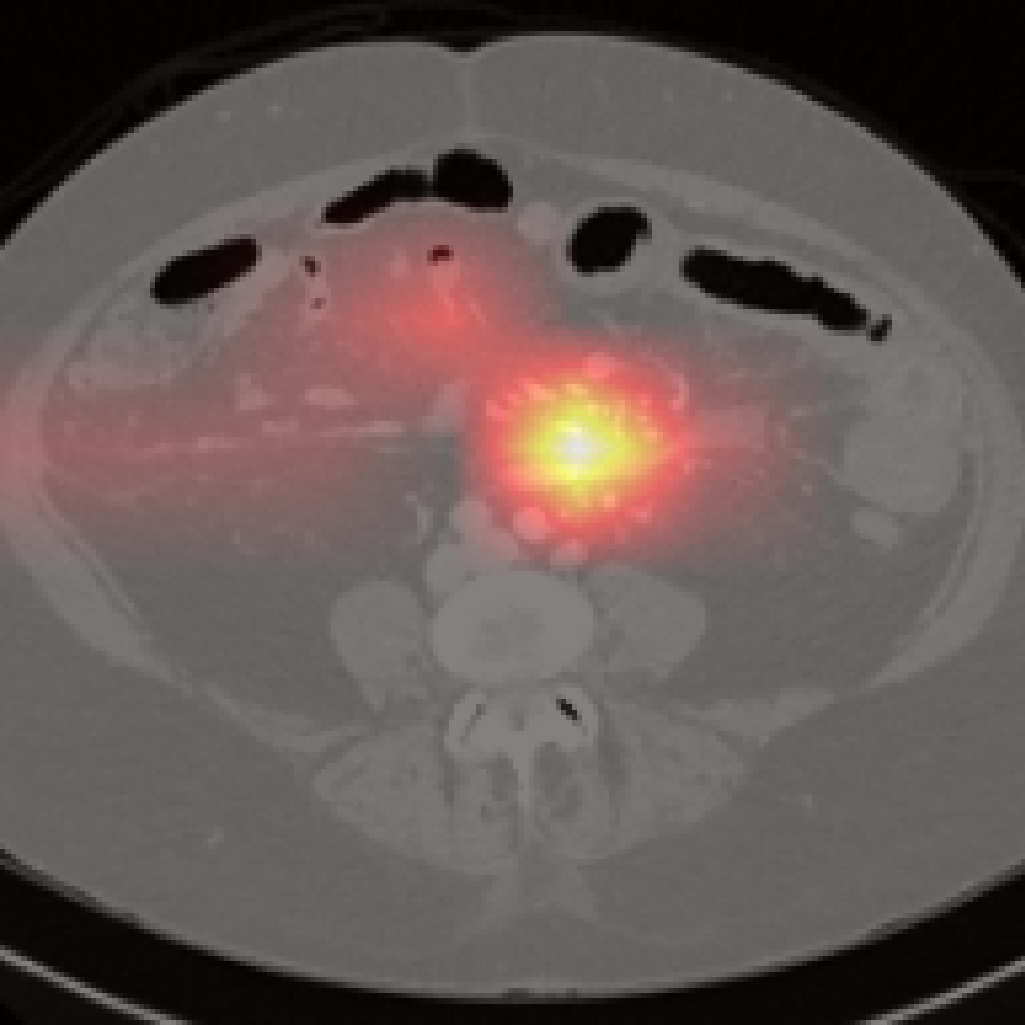} &
        \includegraphics[width=\gridwidth, align=c]{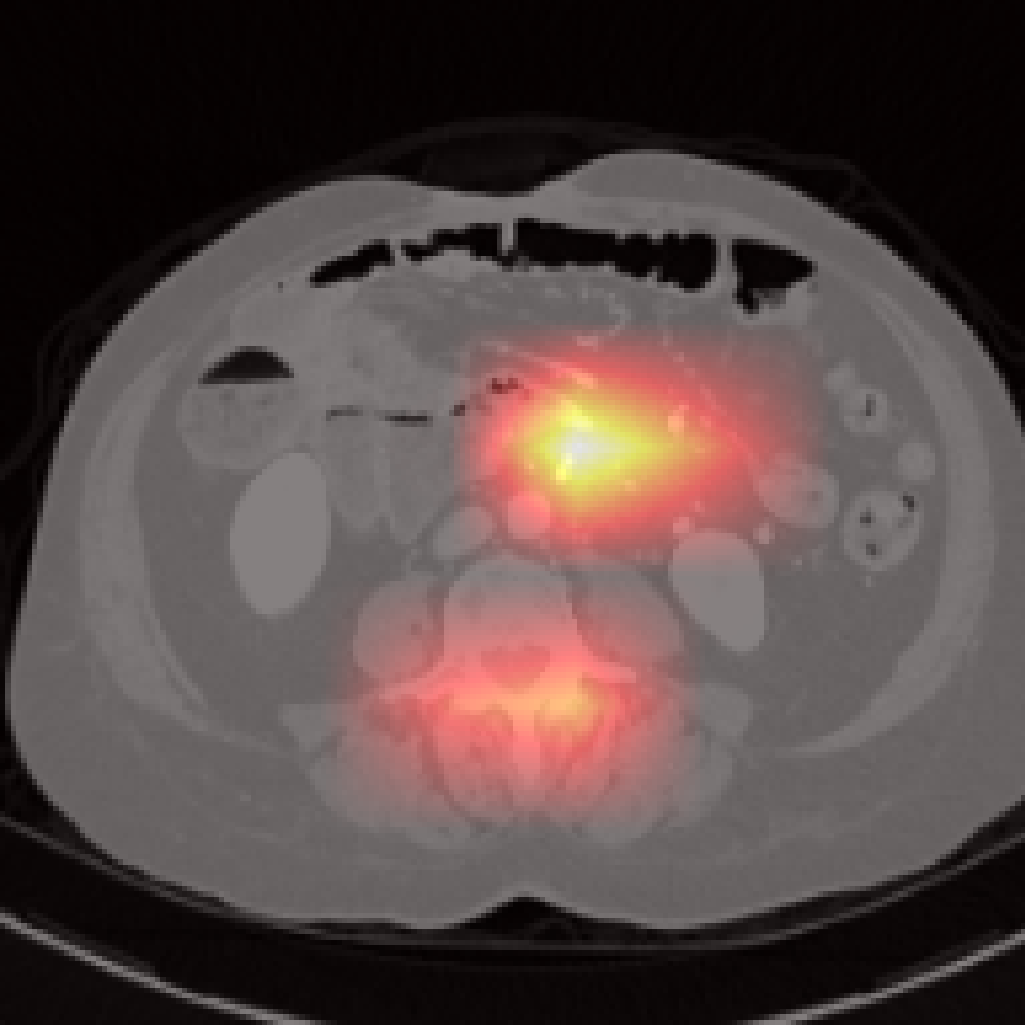} &
        \includegraphics[width=\gridwidth, align=c]{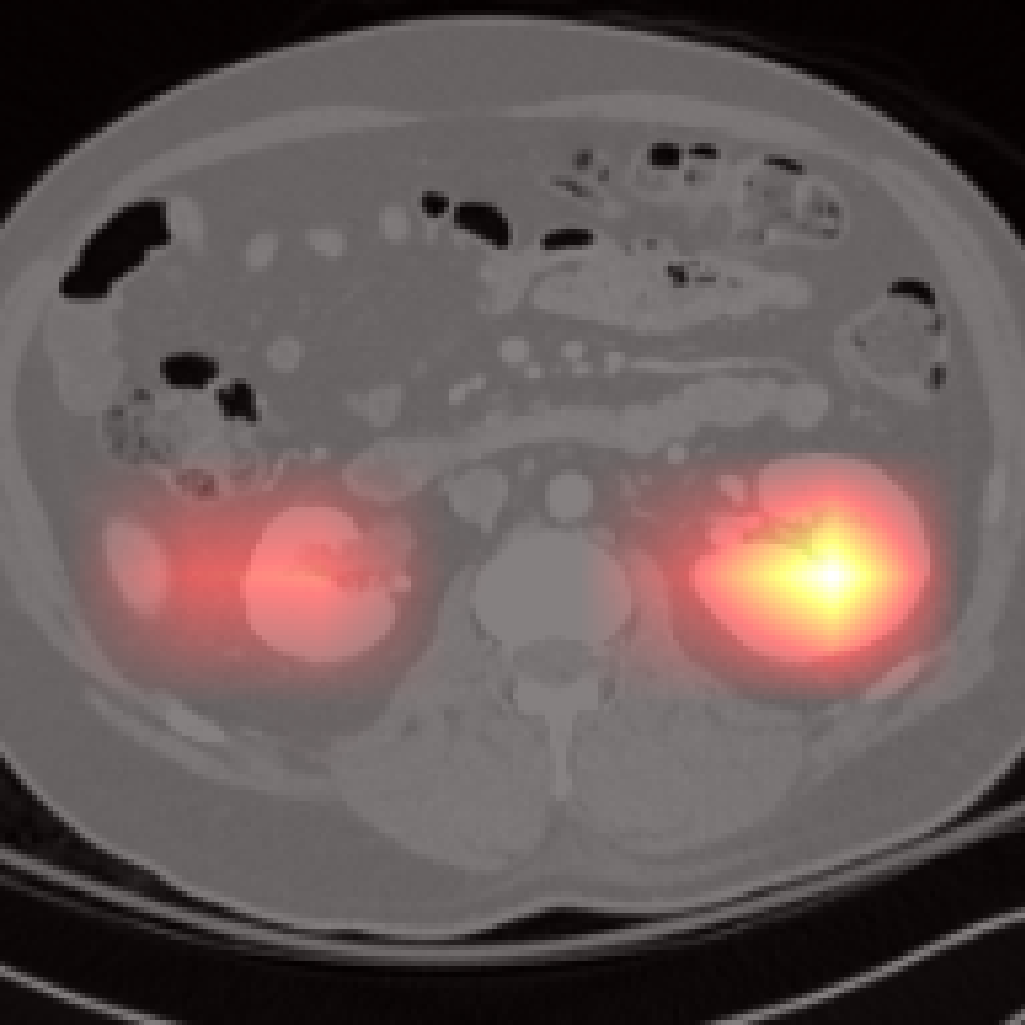} &
        \includegraphics[width=\gridwidth, align=c]{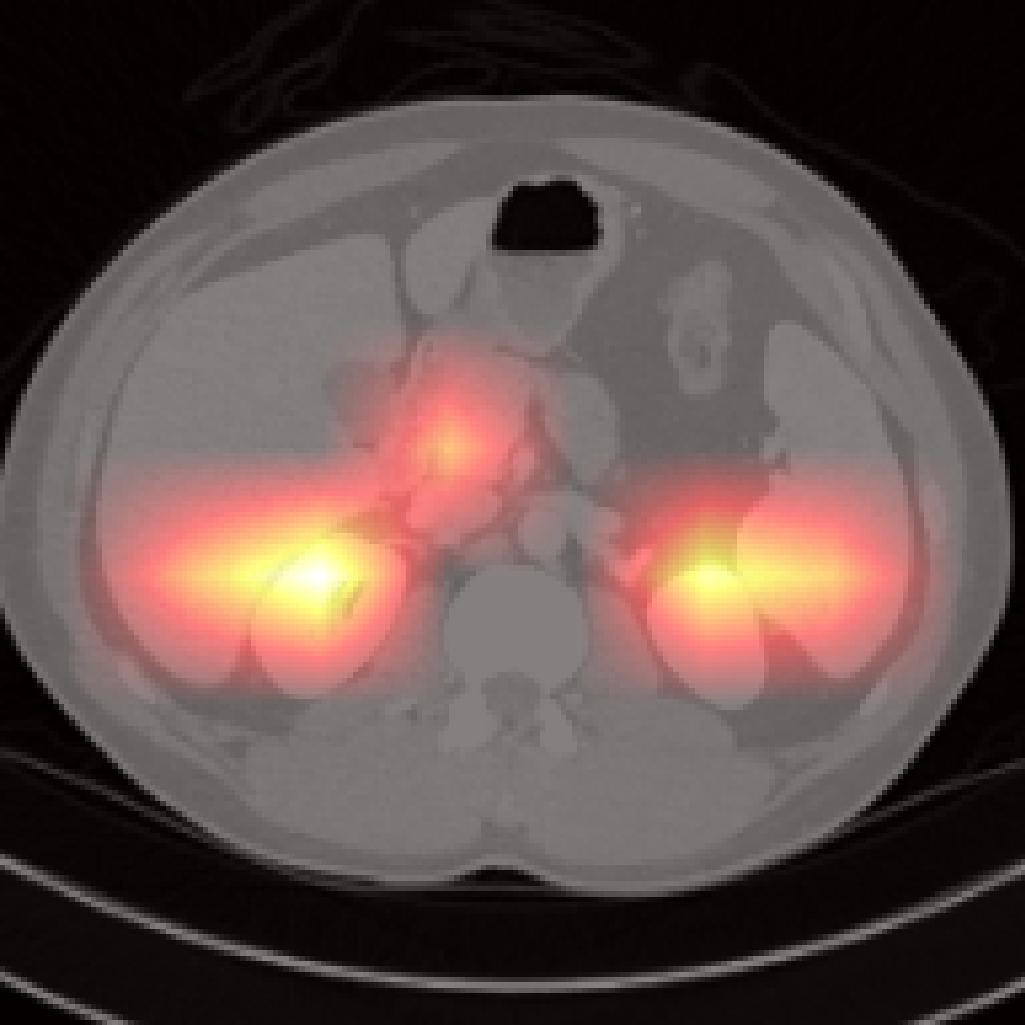} \\[2\tabcolsep]

        \rowlabel{Coronal} &
        \includegraphics[width=\gridwidth, align=c]{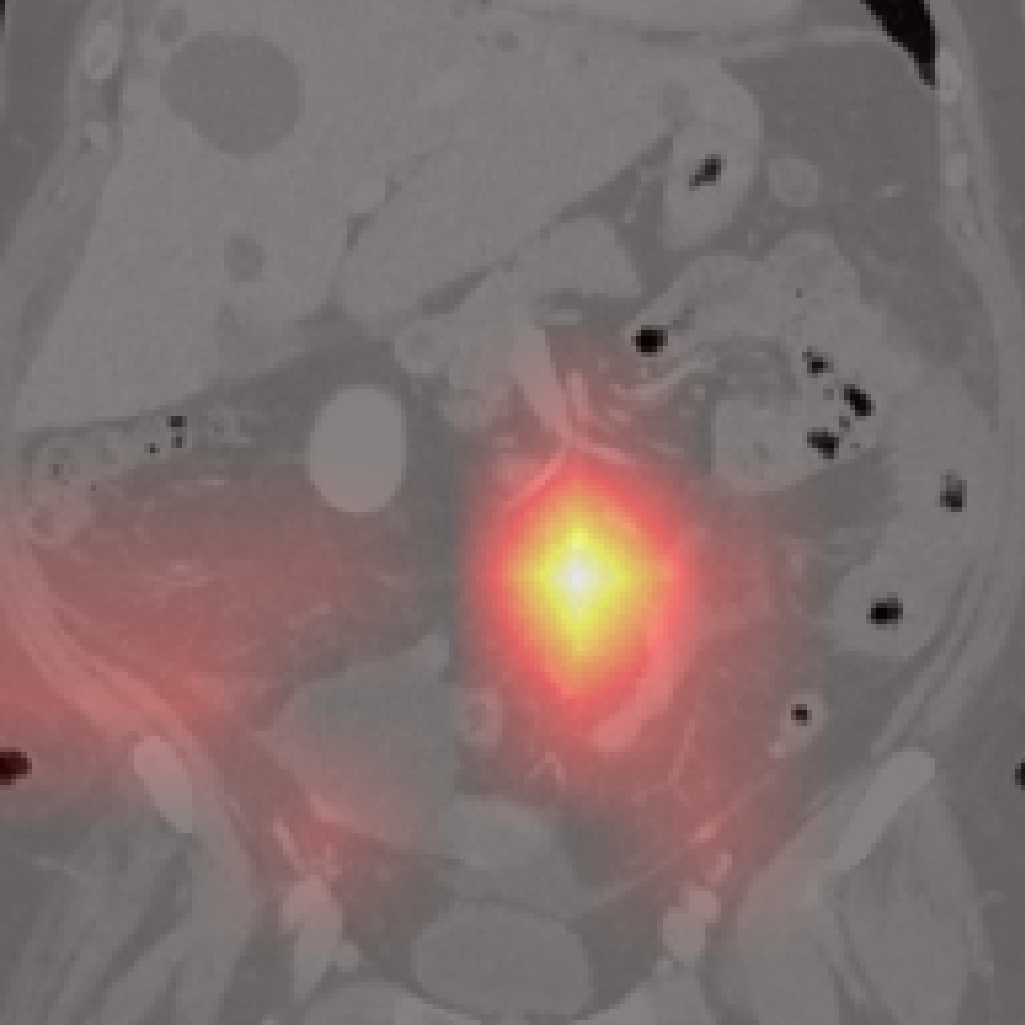} &
        \includegraphics[width=\gridwidth, align=c]{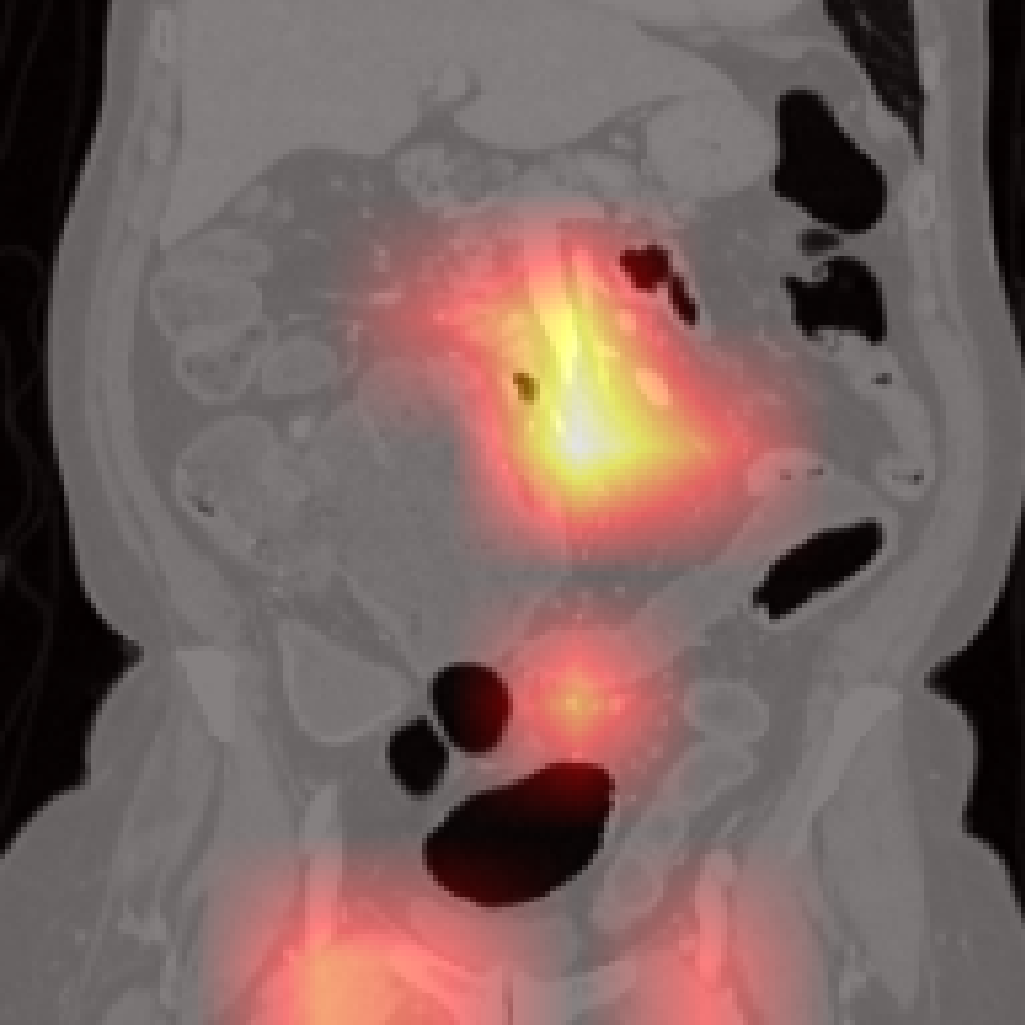} &
        \includegraphics[width=\gridwidth, align=c]{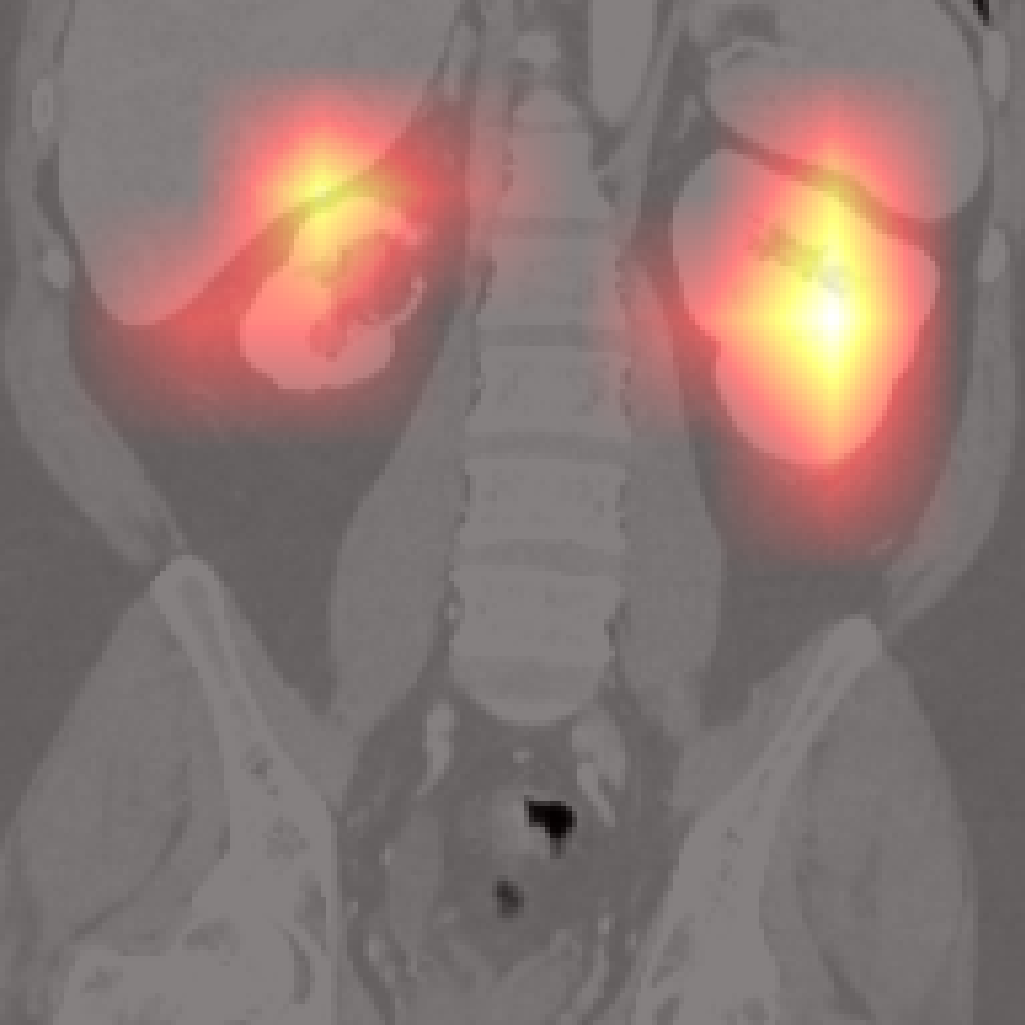} &
        \includegraphics[width=\gridwidth, align=c]{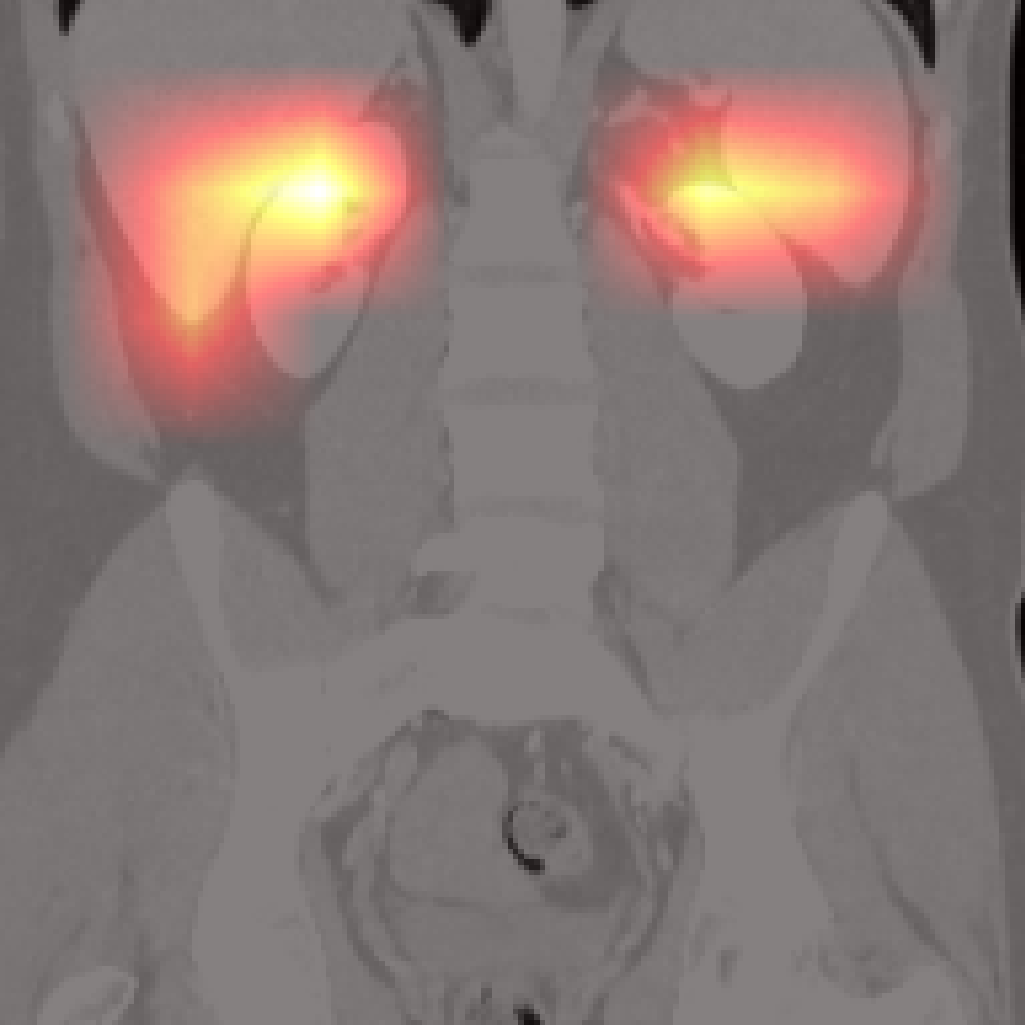} \\[2\tabcolsep]

        \rowlabel{Sagittal} &
        \includegraphics[width=\gridwidth, align=c]{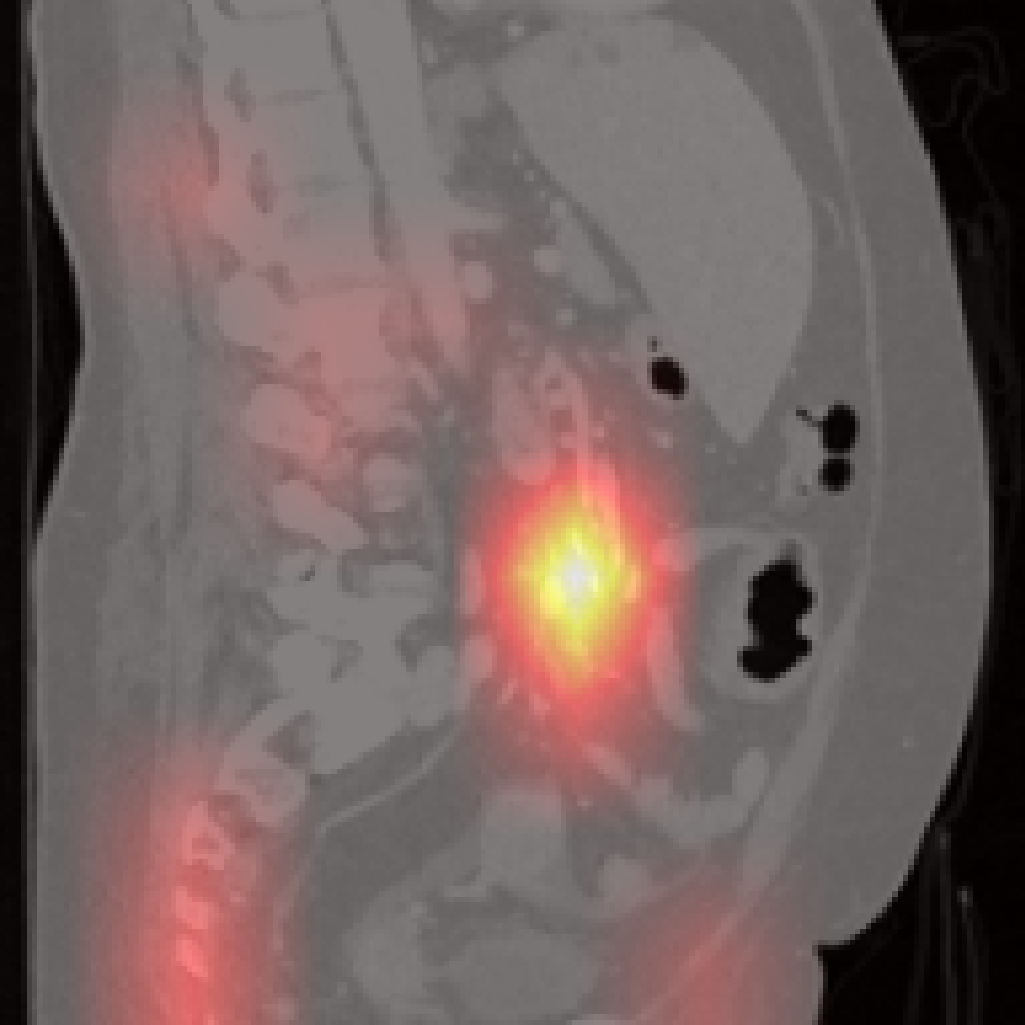} &
        \includegraphics[width=\gridwidth, align=c]{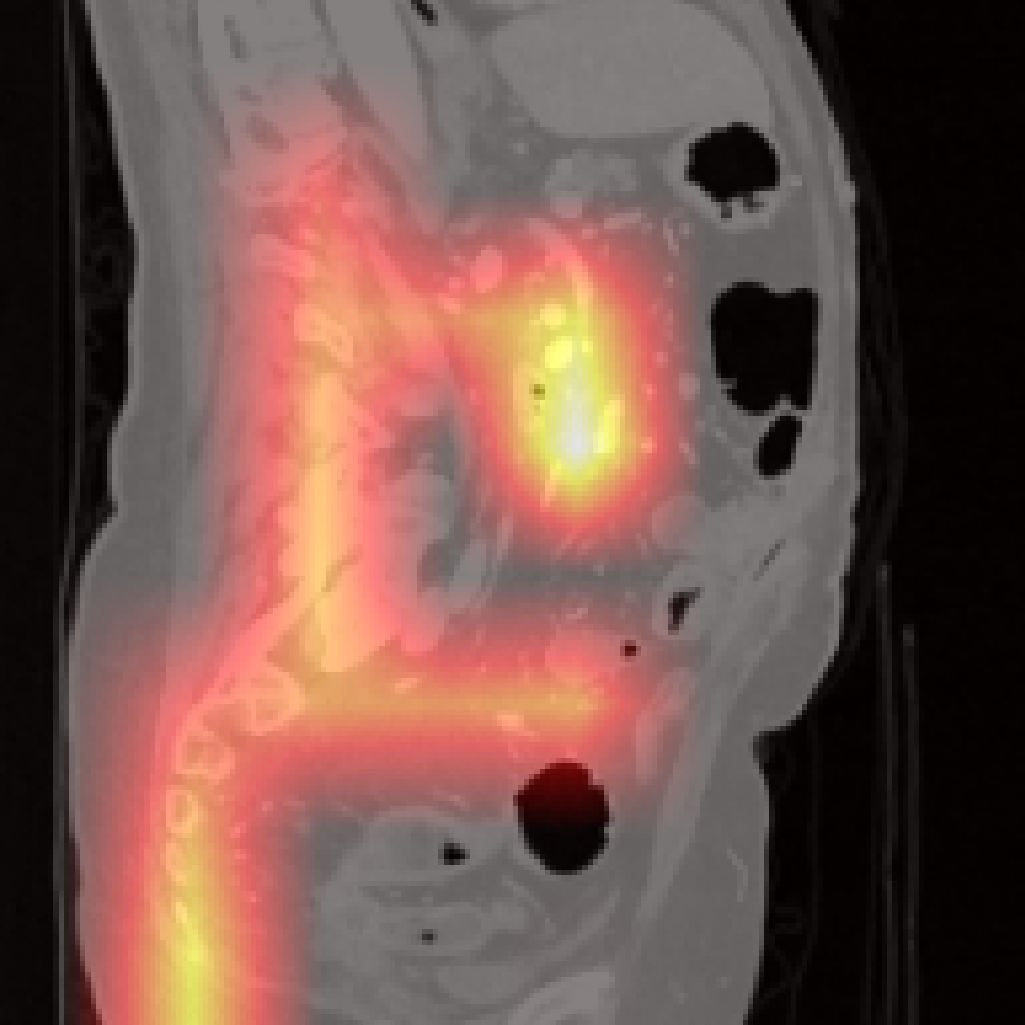} &
        \includegraphics[width=\gridwidth, align=c]{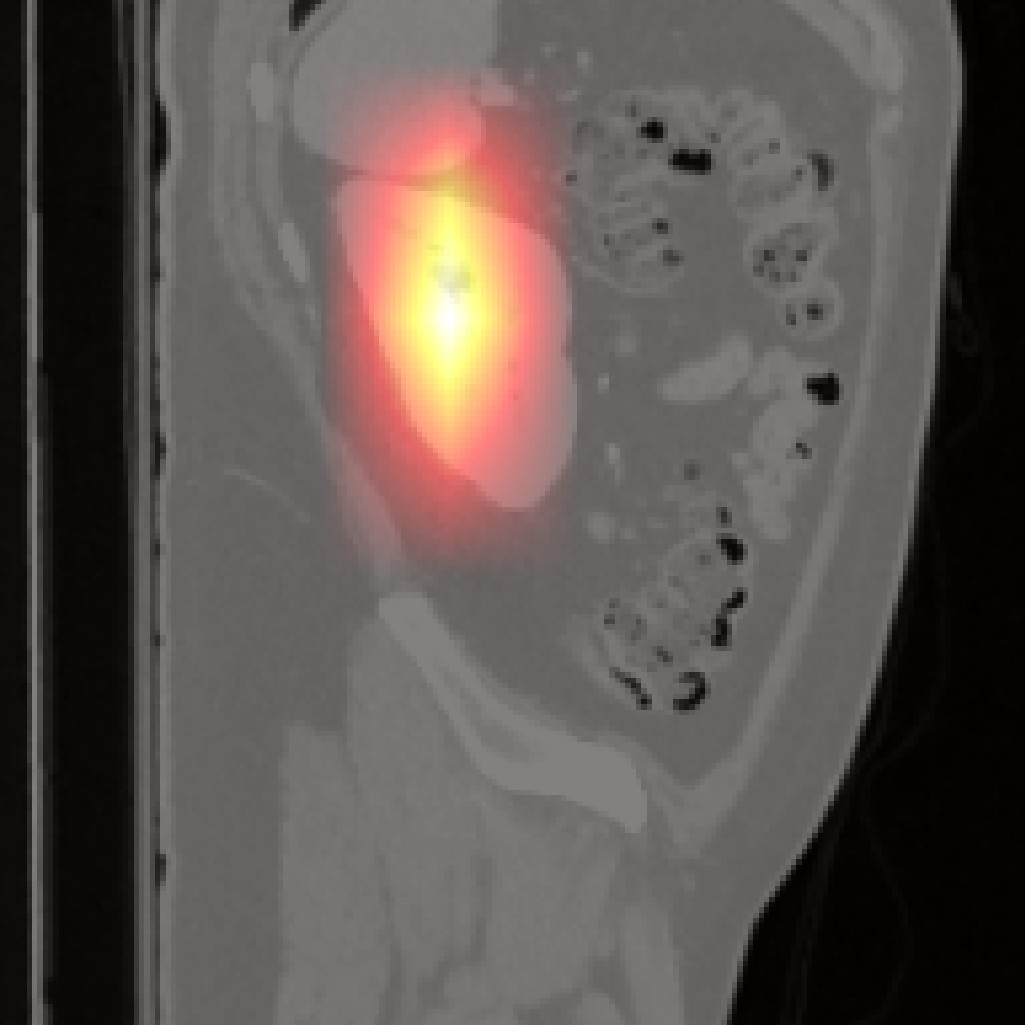} &
        \includegraphics[width=\gridwidth, align=c]{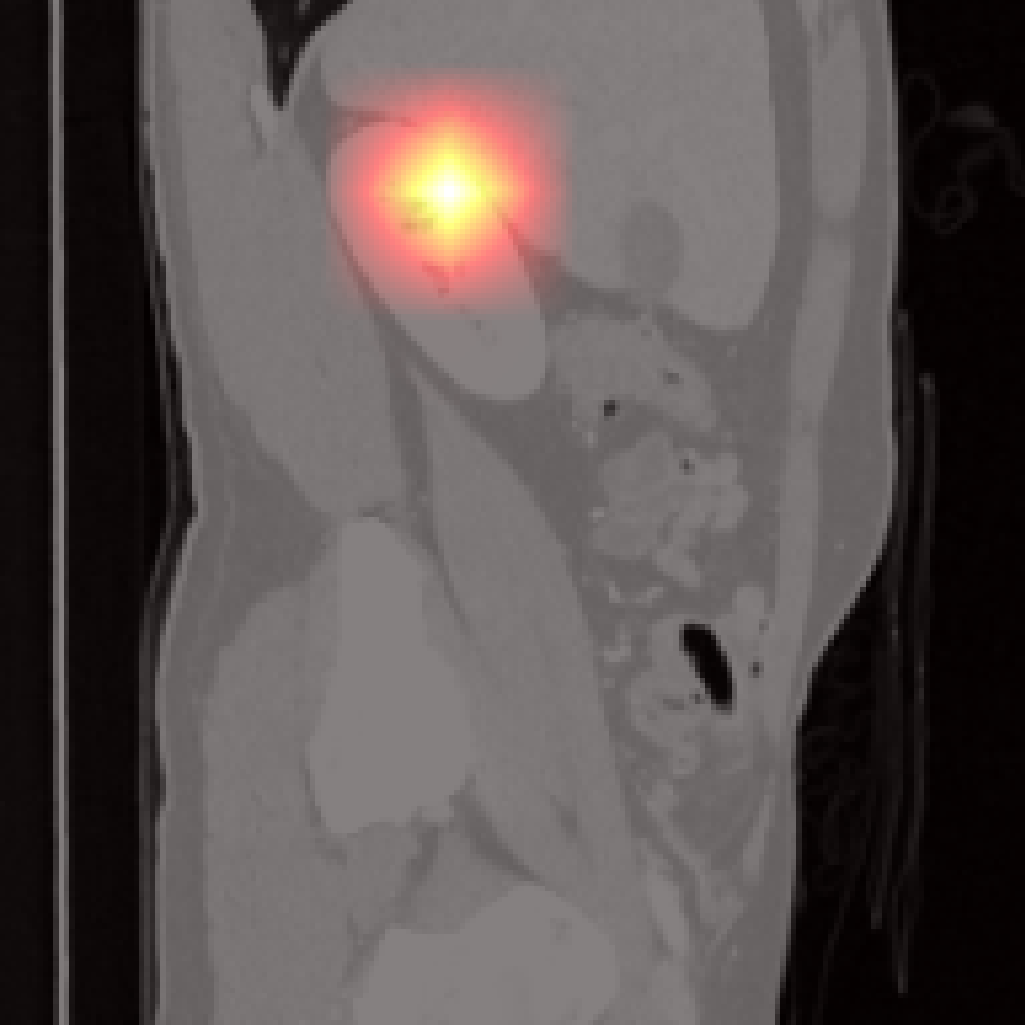} \\
    \end{tabular}

    \caption{\textbf{Failure cases in organ-level attention.} Cross-attention maps for the pancreas and spleen demonstrate lower localization precision compared to other organs.}
    \label{fig:attention_maps_fail}
\end{figure}